\documentclass{article}

\usepackage[preprint, nonatbib]{neurips_2022}

\usepackage[utf8]{inputenc} 
\usepackage[T1]{fontenc}    
\usepackage[colorlinks = true,
            linkcolor = royalazure,
            urlcolor  = royalazure,
            citecolor = royalazure,
            anchorcolor = royalazure]{hyperref}
\usepackage{url}            
\usepackage{booktabs}       
\usepackage{amsfonts}       
\usepackage{nicefrac}       
\usepackage{microtype}      
\usepackage{xcolor}         
\usepackage{authblk}

\usepackage{amsmath,amsfonts,bm}
\newcommand{\etal}{{\emph{et al. }}}

\def\eqref#1{equation~\ref{#1}}

\def\1{\bm{1}}

\DeclareMathAlphabet{\mathsfit}{\encodingdefault}{\sfdefault}{m}{sl}
\SetMathAlphabet{\mathsfit}{bold}{\encodingdefault}{\sfdefault}{bx}{n}

\DeclareMathOperator*{\argmax}{argmax}

\usepackage{graphicx}
\usepackage{caption}
\usepackage{subcaption}
\usepackage{soul}

\usepackage[capitalize,noabbrev]{cleveref}
\usepackage{amsthm}

\theoremstyle{definition}  

\usepackage{enumitem}
\usepackage{booktabs}       
\usepackage{multirow}
\usepackage{multicol}
\usepackage{color}
\usepackage{adjustbox}
\usepackage{wrapfig}
\definecolor{royalazure}{rgb}{0.0, 0.22, 0.66}

\definecolor{ballblue}{rgb}{0.13, 0.67, 0.8}

\usepackage{pifont}

\title{Edge Rewiring Goes Neural: Boosting Network Resilience without Rich Features}

\author[1]{Shanchao~Yang}
\author[2]{Kaili~Ma}
\author[1]{Baoxiang~Wang}
\author[1]{Tianshu~Yu}
\author[1, 3]{Hongyuan~Zha}

\affil[1]{School of Data Science,
The Chinese University of Hong Kong, Shenzhen \authorcr
  \tt shanchaoyang@link.cuhk.edu.cn 
  \\
  \{\tt bxiangwang, yutianshu\}@cuhk.edu.cn
  }
\affil[2]{Department of Computer Science and Engineering, The Chinese University of Hong Kong \authorcr
  \tt klma@cse.cuhk.edu.hk}
\affil[3]{Shenzhen Institute of Artificial Intelligence and Robotics for Society 
\authorcr
  \tt zhahy@cuhk.edu.cn
   }

\begin{document}

\maketitle

\begin{abstract}

Improving the resilience of a network is a fundamental problem in network science, which protects the underlying system from natural disasters and malicious attacks. 
This is traditionally achieved via successive degree-preserving edge rewiring operations, with the major limitation of being transductive.
Inductively solving graph-related tasks with sequential actions is accomplished by adopting graph neural networks (GNNs) coupled with reinforcement learning under the scenario with rich graph features.
However, such frameworks cannot be directly applied to resilience tasks where only pure topological structure is available. In this case, GNNs can barely learn useful information, resulting in prohibitive difficulty in making actions for successively rewiring edges under a reinforcement learning context.
In this paper, we study in depth the reasons why typical GNNs cause such failure.
Based on this investigation, we propose \textbf{ResiNet}, the first end-to-end trainable inductive framework to discover \textbf{Resi}lient \textbf{Net}work topologies while balancing network utility.
To this end, we reformulate resilience optimization as an MDP equipped with edge rewiring action space, and propose a pure topology-oriented variant of GNN called \textbf{Fi}lt\textbf{r}ation \textbf{e}nhanced \textbf{G}raph \textbf{N}eural \textbf{N}etwork (\textbf{FireGNN}), which can learn from graphs without rich features. 
Extensive experiments demonstrate that ResiNet achieves a near-optimal resilience gain on various graphs while balancing the utility, and outperforms existing approaches by a large margin.

\end{abstract}

\section{Introduction}

Network systems, such as infrastructure systems, supply chains, routing networks, and peer-to-peer computing networks, are vulnerable to malicious attacks.
 \textit{Network resilience}, in the context of network science, is a measurement characterizing the ability of a system to defend itself from such failures and attacks~\cite{schneider2011mitigation}.
A resilient network should continue to function and maintain an acceptable level of utility when the network partially fails. Therefore, resilience is crucial when designing new systems or upgrading existing systems toward high reliability.


 

Current network resilience optimization methods improve the resilience of networks over graph topologies based on an atomic operation called \emph{edge rewiring}~\cite{schneider2011mitigation, chan2016optimizing, rong2018heuristic}. Concretely, for a given graph $G=(V,E)$ and two existing edges $AC$ and $BD$, an edge rewiring operation alters the graph structure by removing $AC$ and $BD$ and adding $AB$ and $CD$, where $AC, BD\in E$ and $AB, CD, AD, BC\notin E$.
Edge rewiring operation has some nice properties against simply addition or deletion of edges: 1) it preserves the node degree--the number of edges and the total degree of a graph, while addition/deletion may cause change; 2) it achieves minimal changes in terms of utility measure in terms of graph Laplacian, while addition/deletion may lead to an unpredictable network utility degradation \cite{DBLP:conf/kdd/0001WDWT21, jaume20202}. The challenge of the resilience task lies in determining which two edges are selected for rewiring, with the complexity of $O(E^{2T})$ for $T$ successive steps of edge rewiring.

Traditional non-learning-based resilience optimization methods typically fall into the categories of evolutionary computation \cite{zhou2014memetic} or heuristic-based \cite{schneider2011mitigation, chan2016optimizing, rong2018heuristic, yaziciouglu2015formation}, with the following limitations:

\begin{itemize}
\item \textit{Transductivity}. Traditional methods are transductive since they search the resilience topology on a particular problem instance. This search procedure is performed for every individual graph without generalization. 

\item \textit{Local optimality}. 
It is NP-hard to combinatorially choose the edges to rewire to obtain the globally optimal resilience \cite{mosk2008maximum}. Previous studies predominantly adopt greedy-like algorithms, yielding local optimality in practice \cite{chan2016optimizing}.

\item \textit{Utility Loss}. Rewiring operation in resilience optimization may potentially lead to an degradation of the network utility, which may jeopardize the network functioning. 

\end{itemize}

Although the learning-based paradigm equipped with GNNs has proved powerful for a large variety of graph tasks with rich features~\cite{guided_tree_search, effcient_TSP, TSP_pretrain, CO_on_graph, rl_for_vrp, rl_for_vrp_with_attention, online_vrp}, it still remains opaque how to effectively adapt such approaches to resilience optimization where only topological structure is available. One may infer that such a lack of feature can significantly hinder the learning ability, and a case agreeing with this has been discovered in solving the traveling salesman problem (TSP):
Boffa \etal showed that the performance of GNNs degenerates largely when node coordinates are missing (but pairwise distance is given), compared to the case when coordinates are available \cite{boffa2022neural}. 
Although Boffa \etal adopted a distance encoding strategy to alleviate the performance gap in TSP \cite{boffa2022neural, li2020distance}, we empirically found that this encoding strategy is not working well for the more challenging resilience task (See Sec. \ref{sec:baseline}). As such, it is demanding to devise a novel method that can be applicable for the resilience task without rich features. Readers are referred to Appendix \ref{sec:whygnnfails} for a more detailed analysis.

In this work, we present ResiNet, the first \textit{inductive} learning-based framework for discovering resilient network topology using successive edge rewiring operations. To overcome the above limitation of GNNs in modeling graphs without rich features, we specially design a topology-oriented variant of GNN called \textbf{Fi}lt\textbf{r}ation \textbf{e}nhanced \textbf{GNN} (FireGNN). FireGNN creates a series of subgraphs (the filtration) by successively removing the node with the highest degree from the graph and then learns to aggregate the node representations from each subgraph. This filtration process innovation is inspired by persistent homology and the approximation of the persistence diagram \cite{edelsbrunner2008persistent, aktas2019persistence, hofer2020graph}.  




The main contributions of this paper are summarized as follows:
\begin{enumerate}[label=\arabic*)]
\item We propose ResiNet, the first data-driven learning-based framework to boost network resilience inductively in a degree-preserving manner with moderate loss of the network utility. ResiNet forms resilience optimization into a successive sequential decision process of edge rewiring operations. Extensive experiments show 
that ResiNet achieves near-optimal resilience gain while balancing network utilities. Existing approaches are outperformed by a large margin.

\item  FireGNN, our technical innovation serving as the graph feature extractor, is capable of learning meaningful representations from pure topological structures, which provides sufficient training signals to learn an RL agent to perform successive edge rewiring operations inductively.

\end{enumerate}

\section{Related work}
\paragraph{Network resilience.}
Modern network systems are threatened by various malicious attacks, such as the destruction of critical nodes, critical connections and critical subset of the network via heuristics/learning-based attack~\cite{fan2020finding,zhao2021,fan2017,holme2002attack, iyer2013attack,grassia2021machine,fan2020finding,KCM}. 
Network resilience was proposed and proved as a suitable measurement for describing the robustness and stability of a network system under such attacks \cite{schneider2011mitigation}.
Around optimizing network resilience, various defense strategies have been proposed to protect the network functionality from crashing and preserve network's topologies to some extent. Commonly used manipulations of defense include adding additional edges~\cite{li2019maximizing, carchiolo2019network}, protecting vulnerable edges~\cite{wang2014improving} and rewiring two edges~\cite{schneider2011mitigation, chan2016optimizing, buesser2011optimizing}. Among these manipulations, edge rewiring fits well to real-world applications as it induces fewer functionality changes to the original network and does not impose additional loads to the vertices (degree-preserving)~\cite{schneider2011mitigation, rong2018heuristic, yaziciouglu2015formation}. By now, there has been no learning-based inductive strategy for the resilience task.

\paragraph{GNNs for graph-related tasks.}

GNNs are powerful tools to learn from relational data, providing meaningful representations for the downstream task.
Several successful applications using GNNs as backbones include node classification \cite{kipf2016semi, hamilton2017inductive, DBLP:conf/iclr/VelickovicCCRLB18},  link prediction \cite{li2020deepergcn, kipf2016semi, hamilton2017inductive}, graph property estimation \cite{xu2018how, kipf2016semi, li2020deepergcn, bodnar2021weisfeiler}, and combinatorial problems on graphs (e.g., TSP~\cite{guided_tree_search, effcient_TSP, TSP_pretrain, CO_on_graph, hudson2021graph}, vehicle routing problem~\cite{rl_for_vrp, rl_for_vrp_with_attention, online_vrp}, graph matching~\cite{yu2021deep} and adversarial attack on GNNs \cite{DBLP:conf/kdd/0001WDWT21,dai2018adversarial}). 
Although standard message passing GNNs are powerful, their expressive power is upper-bounded by the 1-Weisfeiler-Lehman (1-WL) test \cite{xu2018how}. 
Many advanced techniques are designed for enabling GNNs with greater expressive power over the 1-WL test, such as distance encoding \cite{li2020distance} and high-order GNNs \cite{morris2019weisfeiler}.
Such strategies are empirically observed to be more expressive than 1-WL versions under rare feature cases \cite{you2021identity, boffa2022neural}.
For example, distance encoding was employed by Boffa~\etal in a learning-based TSP solver, achieving better performance than standard GNNs when only pairwise distance is given \cite{boffa2022neural}.
However, it still obviously under-performs compared to the case when GNNs are fully trained with node coordinates. Till now,
it remains prohibitively challenging to adapt GNNs to graphs effectively
without rich feature, or even with pure topology, as in the resilience task.



\paragraph{Extended related work.} 

The related work on \textit{network resilience and utility}, \textit{multi-views graph augmentation for GNNs} and \textit{deep graph generation} is deferred to Appendix~\ref{appendix:related}.

\section{Problem definition}

An undirected graph is defined as $G= (V, E)$, where $V = \{1, 2, \dots, N\}$ is the set of $N$ nodes, $E$ is the set of $M$ edges, $A \in \{0, 1\}^{N\times N}$ is the adjacency matrix, and $F \in \mathbb{R}^{N \times d}$ is the $d$-dimensional node feature matrix\footnote{A graph without rich feature only has the topology structure and the node feature is not available.}. The degree of a node is defined as $d_i = \sum_{j=1}^N A_{ij}$, and a node with degree 0 is called an isolated node.


Given the network resilience metric $\mathcal{R}(G)$ and the utility metric $\mathcal{U}(G)$, let $\mathbb{G}_G$ denote the set of graphs with the \textit{same} node degrees as $G$.  
The objective of boosting the resilience of $G$ is to find a target graph $G^{\star} \in \mathbb{G}_G $, which maximizes the network resilience while preserving the network utility. Formally, the problem of maximizing network resilience is formulated as
\begin{equation*}
\label{eq:maximizingrobustness}
  G^{\star} = \mathop{\arg \max}_{G^{'} \in \mathbb{G}_G}\limits 
 \,  \alpha \cdot \mathcal{R}(G^{'}) + (1 - \alpha) \cdot \mathcal{U}(G^{'})   \, ,
\end{equation*}
where $\alpha\in\mathbb{R}$ is the scalar weight that balances the resilience and the utility.

\begin{wrapfigure}{r}{0.4\textwidth}
  \begin{center}
    \includegraphics[width=0.4\textwidth]{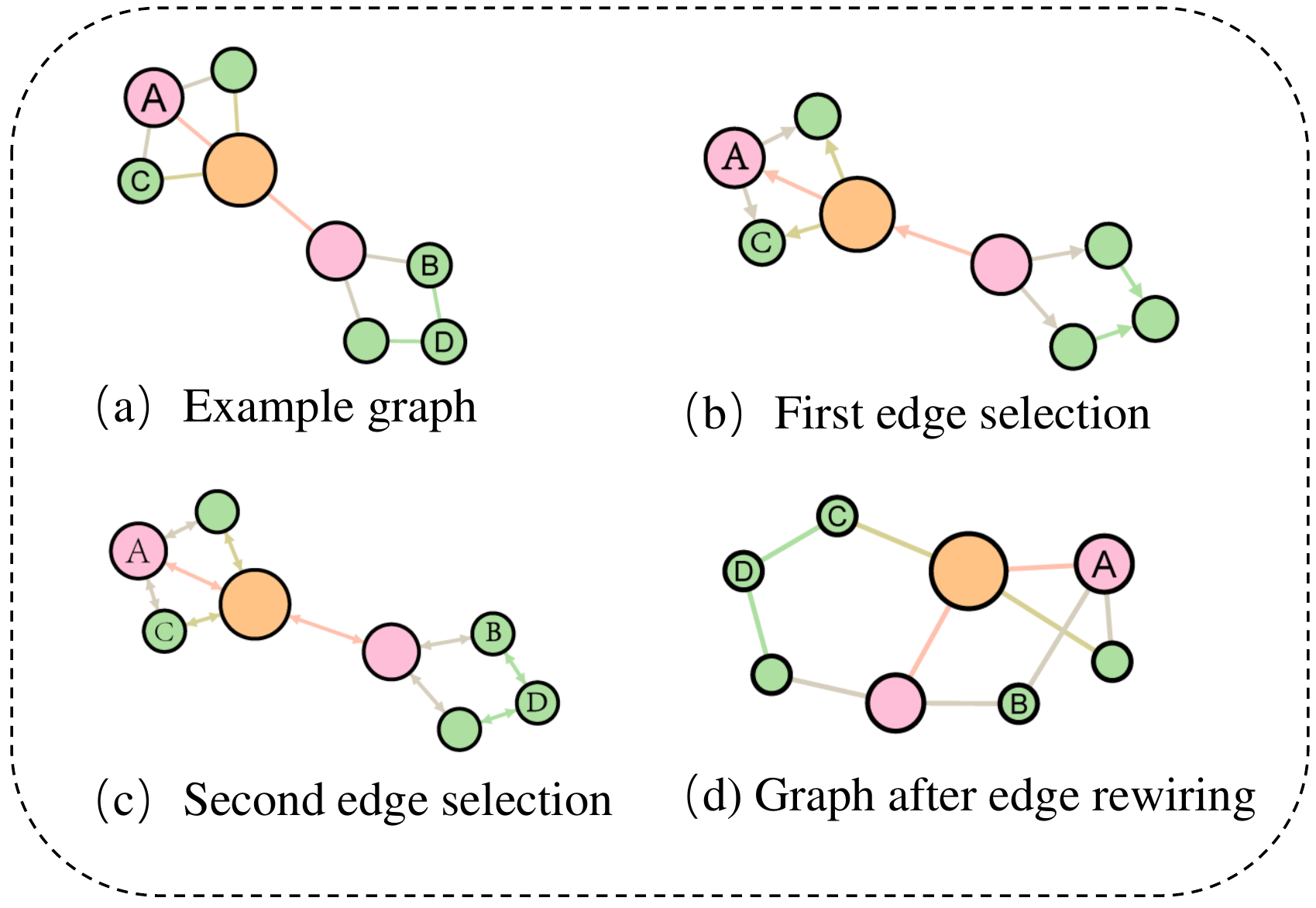}
  \end{center}
   \caption{Illustration of the edge rewiring operation with the removal of $AC$, $BD$ and the addition of $AB$, $CD$.}
     \label{fig:explain_rewire}
          \vspace{-4\intextsep}

   %
\end{wrapfigure}%

Consistent with the conventional setting in network science, two families of resilience metrics $\mathcal{R(\cdot)}$ and two examples of utility metrics $\mathcal{U(\cdot)}$ are used in our experiments, with detailed definitions deferred to Appendix~\ref{sec:appendix_obj}. 

To satisfy the constraint of preserving degree, currently, the edge rewiring operation is the default atomic action for obtaining new graph candidates $G^{'}$ from $G$. As is shown in Figure \ref{fig:explain_rewire}, at each step, two existing edges $AC$ and $BD$ are first selected. Then the edge rewiring alters the graph structure by removing $AC$ and $BD$ and adding $AB$ and $CD$, where $AB, CD, AD, BC\notin E$. Combinatorially, a total of $T$ successive steps of edge rewiring has the complexity of $O(E^{2T})$.

\section{Proposed approach: ResiNet}

\begin{figure*}[h]
     \centering
     
     \begin{subfigure}[b]{0.85\textwidth}
   
    \includegraphics[width=1\textwidth]{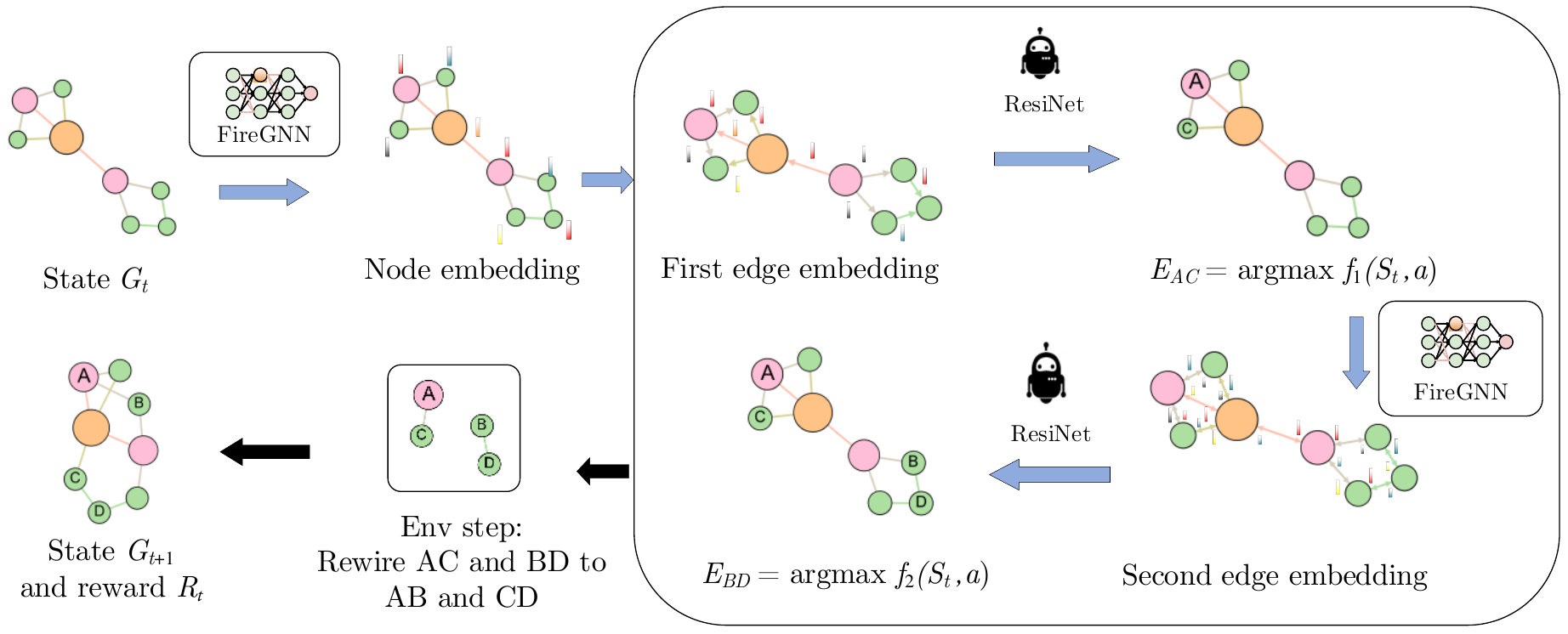}
     
      \end{subfigure}
    \caption{Overview of the architecture of ResiNet to select two  edges for rewiring.}
    \label{fig:policy_net}
   
\end{figure*}

In this section, we formulate the task of boosting network resilience as a reinforcement learning task by learning to select two edges and rewire them successively. We first present the graph resilience-aware environment design and describe our innovation FireGNN in detail. Finally, we present the graph policy network that guides the edge selection and rewiring process.

     
   


\subsection{Boosting network resilience via edge rewiring as Markov decision process}

We formulate the network resilience optimization problem via successive edge rewiring operations into the MDP framework. The Markov property denotes that the graph obtained at time step $t+1$ relies only on the graph at time step $t$ and the rewiring operation, reducing the complexity from original $O(E^{2T})$ to $O(TE^2)$.

As illustrated in Figure \ref{fig:policy_net}, the environment performs the resilience optimization in an auto-regressive step-wise way through a sequence of edge rewiring actions guided by ResiNet. Given an input graph, the agent first decides whether to terminate or not. If it chooses not to terminate, it selects one edge from the graph to remove, receives the very edge it just selected as the auto-regression signal, and then selects another edge to remove.
Four nodes of these two removed edges are re-combined, forming two new edges to be added to the graph. The optimization process repeats until the agent decides to terminate. The details of the design of the state, the action, the transition dynamics, and the reward are presented as follows.

\paragraph{State.} The fully observable state is formulated as $S_t=G_t$, where $G_t$ is the current input graph at step $t$. The detailed node feature initialization strategy is given in Appendix \ref{appendix:node_features}. 

\paragraph{Action.} ResiNet is equipped with a node permutation-invariant, variable-dimensional action space. Given a graph $G_t$, the action $a_t$ is to select two edges and the rewiring order. The agent first chooses an edge $e_1=AC$ and a direction $A\to C$. Then conditioning on the state, $e_1$, and the direction the agents chooses an edge $e_2=BD$ such that $AB, CD, AD, BC\notin E$ and a direction $B\to D$. The heads of the two edges reconnect as a new edge $AB$, and so does the tail $CD$. As $A\to C$, $B\to D$ and $C\to A$, $D\to B$ refer to the same rewiring operation, the choice of the direction of $e_1$ is randomized (this randomized bit is still an input of choosing $e_2$). This effectively reduces the size of the action space by half. In this way, The action space is the set of all feasible pairs of $(e_1, e_2)\in E^2$, with a variable size no larger than $2|E|(|E|-1)$.

\paragraph{Transition dynamics.} The formulation of the action space implies that if the agent does not terminate at step $t$, the selected action must form an edge rewiring. This edge rewiring is executed by the environment and the graph transits to the new graph.

Note that in some other work, infeasible operations are also included in the action space (to make the action space constant through the process) \cite{GCPN, trivedi2020graphopt}. 
In these work, if the operation is infeasible, it is not executed, and the state is not changed.
This reduces the sample efficiency, causes biased gradient estimations \cite{huang2020closer}, and makes the process to be prone to stuck at the state (which requires manually disabling the specific action in the next step).
ResiNet takes advantage of the state-dependent variable action space composed of only feasible operations.

\paragraph{Reward.} ResiNet aims to optimize the resilience while preserving the utility, forming a complicated and possibly unknown objective function. Despite this, by \cite{wakuta1995vector}, an MDP that maximizes a complicated objective is up to an MDP that maximizes the linear combination of resilience and utility for some coefficient factor. This fact motivates us to design the reward as the step-wise gain of such a linear combination as
\begin{equation*}
      R_t =  \alpha \cdot \mathcal{R}(G_{t+1}) + (1 - \alpha) \cdot \mathcal{U}(G_{t+1}) \, - (\alpha \cdot \mathcal{R}(G_{t}) + (1 - \alpha) \cdot \mathcal{U}(G_{t}) )  \,,  
\end{equation*}
where $\mathcal{R}(G)$ and $\mathcal{U}(G)$ are the resilience and the utility functions, respectively. The cumulative reward $\sum_{t=0}^{T-1}R_t$ up to time $T$ is then the total gain of such a linear combination.

\subsection{FireGNN}
\begin{wrapfigure}{r}{0.25\textwidth}
  \begin{center}
    \includegraphics[width=0.25\textwidth]{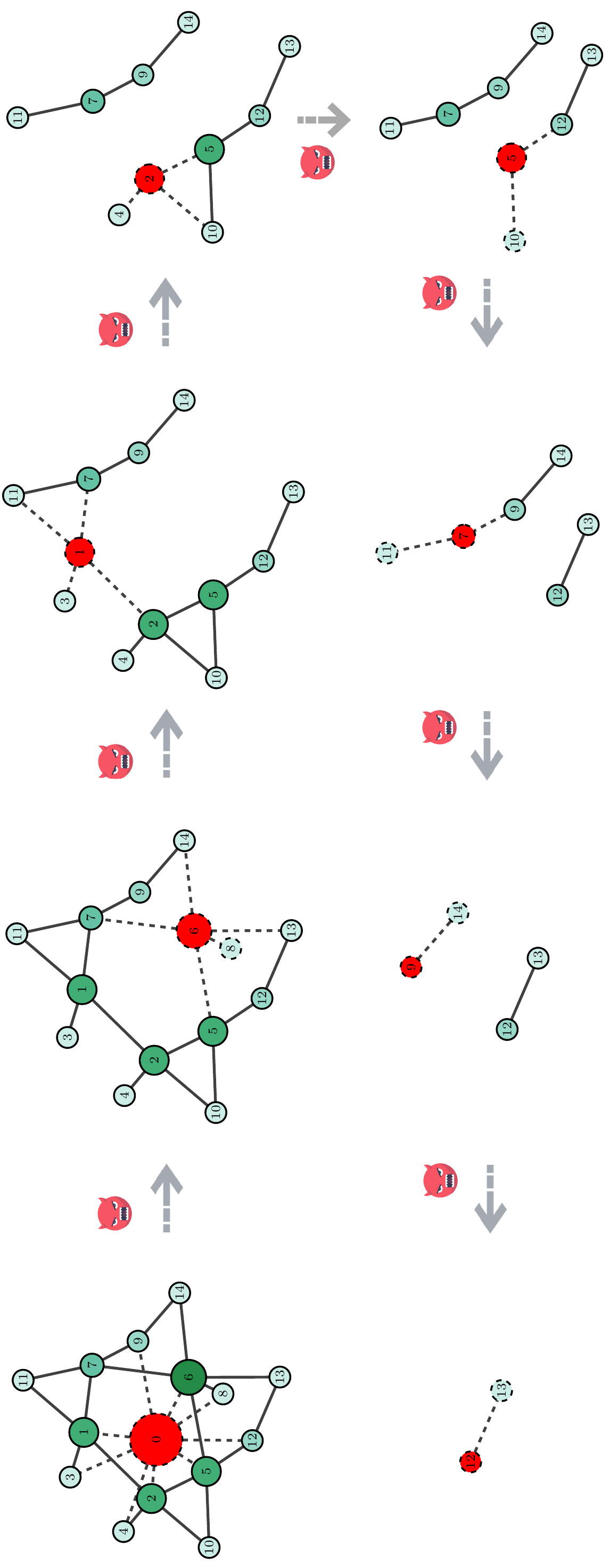}
  \end{center}
   \caption{Filtration process in FireGNN.}
     \label{fig:filtration_process}
\end{wrapfigure}%

 Motivated by graph filtration in persistent homology \cite{edelsbrunner2008persistent}, we design the \textbf{fi}lt\textbf{r}ated graph \textbf{e}nhanced \textbf{GNN} termed FireGNN to model graphs without rich features, or even with only topology. As shown in Figure \ref{fig:filtration_process}, for a given input graph $G$, FireGNN transforms $G$ from the static version to a temporal version 
consisting of a sequence of subgraphs, by repeatedly removing the node with the highest degree\footnote{Removing the node with the highest degree leads to an efficient minimal number of resultant subgraphs, comparing to the uniformly random removal of nodes.}. Observing a sequence of nested subgraphs of $G$ grants FirGNN the capability to observe how $G$ evolves towards being empty. Then FireGNN aligns and aggregates the node, edge and graph embedding from each subgraph, leading to meaningful representations in node, edge, and graph levels. 
Formally, the filtration in FireGNN is constructed as
\begin{equation*}
\begin{split}
 G^{(k-1)} &= G^{(k)} - v_k, \, \quad v_k = \argmax_{ v_i \in G^{(k)}} \text{DEGREE}(v_i) \\
        (V, \emptyset)  &= G^{(0)} \subset G^{(1)} \subset \dots  \subset G^{(N)} = G  \\
        \tilde{G} &= [G^{(0)}, G^{(1)}, \dots, G^{(N)}] \, , 
\end{split}
\end{equation*}
where $G^{(k)}$ denotes the remaining graph after removing $N-k$ nodes with  highest node degrees, $v_k$ denotes the node with highest degree in current subgraph $G^{(k)}$, DEGREE$(\cdot)$ measures the node degree, $G^{(N)}$ is the original graph, and $G^{(0)}$ contains no edge. The sequence of the nested subgraphs of $G$ is termed the filtrated graph $\tilde{G}$.

\paragraph{Node embedding.}  Regular GNN only operates on the original graph $G$ to obtain the node embedding for each node $v_i$ as
  $  h(v_i) = \phi(G^{(N)} = G)_i  \, , $
where $\phi(\cdot)$ denotes a standard GNN model.
In FireGNN, by using the top $K+1$ subgraphs in a graph filtration, the final node embedding $h(v_i)$ of $v_i$ is obtained by
\begin{equation*}\label{eq:filtratedGNN}
      h(v_i) = \text{AGG}_N \left(
      h^{(N-K)}(v_i),  \dots,  h^{(N-1)}(v_i), h^{(N)}(v_i)  \right ) \, ,
\end{equation*}
where $\text{AGG}_N(\cdot)$ denotes a node-level aggregation function, $h^{(k)}(v_i)$ is the node embedding of $i$ in the $k$-th subgraph $G^{(k)}$, and $K \in [N] $.
In practice, 
$h^{(k)}(v_i)$ is discarded when calculating $h(v_i)$ if $v_i$ is isolated or not included in $G^{(k)}$. 

\paragraph{Edge embedding.} The directed edge embedding  $h^{(k)}(e_{ij})$
of the edge from node $i$ to node $j$ in each  subgraph is obtained by combining the embeddings of the two end vertices in $G^{(k)}$ as
\begin{equation*}\label{eq:filtratedGNN_subedge_embedding}
     h^{(k)}(e_{ij})  = m_f \left (\text{AGG}_{N \to E} \left (h^{(k)}(v_i), h^{(k)}(v_j)
    \right ) \right) \, ,
\end{equation*}
where $\text{AGG}_{N \to E}(\cdot)$ denotes an aggregation function for obtaining edge embedding from two end vertices (typically chosen from $\mathtt{min}$, $\mathtt{max}$, $\mathtt{sum}$, $\mathtt{difference}$, and $\mathtt{multiplication}$).  
$m_f(\cdot)$ is a multilayer perceptron (MLP) model that ensures the consistence between the dimensions of edge embedding and graph embedding.

The final embedding of the directed edge $e_{ij}$ 
of the filtrated graph $\tilde{G}$ is given by
\begin{equation*}\label{eq:filtratedGNN_edge_embedding}
     h(e_{ij}) = \text{AGG}_E \left (
        h^{(N-K)}(e_{ij}) , \dots,  h^{(N-1)}(e_{ij}), h^{(N)}(e_{ij})  \right ) \, ,
\end{equation*}
where $\text{AGG}_E(\cdot)$ denotes an edge-level aggregation function.

\paragraph{Graph embedding.} With the node embedding $h^{(k)}(v_i)$ of each subgraph $G^{(k)}$ available, the graph embedding $h^{(k)}(G)$ of each subgraph $G^{(k)}$ is calculated by a readout functions (e.g., $\mathtt{mean}$, $\mathtt{sum}$) on all non-isolated nodes in $G^{(k)}$ as
\begin{equation*}\label{eq:filtratedGNN_subgraph_embedding}
     h^{(k)}(G) = \, \text{READOUT} \left ( {h^{(k)}(v_i)} 
    \right)  \, 
    \forall v_i \in G^{(k)} \, \text{and} \, \, d^{(k)}_i \ge 0 \,. \\
\end{equation*}
The final graph embedding of the filtrated graph $\tilde{G}$ is given by
\begin{equation*}\label{eq:filtratedGNN_graph_embedding}
    h(G) = \text{AGG}_G \left (h^{(N-K)}(G), \dots,  h^{(N-1)}(G), h^{(N)}(G)  \right ) \, ,
\end{equation*}
where $\text{AGG}_G(\cdot)$ denotes a graph-level aggregation function.

\subsection{Edge rewiring policy network}
Having presented the details of the graph resilience environment and FireGNN, in this section, we describe the policy network architecture of ResiNet in detail, which learns to select two existing edges for rewiring at each step.  At time step $t$, the policy network uses FireGNN as the graph extractor to obtain the directed edge embedding $h(e_{ij}) \in \mathbb{R}^{2|E| \times d}$ and the graph embedding $h(G)  \in \mathbb{R}^{d}$ from the filtrated graph $\tilde{G}_t$, and outputs an action $a_{t}$ representing two selected rewired edges,  leading to the new state $G_{t+1}$ and the reward $R_t$. 


To be inductive, we adapt a special autoregressive node permutation-invariant dimension-variable action space to model the selection of edges from graphs with arbitrary sizes and permutations. The detailed mechanism of obtaining the action $a_t$ based on edge embedding and graph embedding is presented as follows, further reducing the complexity from $O(TE^2)$ to $O(TE)$.

\paragraph{Auto-regressive latent edge selection.}
An edge rewiring action  $a_t$ at time step $t$ involves the prediction of the termination probability $a_t^{(0)}$ and the selection of two edges $(a_t^{(1)}$ and $a_t^{(2)})$ and the rewiring order. The action space of $a_t^{(0)}$ is binary, however, the selection of two edges imposes a huge action space in $O(|E|^2)$, which is too expensive to sample from even for a small graph. Instead of selecting two edges simultaneously, we decompose the joint action $a_t$ into $a_t = (a_t^{(0)}, a_t^{(1)}, a_t^{(2)})$, where $a_t^{(1)}$ and $a_t^{(2)}$ are two existing edges which do not share any common node (recall that $a_t^{(1)}$ and $a_t^{(2)}$ are directed edges for an undirected graph). Thus the probability of  $a_t$ is formulated as
\begin{equation*}
  \mathbb{P}(a_t|s_t) = \mathbb{P}(a_t^{(0)} |s_t) \mathbb{P}(a_t^{(1)} |s_t, a_t^{(0)}) \mathbb{P}(a_t^{(2)} |s_t,  a_t^{(0)}, a_t^{(1)}) \,. 
\end{equation*}

\paragraph{Predicting the termination probability.}
The first policy network $\pi_0(\cdot)$ takes the graph embedding as input and outputs the probability distribution of the first action that decides to terminate or not as
\begin{equation*}\label{eq:prob_ac0}
  \mathbb{P}(a_{t}^{(0)} |s_t) = \pi_0(h(G)) \,,
\end{equation*}
where $\pi_0(\cdot)$ is implemented by a two layer MLP. Then $a_t^{(0)} \sim \text{Bernoulli}(\mathbb{P}(a_{t}^{(0)} |s_t)) \in \{0, 1\}$.

\paragraph{Selecting edges.} If the signal $a_t^{(0)}$ given by the agent decides to continue to rewire, two edges are then selected in an auto-regressive way. The signal of continuing to rewire $a_t^{(0)}$ is input to the selection of two edges as a one-hot encoding vector $l_c$.
The second policy network $\pi_1(\cdot)$ takes the graph embedding and $l_c$ as input and outputs a latent vector $l_1 \in \mathbb{R}^{d}$.
The pointer network \cite{vinyals2015pointer} is used to measure the proximity between $l_1$ and each edge embedding $h(e_{ij})$ in $G$ to obtain the first edge selection probability distribution. Then, to select the second edge, the graph embedding $h(G)$ and the first selected edge embedding $ h(e_t^{(1)})$ and $l_c$ are concatenated and fed into the third policy network $\pi_2(\cdot)$. $\pi_2(\cdot)$ obtains the latent vector $l_2$ for selecting the second edge using a respective pointer network. The overall process can be formulated as
\begin{equation*}\label{eq:action:probs}
\begin{split}
 l_1 &= \pi_1([h(G), l_c])  \\
\mathbb{P}(a_{t}^{(1)} |s_t, a_t^{(0)}) &= f_1(l_1, h(e_{ij})),  \, \forall e_{ij} \in E  \\ 
 l_2 &= \pi_2([h(G),  h_{e_t^{(1)}}, l_c]) \\
\mathbb{P}(a_{t}^{(2)} |s_t, a_t^{(1)}, a_t^{(0)}) &= f_2(l_2, h(e_{ij})), \, \forall e_{ij} \in E   \, ,
\end{split}
\end{equation*}
where  $\pi_i(\cdot)$ is a two-layer MLP model, $[\cdot,\cdot]$ denotes the concatenation operator,  $ h_{e_t^{(1)}}$ is the embedding of the first selected edge at step $t$, and $f_i(\cdot)$ is a pointer network.

\section{Experiments}
\label{sec:experiments}
In this section, we demonstrate the advantages of ResiNet over existing non-learning-based and learning-based methods in achieving superior network resilience, inductively generalizing to unseen graphs, and accommodating multiple resilience and utility metrics. Moreover, we show that FireGNN can learn meaningful representations from graph data without rich features,  while current GNNs (including GNNs with stronger power than 1-WL test) fail. 
\textit{Our implementation is already open sourced\footnote{Link to code and datasets: \href{https://github.com/yangysc/ResiNet}{https://github.com/yangysc/ResiNet}}}.

\subsection{Experimental settings}

\paragraph{Datasets.}  Synthetic datasets, real EU power network \cite{zhou2005approximate} and Internet peer-to-peer networks~\cite{leskovec2007graph, ripeanu2002mapping} are used to demonstrate the performance of ResiNet in transductive and inductive settings.
The details of data generation and the statistics of the datasets are presented in Appendix \ref{appendix:datasets}. Following the conventional experimental settings in network science, the maximal node size is set to be around 1000 \cite{schneider2011mitigation}, taking into account:
1) the high complexity of selecting two edges at each step is $O(E^2)$; 2) evaluating the resilience metric is time-consuming for large graphs. 

\paragraph{Baselines.} We compare ResiNet with existing graph resilience optimization algorithms, including non-learning-based methods and learning-based algorithms. Non-learning-based methods (upper half of Table~\ref{tb:res_resi_gain_fixed_budget}) include the hill climbing (HC) \cite{schneider2011mitigation}, the greedy algorithm \cite{chan2016optimizing}, 
the simulated annealing (SA) \cite{buesser2011optimizing}, and the evolutionary algorithm (EA) \cite{zhou2014memetic}. 
Since there is no previous learning-based baseline, 
we specifically devise two counterparts based on our framework by replacing FireGNN with existing well-known powerful GNNs (DE-GNN~\cite{li2020distance} and $k$-GNN~\cite{morris2019weisfeiler}) (lower half of Table~\ref{tb:res_resi_gain_fixed_budget}). The classical GIN model is used as the backbone for FireGNN and DE-GNN \cite{xu2018how}. 
All devised counterparts and selected variants other than FireGNN cannot be successfully applied to the resilience task in an inductive fashion, when only topological structures are available.

The ResiNet's training setup 
is detailed in Appendix \ref{appendix:resinet_setup}. All algorithms are repeated for 3 random seeds using default hyper-parameters.

\paragraph{Metrics.} Various definitions of resilience and utility used for evaluating algorithms are deferred to Appendix \ref{sec:appendix_obj}.


\subsection{Comparisons to the baselines}\label{sec:baseline}

\begin{table*}
  \caption{Resilience optimization algorithm under the fixed maximal rewiring number budget of 20. 
  Entries are in the format of $X(Y)$, where 1) $X$: weighted sum of the graph connectivity-based resilience and the network efficiency improvement (in percentage); 2) $Y$: required rewiring number.
  Results are averaged over 3 runs and the best performance is in bold.} 
  \label{tb:res_resi_gain_fixed_budget}

  \centering
  \scalebox{0.72}{
  \begin{tabular}{llllllllllll}
    \toprule
    \textbf{Method}  & $\alpha$ & BA-15 & BA-50 &  BA-100  & BA-500 & BA-1000 &  EU & P2P-Gnutella05 & P2P-Gnutella09 \\
     
    \midrule
    
    \multirow{2}{4em}{HC} & $0$ & 26.8 (10)  & 30.0 (20) & 24.1 (20) & 6.4 (20) & 66.6 (20) & 19.8 (20) & 6.2 (20) & 8.4 (20)   \\
    &$0.5$ & 18.6 (11.3) & 22.1 (20) & 14.9 (20) & 5.9 (20) & 16.4 (20) & 16.3 (20) & 5.2 (20) & 7.0 (20) \\

\hline
   \multirow{2}{4em}{SA} & $0$ & 21.6 (17.3) & 11.9 (20) & 12.5 (20) & 3.8 (20) & 42.9 (20) & 14.9 (20) & 3.9 (20) & 3.7 (20) \\
    &$0.5$  & 16.8 (19.0)  & 11.4 (20) & 13.4 (20) & 4.0 (20) & 15.4 (20) & 14.0 (20) & 6.3 (20) & 4.8 (20) \\
    
 \hline
 \multirow{2}{4em}{Greedy} & $0$ & 23.5 (6)  & 48.6 (13) & 64.3 (20) & \ding{55} & \ding{55} & 0.5 (3) & \ding{55} & \ding{55} \\
    &$0.5$ & 5.3 (15)  & 34.7 (13) & 42.7 (20)  & \ding{55} & \ding{55} & 0.3 (3) & \ding{55} & \ding{55} \\

\hline

 \multirow{2}{4em}{EA} & $0$ & 8.5 (20) & 6.4 (20)  & 4.0 (20) & 8.5 (20) & \textbf{174.1} (20) & 8.2 (20) & 2.7 (20) &  0 (20) \\
    &$0.5$ & 6.4 (20)  & 4.7 (20) & 2.8 (20)  & 5.6 (20) & 18.7 (20) & 9.3 (20) & 3.7 (20) & 0.1 (20)  \\
\hline
\hline
\multirow{2}{4em}{DE-GNN-RL} & $0$ & 13.7 (2) &  0 (1)  & 0 (1)  & 1.6 (20)  & 41.7 (20) & 9.0 (20) & 2.2 (20) & 0 (1) \\
&$0.5$ &  10.9 (2) &  0 (1) & 0 (1) & 2.7 (20) & 20.1 (14) & 2.1 (20) & 0 (1) &  1.0 (20) \\

\hline
\multirow{2}{4em}{$k$-GNN-RL} & $0$ & 13.7 (2) &  0 (1)  & 0 (1)  & 0 (1)  & 8.8 (20) & 4.5 (20) & -0.2 (20) & 0 (1) \\
&$0.5$ &  6.3 (2) &  0 (1) & 0 (1) & 0 (20) & -24.9 (20) & 4.8 (20) & -0.1 (20) &  0 (1) \\

\hline
 \multirow{2}{4em}{ResiNet} & $0$ & \textbf{35.3} (6) & \textbf{61.5} (20) & \textbf{70.0} (20) & \textbf{10.2} (20) & 172.8 (20) &  \textbf{54.2} (20) & \textbf{14.0} (20) & \textbf{18.6} (20) \\
    &$0.5$ & \textbf{26.9} (20) & \textbf{53.9} (20) &  \textbf{53.1} (20) &\textbf{ 15.7} (20) & \textbf{43.7} (20) & \textbf{51.8} (20) & \textbf{12.4} (20) & \textbf{15.1} (20) \\

    \bottomrule
  \end{tabular}
  }
  
\end{table*}

In this section, we compare ResiNet to baselines in optimizing the combination of resilience and utility with weight coefficient $\alpha \in \{0, 0.5\}$. Following conventional setting, the graph connectivity-based metric is used as resilience metric \cite{schneider2011mitigation} and the global efficiency is used as utility metric \cite{latora2003economic, boccaletti2006complex}. 

Table \ref{tb:res_resi_gain_fixed_budget} records the metric gain and the required number of rewiring operations of different methods under the same rewiring budget. ResiNet outperforms all baselines consistently on all datasets. 
Note that this performance may be achieved by ResiNet under a much fewer number of rewiring operations, such as on BA-15 with $\alpha=0$.
In contrast, despite approximately searching for all possible new edges, the greedy algorithm is trapped in a local optimum (as it maximizes the one-step resilience gain) and is too expensive to optimize the resilience of a network with more than 300 nodes.
For SA, the initial temperature and the temperature decay rate need to be carefully tuned for each network. EA performs suboptimally with a limited rewiring budget due to the numerous rewiring operations required in the internal process (e.g., the crossover operator). Learning-based methods (DE-GNN-RL and $k$-GNN-RL) using existing GNNs coupled with distance encoding cannot learn effectively compared to ResiNet, supporting our claim about the effectiveness of FireGNN on graphs without rich features. 

All baselines are compared under the same rewiring budget of 20 since each edge rewiring introduces economic costs. We record the performance and speed of each algorithm for a maximal rewiring budget of 200 in Table \ref{tb:res_resi_gain_budget_200}. Solving the resilience task under such a large rewiring budget (200) will not be applicable in practice due to the high cost of many rewiring operations in general, and as such this result is only for the completeness of the presentation.

\subsection{Ablation study of FireGNN}


\begin{figure*}
    \centering
   \begin{subfigure}{0.41\textwidth}  
         \centering
    \includegraphics[width=0.95\textwidth]{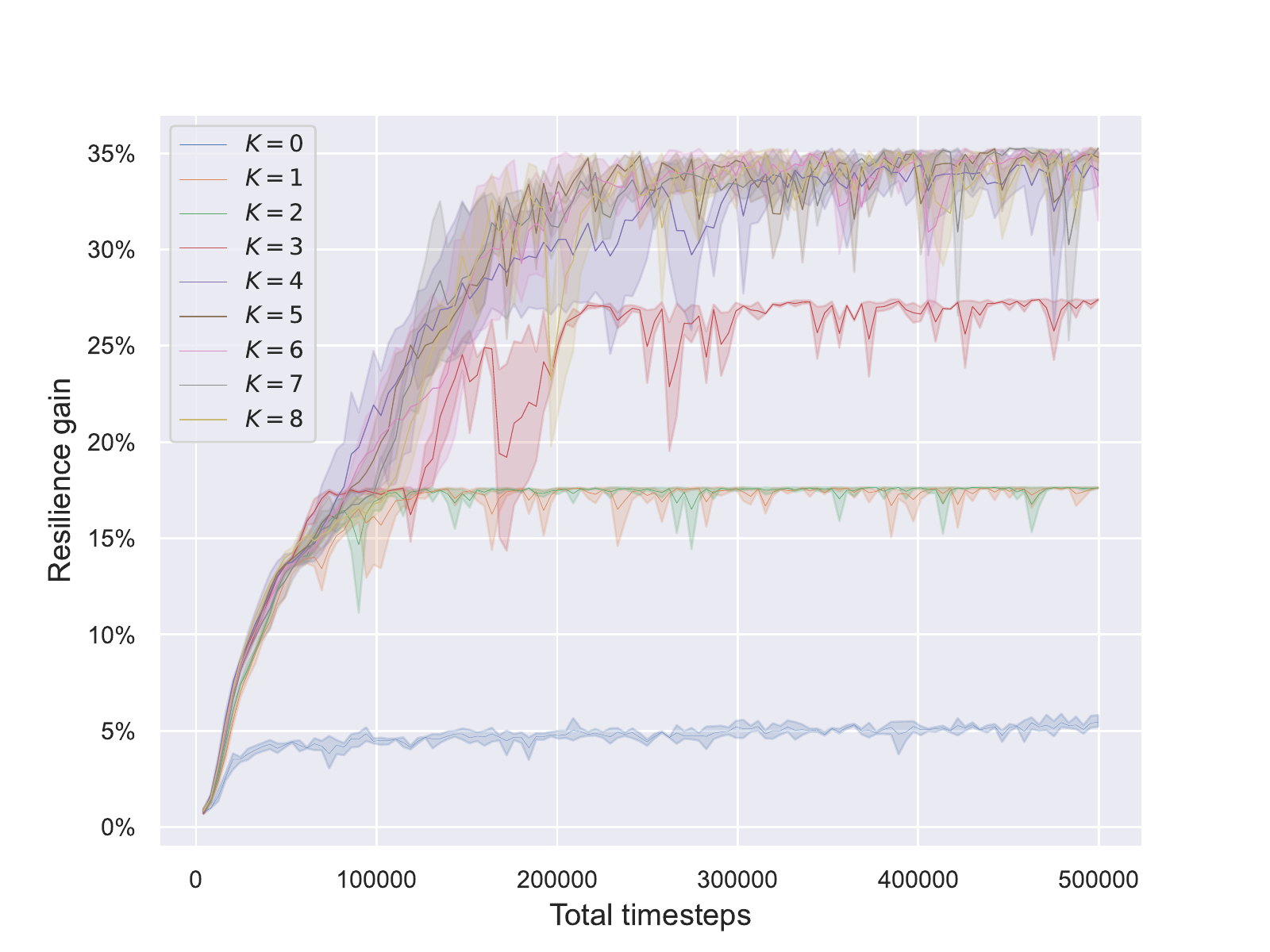}
    \caption{}
    \label{fig:ablation_filtration_order}
     \end{subfigure}
     \begin{subfigure}{0.41\textwidth}
         \centering
       \includegraphics[width=0.95\textwidth]{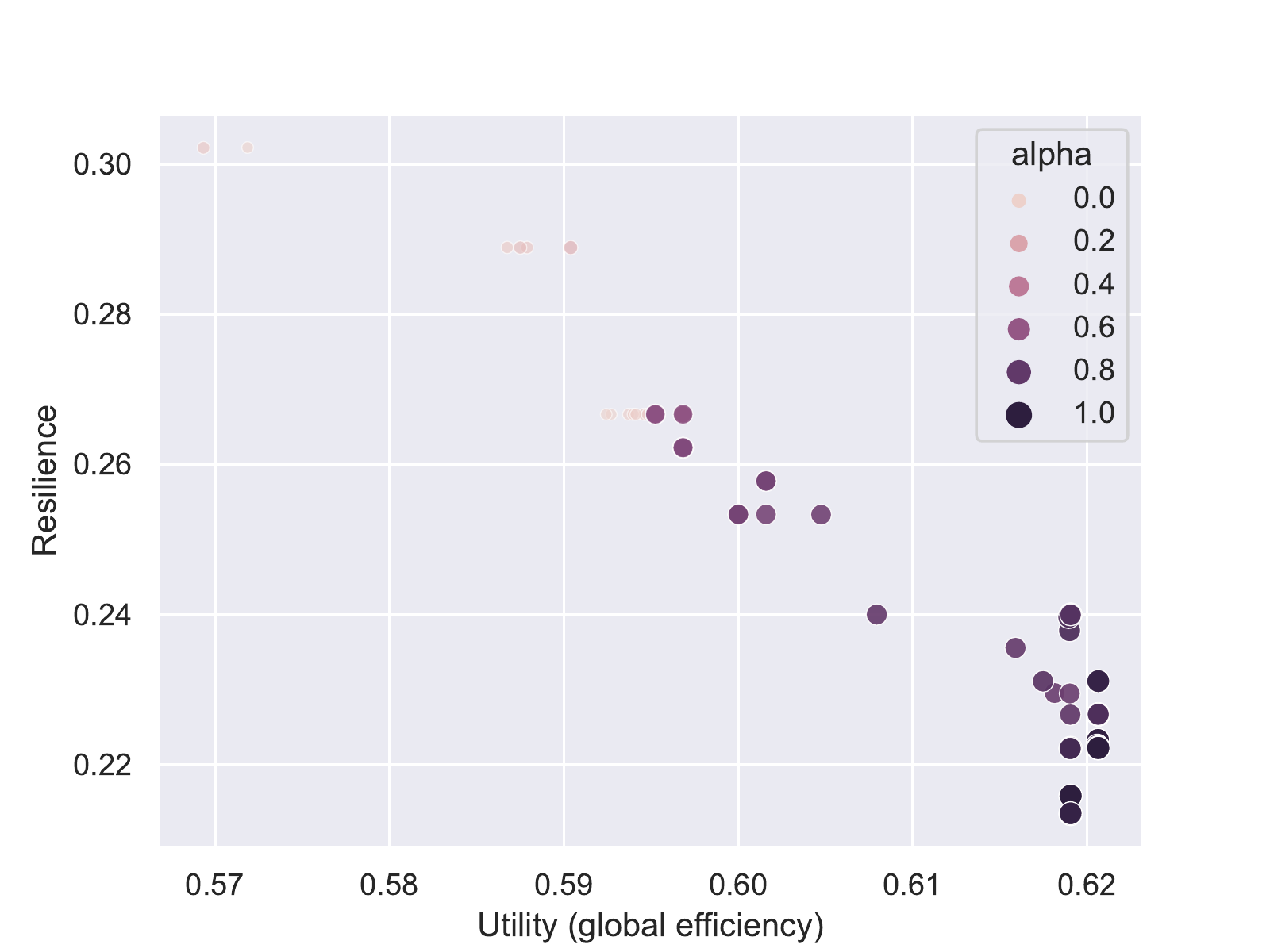}        
       \caption{}
         \label{fig:pareto}
     \end{subfigure}
     
\caption{Experimental results on ablation study and generalization of ResiNet with (a) effect of the filtration order $K$  on FireGNN, (b) pareto points obtained by ResiNet of balancing various combinations of resilience and utility. Results are averaged over 3 runs on the BA-15 dataset.}
\label{fig:test}
\end{figure*}

In FireGNN, the filtration order $K \in \{0,\dots, N-1\}$ determines the total number of subgraphs involved in calculating the final node embedding, edge embedding, and graph embedding, where $K=0$ means that only the original graph is used and $K=N-1$ means that all $N$ subgraphs are used. We run a grid search on BA-15 to explore the effect of the filtration order $K$ on ResiNet. As shown in Figure \ref{fig:ablation_filtration_order}, without FireGNN, ResiNet only achieves a minor gain of around 5\%. The performance improved significantly with $K > 0$ and ResiNet obtained the optimal resilience gain of about 35\% on BA-15 when $K \geq 5$. We only report the performance when $K \leq 8$ since BA-15 loses all connections when losing more than $8$ critical nodes.

FireGNN degenerates to existing GNNs when the filtration order $K$ is 0. Aside from the BA-15 dataset, the comparisons between ResiNet and other learning-based methods on other large datasets further validate the effectiveness of FireGNN.
Table \ref{tb:res_resi_gain_fixed_budget} shows
that without FireGNN (replaced by DE-GNN or $k$-GNN),  it is generally challenging for ResiNet to find a positive gain and ResiNet cannot learn to select the correct edges with the incorrect learned edge embeddings.

\begin{figure*}
    \centering
   \begin{subfigure}{0.32\textwidth}  
         \centering
    \includegraphics[width=\textwidth]{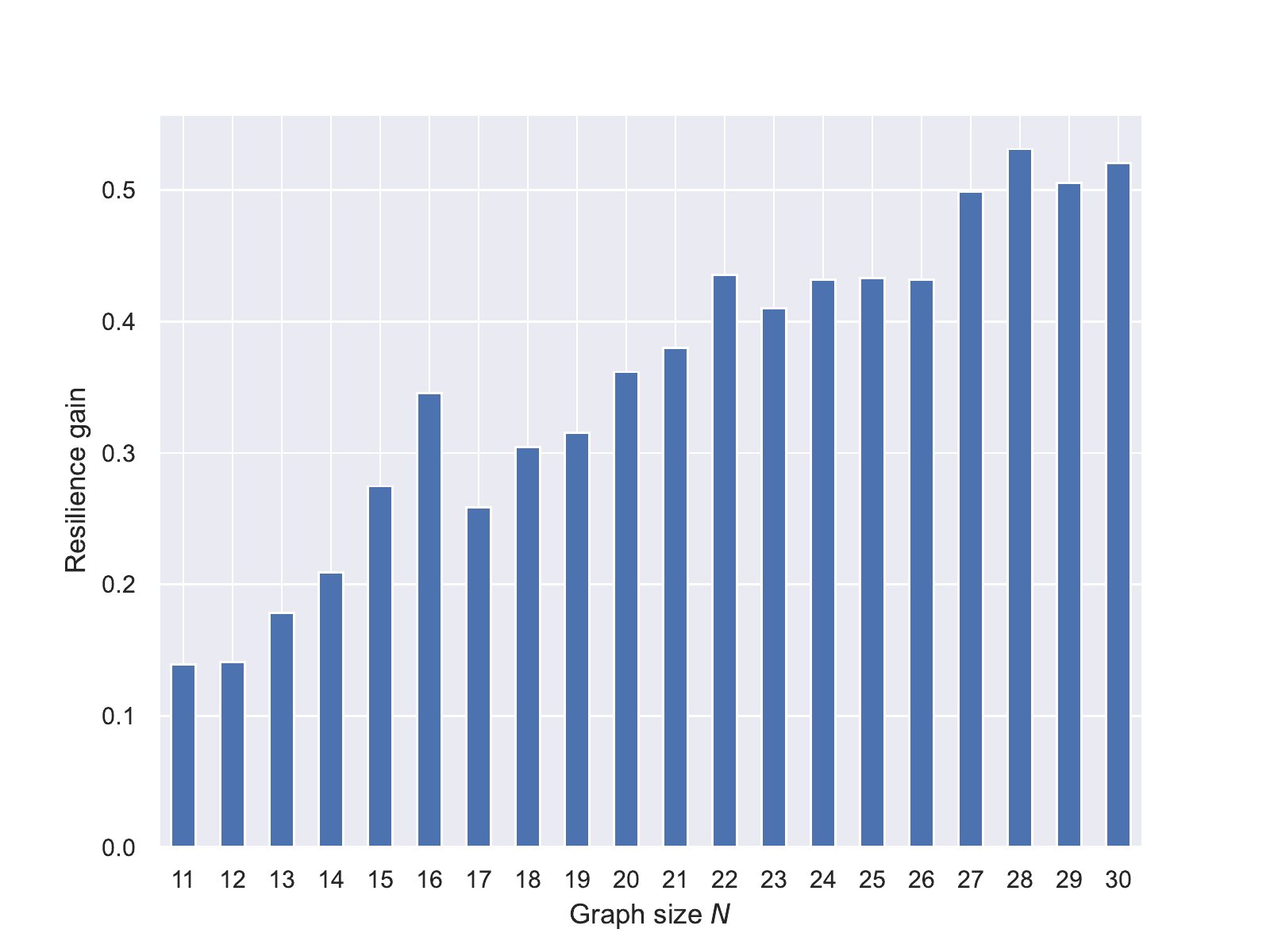}
    \caption{Inductivity on resilience}
     \end{subfigure}
      \hfill
     \begin{subfigure}{0.32\textwidth}
         \centering
       \includegraphics[width=\textwidth]{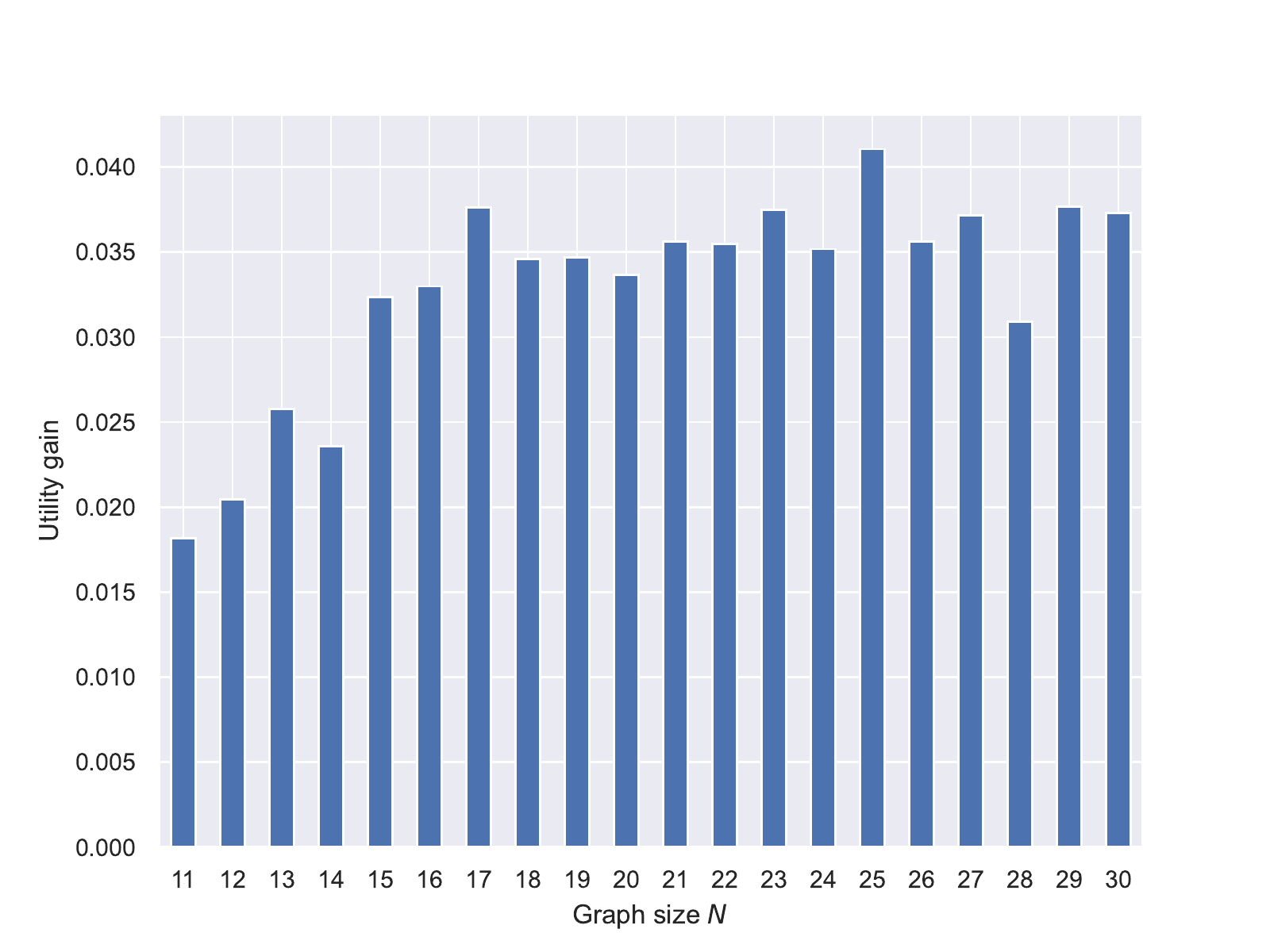}
       \caption{Inductivity on utility}
     \end{subfigure}
       \hfill
      \begin{subfigure}{0.32\textwidth}  
         \centering
    \includegraphics[width=\textwidth]{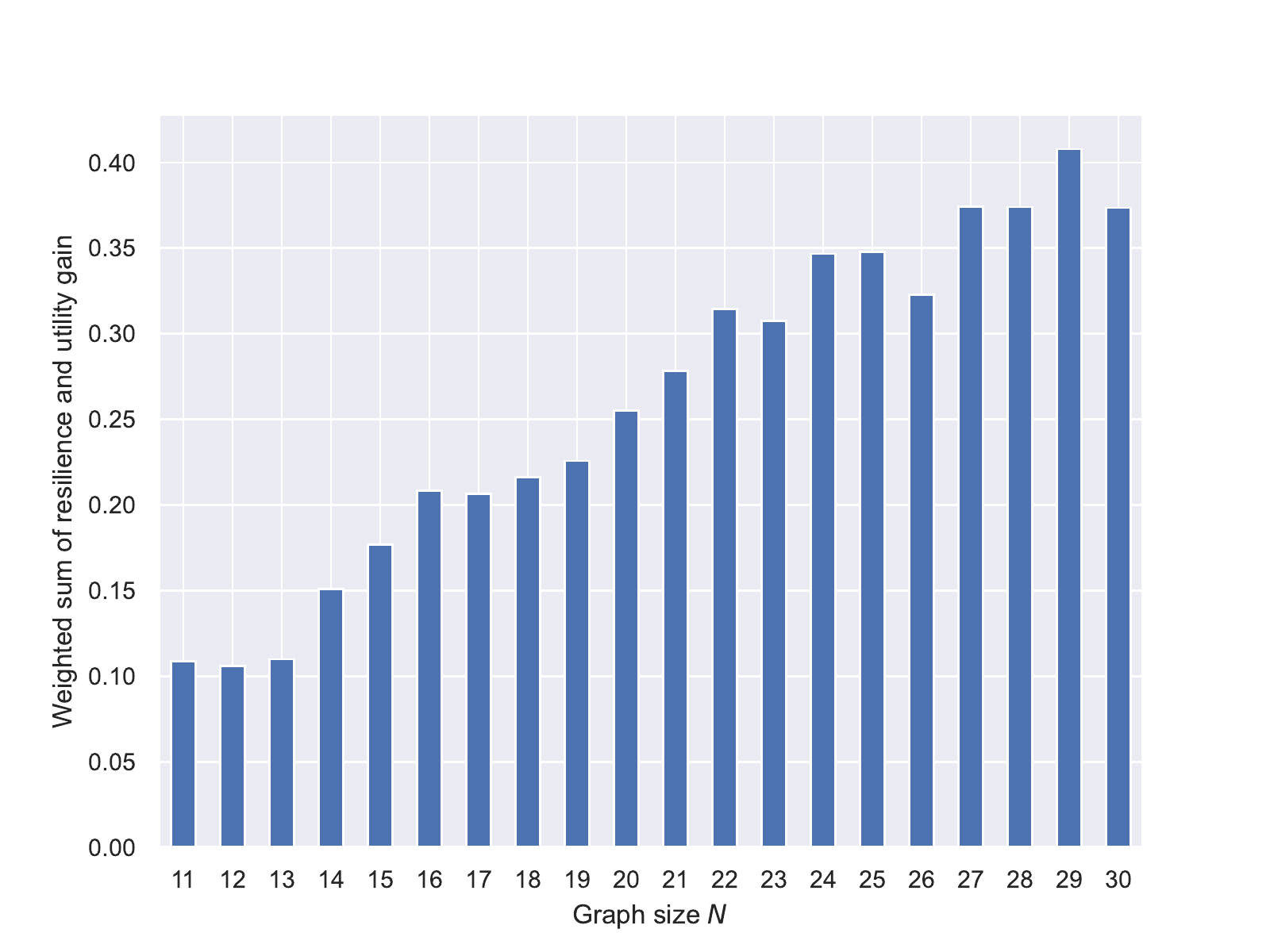}
    \caption{Inductivity on both metrics}
     \end{subfigure}
     
\caption{ The inductive ability of ResiNet on the test dataset (BA-10-30) when optimizing (a) network resilience, (b) network utility, and (c) their combination.}
\label{fig:inductivity_ba_30}
\end{figure*}

Experimental results in Table \ref{tb:res_resi_gain_fixed_budget} validate that FireGNN is critical for the network resilience task without rich features, by exploiting node information of each subgraph in a filtration process. Empirically, we found that ResiNet avoids the following problematic phenomena of existing GNNs during the training.
    As each rewiring only changes a graph by 4 edges, the graph embedding and the edge embedding may not vary significantly between two consecutive steps. Thus, existing GNNs fail to provide enough information for the RL agent to make correct edge selections. For example, we empirically found that with existing GNNs the RL agent can be stuck into an action loop, where after the rewiring of $AC$ and $BD$ to obtain $AB$ and $CD$ for $G_t$ at step $t$, the agent may choose to reverse the rewiring (rewire $AB$ and $CD$) for $G_{t+1}$ at step $t+1$, returning to $G_t$ and then trapped in an infinite loop between $G_t$ and $G_{t+1}$.

\subsection{Generalization}

    

In this section, we conduct extensive experiments to show that ResiNet generalizes to unseen graphs, different utility and resilience metrics. 

To demonstrate the inductivity of ResiNet, we first train ResiNet on two different datasets with the data setting listed in Table \ref{tb:dataset}, and then evaluate its performance on an individual test dataset. The test dataset is not observed to ResiNet during the training process and fine-tuning is not allowed. We report the averaged resilience gain for the graphs of the same size for each dataset. The performance of ResiNet on BA-10-30 is shown in Figure \ref{fig:inductivity_ba_30} and the results of other datasets are deferred to Figure~\ref{fig:inductivity_ba_mixed} in  Appendix~\ref{appendix:extra_results}. Figure \ref{fig:inductivity_ba_30} shows a nearly linear improvement of resilience with the increase of graph size, which is also consistent with the results in the transductive setting that larger graphs usually have a larger room to improve.

To demonstrate that ResiNet can learn from networks to accommodate different utility and resilience metrics, we conduct experiments based on the BA-15 using multiple resilience and utility metrics. 
The Pareto points shown in Figure \ref{fig:pareto} denote the optimum under different objectives on BA-15, implying that ResiNet can obtain the approximate Pareto frontier. Surprisingly, the initial gain of resilience (from around 0.21 to around 0.24) is obtained without loss of the utility, which incentivizes almost every network to conduct such optimization to some extent when feasible.
More results are included in Appendix~\ref{sec:res_multiobjs} and the optimized network structures are visualized in Figure \ref{fig:casestudy} and Figure~\ref{fig:results_EU}.

\section{Conclusion}\label{sec:conclusion}
In this work, we propose a general learning-based framework, ResiNet, for the discovery of resilient network topologies with minimal changes to the graph structure. ResiNet is the \textit{first} inductive framework that formulates the task of boosting network resilience as an MDP of successive edge rewiring operations.
Our technical innovation, FireGNN, as the graph feature extractor in ResiNet, is motivated by persistent homology. FireGNN alleviates the insufficiency of current GNNs (including GNNs more powerful than 1-WL test) on modeling graphs without rich features. FireGNN can learn meaningful representations on the resilience task to provide sufficient gradients for training the RL agent while current GNNs fail.
Our framework is practically feasible as it preserves the utility of the networks when boosting resilience.
Both ResiNet and FireGNN are potentially general enough to be applied to solve various graph problems without rich features.

\textbf{Limitation}. Similar to existing baselines, the exact objective oracle in ResiNet is time-consuming for evaluation on giant graphs. Future work should investigate how to combine FireGNN with techniques for handling out-of-distribution data to learn a reliable neural-version oracle for fast evaluation.

\textbf{Negative social impact}. The authors do not foresee the negative social impact of this work.




{
\small


\begin{thebibliography}{10}

\bibitem{schneider2011mitigation}
Christian~M Schneider, Andr{\'e}~A Moreira, Jos{\'e}~S Andrade, Shlomo Havlin,
  and Hans~J Herrmann.
\newblock Mitigation of malicious attacks on networks.
\newblock {\em Proceedings of the National Academy of Sciences},
  108(10):3838--3841, 2011.

\bibitem{chan2016optimizing}
Hau Chan and Leman Akoglu.
\newblock Optimizing network robustness by edge rewiring: a general framework.
\newblock {\em Data Mining and Knowledge Discovery}, 30(5):1395--1425, 2016.

\bibitem{rong2018heuristic}
Lei Rong and Jing Liu.
\newblock A heuristic algorithm for enhancing the robustness of scale-free
  networks based on edge classification.
\newblock {\em Physica A: Statistical Mechanics and its Applications},
  503:503--515, 2018.

\bibitem{DBLP:conf/kdd/0001WDWT21}
Yao Ma, Suhang Wang, Tyler Derr, Lingfei Wu, and Jiliang Tang.
\newblock Graph adversarial attack via rewiring.
\newblock In {\em {KDD}}, pages 1161--1169. {ACM}, 2021.

\bibitem{jaume20202}
Daniel~A Jaume, Adri{\'a}n Pastine, and Victor~Nicolas Schv{\"o}llner.
\newblock 2-switch: transition and stability on graphs and forests.
\newblock {\em arXiv preprint arXiv:2004.11164}, 2020.

\bibitem{zhou2014memetic}
Mingxing Zhou and Jing Liu.
\newblock A memetic algorithm for enhancing the robustness of scale-free
  networks against malicious attacks.
\newblock {\em Physica A: Statistical Mechanics and its Applications},
  410:131--143, 2014.

\bibitem{yaziciouglu2015formation}
A~Yasin Yaz{\i}c{\i}o{\u{g}}lu, Magnus Egerstedt, and Jeff~S Shamma.
\newblock Formation of robust multi-agent networks through self-organizing
  random regular graphs.
\newblock {\em IEEE Transactions on Network Science and Engineering},
  2(4):139--151, 2015.

\bibitem{mosk2008maximum}
Damon Mosk-Aoyama.
\newblock Maximum algebraic connectivity augmentation is {NP}-hard.
\newblock {\em Operations Research Letters}, 36(6):677--679, 2008.

\bibitem{guided_tree_search}
Zhuwen Li, Qifeng Chen, and Vladlen Koltun.
\newblock Combinatorial optimization with graph convolutional networks and
  guided tree search.
\newblock In {\em Advances in Neural Information Processing Systems}, 2018.

\bibitem{effcient_TSP}
Chaitanya~K. Joshi, Thomas Laurent, and Xavier Bresson.
\newblock An efficient graph convolutional network technique for the travelling
  salesman problem.
\newblock {\em CoRR}, abs/1906.01227, 2019.

\bibitem{TSP_pretrain}
Zhang{-}Hua Fu, Kai{-}Bin Qiu, and Hongyuan Zha.
\newblock Generalize a small pre-trained model to arbitrarily large {TSP}
  instances.
\newblock In {\em Proceedings of the AAAI Conference on Artificial
  Intelligence}, 2020.

\bibitem{CO_on_graph}
Elias~B. Khalil, Hanjun Dai, Yuyu Zhang, Bistra Dilkina, and Le~Song.
\newblock Learning combinatorial optimization algorithms over graphs.
\newblock In {\em Advances in Neural Information Processing Systems}, 2017.

\bibitem{rl_for_vrp}
MohammadReza Nazari, Afshin Oroojlooy, Lawrence~V. Snyder, and Martin
  Tak{\'{a}}c.
\newblock Reinforcement learning for solving the vehicle routing problem.
\newblock In {\em Advances in Neural Information Processing Systems}, 2018.

\bibitem{rl_for_vrp_with_attention}
Bo~Peng, Jiahai Wang, and Zizhen Zhang.
\newblock A deep reinforcement learning algorithm using dynamic attention model
  for vehicle routing problems.
\newblock {\em CoRR}, abs/2002.03282, 2020.

\bibitem{online_vrp}
James Jian~Qiao Yu, Wen Yu, and Jiatao Gu.
\newblock Online vehicle routing with neural combinatorial optimization and
  deep reinforcement learning.
\newblock {\em {IEEE} Transactions on Intelligent Transportation Systems},
  20(10):3806--3817, 2019.

\bibitem{boffa2022neural}
Matteo Boffa, Zied Ben{-}Houidi, Jonatan Krolikowski, and Dario Rossi.
\newblock Neural combinatorial optimization beyond the {TSP:} existing
  architectures under-represent graph structure.
\newblock In {\em Proceedings of the AAAI Conference on Artificial
  Intelligence}, 2022.

\bibitem{li2020distance}
Pan Li, Yanbang Wang, Hongwei Wang, and Jure Leskovec.
\newblock Distance encoding: Design provably more powerful neural networks for
  graph representation learning.
\newblock In {\em Advances in Neural Information Processing Systems}, 2020.

\bibitem{edelsbrunner2008persistent}
Herbert Edelsbrunner and John Harer.
\newblock Persistent homology-{A} survey.
\newblock {\em Contemporary Mathematics}, 453:257--282, 2008.

\bibitem{aktas2019persistence}
Mehmet~E Aktas, Esra Akbas, and Ahmed El~Fatmaoui.
\newblock Persistence homology of networks: methods and applications.
\newblock {\em Applied Network Science}, 4(1):1--28, 2019.

\bibitem{hofer2020graph}
Christoph Hofer, Florian Graf, Bastian Rieck, Marc Niethammer, and Roland
  Kwitt.
\newblock Graph filtration learning.
\newblock In {\em International Conference on Machine Learning}, 2020.

\bibitem{fan2020finding}
Changjun Fan, Li~Zeng, Yizhou Sun, and Yang-Yu Liu.
\newblock Finding key players in complex networks through deep reinforcement
  learning.
\newblock {\em Nature Machine Intelligence}, pages 1--8, 2020.

\bibitem{zhao2021}
Kangfei Zhao, Zhiwei Zhang, Yu~Rong, Jeffrey~Xu Yu, and Junzhou Huang.
\newblock Finding critical users in social communities via graph convolutions.
\newblock {\em IEEE Transactions on Knowledge and Data Engineering}, pages
  1--1, 2021.

\bibitem{fan2017}
Fan Zhang, Ying Zhang, Lu~Qin, Wenjie Zhang, and Xuemin Lin.
\newblock Finding critical users for social network engagement: The collapsed
  k-core problem.
\newblock In {\em Proceedings of the AAAI Conference on Artificial
  Intelligence}, 2017.

\bibitem{holme2002attack}
Petter Holme, Beom~Jun Kim, Chang~No Yoon, and Seung~Kee Han.
\newblock Attack vulnerability of complex networks.
\newblock {\em Physical Review E}, 65(5):056109, 2002.

\bibitem{iyer2013attack}
Swami Iyer, Timothy Killingback, Bala Sundaram, and Zhen Wang.
\newblock Attack robustness and centrality of complex networks.
\newblock {\em PloS One}, 8(4):e59613, 2013.

\bibitem{grassia2021machine}
Marco Grassia, Manlio De~Domenico, and Giuseppe Mangioni.
\newblock Machine learning dismantling and early-warning signals of
  disintegration in complex systems.
\newblock {\em Nature Communications}, 12(1):1--10, 2021.

\bibitem{KCM}
Sourav Medya, Tianyi Ma, Arlei Silva, and Ambuj Singh.
\newblock A game theoretic approach for k-core minimization.
\newblock In {\em Proceedings of the International Conference on Autonomous
  Agents and MultiAgent Systems}, 2020.

\bibitem{li2019maximizing}
Wenguo Li, Yong Li, Yi~Tan, Yijia Cao, Chun Chen, Ye~Cai, Kwang~Y Lee, and
  Michael Pecht.
\newblock Maximizing network resilience against malicious attacks.
\newblock {\em Scientific Reports}, 9(1):1--9, 2019.

\bibitem{carchiolo2019network}
Vincenza Carchiolo, Marco Grassia, Alessandro Longheu, Michele Malgeri, and
  Giuseppe Mangioni.
\newblock Network robustness improvement via long-range links.
\newblock {\em Computational Social Networks}, 6(1):1--16, 2019.

\bibitem{wang2014improving}
Xiangrong Wang, Evangelos Pournaras, Robert~E Kooij, and Piet Van~Mieghem.
\newblock Improving robustness of complex networks via the effective graph
  resistance.
\newblock {\em The European Physical Journal B}, 87(9):1--12, 2014.

\bibitem{buesser2011optimizing}
Pierre Buesser, Fabio Daolio, and Marco Tomassini.
\newblock Optimizing the robustness of scale-free networks with simulated
  annealing.
\newblock In {\em International Conference on Adaptive and Natural Computing
  Algorithms}, pages 167--176. Springer, 2011.

\bibitem{kipf2016semi}
Thomas~N Kipf and Max Welling.
\newblock Semi-supervised classification with graph convolutional networks.
\newblock In {\em International Conference on Learning Representations}, 2017.

\bibitem{hamilton2017inductive}
Will Hamilton, Zhitao Ying, and Jure Leskovec.
\newblock Inductive representation learning on large graphs.
\newblock In {\em Advances in Neural Information Processing Systems}, 2017.

\bibitem{DBLP:conf/iclr/VelickovicCCRLB18}
Petar Velickovic, Guillem Cucurull, Arantxa Casanova, Adriana Romero, Pietro
  Li{\`{o}}, and Yoshua Bengio.
\newblock Graph attention networks.
\newblock In {\em International Conference on Learning Representations}, 2018.

\bibitem{li2020deepergcn}
Guohao Li, Chenxin Xiong, Ali Thabet, and Bernard Ghanem.
\newblock Deeper{GCN}: All you need to train deeper {GCN}s.
\newblock {\em arXiv preprint arXiv:2006.07739}, 2020.

\bibitem{xu2018how}
Keyulu Xu, Weihua Hu, Jure Leskovec, and Stefanie Jegelka.
\newblock How powerful are graph neural networks?
\newblock In {\em International Conference on Learning Representations}, 2019.

\bibitem{bodnar2021weisfeiler}
Cristian Bodnar, Fabrizio Frasca, Nina Otter, Yuguang Wang, Pietro Lio, Guido~F
  Montufar, and Michael Bronstein.
\newblock Weisfeiler and lehman go cellular: {CW} networks.
\newblock In {\em Advances in Neural Information Processing Systems}, 2021.

\bibitem{hudson2021graph}
Benjamin Hudson, Qingbiao Li, Matthew Malencia, and Amanda Prorok.
\newblock Graph neural network guided local search for the traveling
  salesperson problem.
\newblock In {\em International Conference on Learning Representations}, 2022.

\bibitem{yu2021deep}
Tianshu Yu, Runzhong Wang, Junchi Yan, and Baoxin Li.
\newblock Deep latent graph matching.
\newblock In {\em International Conference on Machine Learning}, 2021.

\bibitem{dai2018adversarial}
Hanjun Dai, Hui Li, Tian Tian, Xin Huang, Lin Wang, Jun Zhu, and Le~Song.
\newblock Adversarial attack on graph structured data.
\newblock In {\em International Conference on Machine Learning}, 2018.

\bibitem{morris2019weisfeiler}
Christopher Morris, Martin Ritzert, Matthias Fey, William~L Hamilton, Jan~Eric
  Lenssen, Gaurav Rattan, and Martin Grohe.
\newblock Weisfeiler and {Leman} go neural: Higher-order graph neural networks.
\newblock In {\em Proceedings of the AAAI Conference on Artificial
  Intelligence}, 2019.

\bibitem{you2021identity}
Jiaxuan You, Jonathan Gomes{-}Selman, Rex Ying, and Jure Leskovec.
\newblock Identity-aware graph neural networks.
\newblock In {\em Proceedings of the AAAI Conference on Artificial
  Intelligence}, 2021.

\bibitem{GCPN}
Jiaxuan You, Bowen Liu, Zhitao Ying, Vijay~S. Pande, and Jure Leskovec.
\newblock Graph convolutional policy network for goal-directed molecular graph
  generation.
\newblock In {\em Advances in Neural Information Processing Systems}, 2018.

\bibitem{trivedi2020graphopt}
Rakshit Trivedi, Jiachen Yang, and Hongyuan Zha.
\newblock Graph{O}pt: Learning optimization models of graph formation.
\newblock In {\em International Conference on Machine Learning}, pages
  9603--9613. PMLR, 2020.

\bibitem{huang2020closer}
Shengyi Huang and Santiago Onta{\~{n}}{\'{o}}n.
\newblock A closer look at invalid action masking in policy gradient
  algorithms.
\newblock {\em CoRR}, abs/2006.14171, 2020.

\bibitem{wakuta1995vector}
Kazuyoshi Wakuta.
\newblock Vector-valued markov decision processes and the systems of linear
  inequalities.
\newblock {\em Stochastic Processes and Their Applications}, 56(1):159--169,
  1995.

\bibitem{vinyals2015pointer}
Oriol Vinyals, Meire Fortunato, and Navdeep Jaitly.
\newblock Pointer networks.
\newblock In {\em Advances in Neural Information Processing Systems}, 2015.

\bibitem{zhou2005approximate}
Qiong Zhou and Janusz~W Bialek.
\newblock Approximate model of european interconnected system as a benchmark
  system to study effects of cross-border trades.
\newblock {\em IEEE Transactions on Power Systems}, 20(2):782--788, 2005.

\bibitem{leskovec2007graph}
Jure Leskovec, Jon Kleinberg, and Christos Faloutsos.
\newblock Graph evolution: Densification and shrinking diameters.
\newblock {\em ACM transactions on Knowledge Discovery from Data (TKDD)},
  1(1):2--es, 2007.

\bibitem{ripeanu2002mapping}
Matei Ripeanu, Ian Foster, and Adriana Iamnitchi.
\newblock Mapping the gnutella network: Properties of large-scale peer-to-peer
  systems and implications for system design.
\newblock {\em arXiv preprint cs/0209028}, 2002.

\bibitem{latora2003economic}
Vito Latora and Massimo Marchiori.
\newblock Economic small-world behavior in weighted networks.
\newblock {\em The European Physical Journal B-Condensed Matter and Complex
  Systems}, 32(2):249--263, 2003.

\bibitem{boccaletti2006complex}
Stefano Boccaletti, Vito Latora, Yamir Moreno, Martin Chavez, and D-U Hwang.
\newblock Complex networks: Structure and dynamics.
\newblock {\em Physics Reports}, 424(4-5):175--308, 2006.

\bibitem{pretraining_strategy}
Weihua Hu, Bowen Liu, Joseph Gomes, Marinka Zitnik, Percy Liang, Vijay~S.
  Pande, and Jure Leskovec.
\newblock Strategies for pre-training graph neural networks.
\newblock In {\em International Conference on Learning Representations}, 2020.

\bibitem{GCC}
Jiezhong Qiu, Qibin Chen, Yuxiao Dong, Jing Zhang, Hongxia Yang, Ming Ding,
  Kuansan Wang, and Jie Tang.
\newblock {GCC:} graph contrastive coding for graph neural network
  pre-training.
\newblock In {\em {KDD} '20: The 26th {ACM} {SIGKDD} Conference on Knowledge
  Discovery and Data Mining}, 2020.

\bibitem{DGI}
Petar Velickovic, William Fedus, William~L. Hamilton, Pietro Li{\`{o}}, Yoshua
  Bengio, and R.~Devon Hjelm.
\newblock Deep graph infomax.
\newblock In {\em International Conference on Learning Representations}, 2019.

\bibitem{GCA}
Yuning You, Tianlong Chen, Yongduo Sui, Ting Chen, Zhangyang Wang, and Yang
  Shen.
\newblock Graph contrastive learning with augmentations.
\newblock In {\em Advances in Neural Information Processing Systems}, 2020.

\bibitem{pmlr-v119-hassani20a}
Kaveh Hassani and Amir~Hosein Khasahmadi.
\newblock Contrastive multi-view representation learning on graphs.
\newblock In {\em Proceedings of the International Conference on Machine
  Learning}, 2020.

\bibitem{jin2020graph}
Wei Jin, Yao Ma, Xiaorui Liu, Xianfeng Tang, Suhang Wang, and Jiliang Tang.
\newblock Graph structure learning for robust graph neural networks.
\newblock In {\em Proceedings of the 26th ACM SIGKDD International Conference
  on Knowledge Discovery \& Data Mining}, 2020.

\bibitem{GT-GAN}
Xiaojie Guo, Lingfei Wu, and Liang Zhao.
\newblock Deep graph translation.
\newblock {\em CoRR}, abs/1805.09980, 2018.

\bibitem{bengio2021flow}
Emmanuel Bengio, Moksh Jain, Maksym Korablyov, Doina Precup, and Yoshua Bengio.
\newblock Flow network based generative models for non-iterative diverse
  candidate generation.
\newblock {\em Advances in Neural Information Processing Systems}, 34, 2021.

\bibitem{latora2001efficient}
Vito Latora and Massimo Marchiori.
\newblock Efficient behavior of small-world networks.
\newblock {\em Physical Review Letters}, 87(19):198701, 2001.

\bibitem{albert2002statistical}
R{\'e}ka Albert and Albert-L{\'a}szl{\'o} Barab{\'a}si.
\newblock Statistical mechanics of complex networks.
\newblock {\em Reviews of Modern Physics}, 74(1):47, 2002.

\bibitem{bollobas2004robustness}
B{\'e}la Bollob{\'a}s and Oliver Riordan.
\newblock Robustness and vulnerability of scale-free random graphs.
\newblock {\em Internet Mathematics}, 1(1):1--35, 2004.

\bibitem{leskovec2006sampling}
Jure Leskovec and Christos Faloutsos.
\newblock Sampling from large graphs.
\newblock In {\em Proceedings of the 12th ACM SIGKDD International conference
  on Knowledge Discovery and Data Mining}, 2006.

\bibitem{DBLP:conf/nips/KlambauerUMH17}
G{\"{u}}nter Klambauer, Thomas Unterthiner, Andreas Mayr, and Sepp Hochreiter.
\newblock Self-normalizing neural networks.
\newblock In {\em Advances in Neural Information Processing Systems}, 2017.

\bibitem{cai2020graphnorm}
Tianle Cai, Shengjie Luo, Keyulu Xu, Di~He, Tie{-}Yan Liu, and Liwei Wang.
\newblock {GraphNorm}: {A} principled approach to accelerating graph neural
  network training.
\newblock In {\em Proceedings of the International Conference on Machine
  Learning}, 2021.

\bibitem{xu2018representation}
Keyulu Xu, Chengtao Li, Yonglong Tian, Tomohiro Sonobe, Ken-ichi Kawarabayashi,
  and Stefanie Jegelka.
\newblock Representation learning on graphs with jumping knowledge networks.
\newblock In {\em International Conference on Machine Learning}, 2018.

\bibitem{schulman2017proximal}
John Schulman, Filip Wolski, Prafulla Dhariwal, Alec Radford, and Oleg Klimov.
\newblock Proximal policy optimization algorithms.
\newblock {\em CoRR}, abs/1707.06347, 2017.

\bibitem{andrychowicz2020matters}
Marcin Andrychowicz, Anton Raichuk, Piotr Sta{\'n}czyk, Manu Orsini, Sertan
  Girgin, Rapha{\"e}l Marinier, Leonard Hussenot, Matthieu Geist, Olivier
  Pietquin, Marcin Michalski, Sylvain Gelly, and Olivier Bachem.
\newblock What matters for on-policy deep actor-critic methods? {A} large-scale
  study.
\newblock In {\em International Conference on Learning Representations}, 2021.

\bibitem{ye2020mastering}
Deheng Ye, Zhao Liu, Mingfei Sun, Bei Shi, Peilin Zhao, Hao Wu, Hongsheng Yu,
  Shaojie Yang, Xipeng Wu, Qingwei Guo, et~al.
\newblock Mastering complex control in moba games with deep reinforcement
  learning.
\newblock In {\em Proceedings of the AAAI Conference on Artificial
  Intelligence}, 2020.

\bibitem{kingma2015adam}
P.~Diederik Kingma and Lei~Jimmy Ba.
\newblock Adam: A method for stochastic optimization.
\newblock In {\em International Conference on Learning Representations}, 2015.

\bibitem{NIPS2017_3f5ee243}
Ashish Vaswani, Noam Shazeer, Niki Parmar, Jakob Uszkoreit, Llion Jones,
  Aidan~N Gomez, {\L}ukasz Kaiser, and Illia Polosukhin.
\newblock Attention is all you need.
\newblock In {\em Advances in Neural Information Processing Systems}, 2017.

\bibitem{bello2016neural}
Irwan Bello, Hieu Pham, Quoc~V Le, Mohammad Norouzi, and Samy Bengio.
\newblock Neural combinatorial optimization with reinforcement learning.
\newblock In {\em International Conference on Learning Representations}, 2016.

\bibitem{joshi2022learning}
Chaitanya~K Joshi, Quentin Cappart, Louis-Martin Rousseau, and Thomas Laurent.
\newblock Learning the travelling salesperson problem requires rethinking
  generalization.
\newblock {\em Constraints}, pages 1--29, 2022.

\bibitem{kool2018attention}
Wouter Kool, Herke Van~Hoof, and Max Welling.
\newblock Attention, learn to solve routing problems!
\newblock In {\em International Conference on Learning Representations}, 2018.

\bibitem{DBLP:journals/corr/abs-2201-10494}
Maximilian B{\"{o}}ther, Otto Ki{\ss}ig, Martin Taraz, Sarel Cohen, Karen
  Seidel, and Tobias Friedrich.
\newblock What's wrong with deep learning in tree search for combinatorial
  optimization.
\newblock In {\em International Conference on Learning Representations}, 2022.

\end{thebibliography}



%
}

\clearpage
\newpage

\clearpage
\newpage
\appendix

\section*{Appendix}

\appendix

\section{Extended related work}\label{appendix:related}

\paragraph{Network resilience and utility.} Network utility refers to the system's quality to provide a specific service, for example, transmitting electricity in power networks and transmitting packages in routing networks. A popular metric for network utility is the network efficiency \cite{latora2003economic, boccaletti2006complex}. Network resilience measures the ability to prevent utility loss under failures and attacks. 
In many previous work, despite that network resilience could be improved, the utility may dramatically drop at the same time~\cite{li2019maximizing, carchiolo2019network, wang2014improving,schneider2011mitigation, chan2016optimizing, buesser2011optimizing}. This contradicts the idea behind improving network resilience and will be infeasible in real-world applications. Our goal is to enhance network resilience with moderate loss of network utility by network structure manipulations.

\paragraph{Multi-views graph augmentation for GNNs.} Multi-views graph augmentation is one efficient way to improve the expressive power of GNNs or combine domain knowledge, which is adapted based on the task's prior~\cite{pretraining_strategy}. For example, GCC generates multiple subgraphs from the same ego network~\cite{GCC}. DGI maximizes the mutual information between global and local information~\cite{DGI}. GCA adaptively incorporates various priors for topological and semantic aspects of the graph~\cite{GCA}. \cite {pmlr-v119-hassani20a} contrasts representations from first-order neighbors and a graph diffusion. DeGNN\cite{jin2020graph} was proposed as an automatic graph decomposition algorithm to
improve the performance of deeper GNNs. These techniques rely on the existence of rich graph feature and the resultant GNNs cannot work well on graphs without rich features. In the resilience task, only the graph topological structure is available. Motivated by the calculation process of persistent homology~\cite{edelsbrunner2008persistent}, we apply the filtration process to enhance the expressive power of GNNs for handling graphs without rich features.

\paragraph{Deep graph generation.} Deep graph generation models learn the distribution of given graphs and generate more novel graphs. Some work use the encoder-decoder framework by learning latent representation of the input graph through the encoder and then generating the target graph through the decoder. For example, GCPN~\cite{GCPN} incorporates chemistry domain rules on molecular graph generation. GT-GAN~\cite{GT-GAN} proposes a GAN-based model on malware cyber-network synthesis. GraphOpt \cite{trivedi2020graphopt} learns an implicit model to discover an underlying optimization mechanism of the graph generation using inverse reinforcement learning. GFlowNet learns a stochastic policy for generating molecules with the probability proportional to a given reward based on flow networks and local flow-matching conditions~\cite{bengio2021flow}. Graph structure learning aims to learn an optimized graph structure and corresponding graph representations~\cite{jin2020graph}. However, constrained version of graph generation is still under development and none of existing methods can generate desired graphs with the exact node degree preserving constraint, which is required by the resilience task.

\section{Definitions of different objective functions}
\label{sec:appendix_obj}
In this section, we present resilience definitions and utility definitions used in our experiments. 

\subsection{Resilience definitions} \label{appendix:extra_resilience}
Three kinds of resilience metrics are considered:
\begin{itemize}
    \item  The graph connectivity-based measurement is defined as \cite{schneider2011mitigation} 
\begin{equation*}\label{eq:robustness_R}
  \mathcal{R}(G) = \frac{1}{N}\sum_{q=1}^{N}s(q) \, ,
\end{equation*}
where $s(q)$ is the fraction of nodes in the largest connected remaining graph after removing $q$ nodes from graph $G$ according to certain attack strategy. The range of possible values of $\mathcal{R}$ is  $[1/N, 1/2]$, where these two extreme values correspond to a star network and a fully connected network, respectively. 
    \item The spectral radius ($\mathcal{SR}$) denotes the largest eigenvalue $\lambda_1$ of an adjacency matrix. 
    \item The algebraic connectivity ($\mathcal{AC}$) represents the second smallest eigenvalue of the Laplacian matrix of $G$.
\end{itemize}



\subsection{Utility definitions}\label{appendix:utility_definition}
In this paper, the global and local communication efficiency are used as two measurements of the network utility, which are widely applied across diverse applications of network science, such as transportation and communication networks \cite{latora2003economic, boccaletti2006complex}. 

The average efficiency of a network $G$ is defined as inversely proportional to the average over pairwise distances \cite{latora2001efficient} as
\begin{equation*}
   E(G) = \frac{1}{N(N-1)}\sum_{i\neq j \in V} \frac{1}{d(i, j)} \, ,
\end{equation*}
where $N$ denotes the total nodes in a network and $d(i, j)$ is the length of the shortest path between a node $i$ and another node $j$. 

We can calculate the global and local efficiency given the average efficiency.

\begin{itemize}
    \item The global efficiency of a network $G$ is defined as \cite{latora2001efficient, latora2003economic}
\begin{equation*}
   E_{global}(G) = \frac{E(G)}{E(G^{ideal})} \, ,
\end{equation*}
where $G^{ideal}$ is the ``ideal'' fully-connected graph on $N$ nodes and the range of $E_{global}(G)$ is [0, 1].
    
    \item The local efficiency of a network $G$ measures a  local average of pairwise communication efficiencies and is defined as \cite{latora2001efficient}
\begin{equation*}
   E_{local}(G) = \frac{1}{N}{\sum_{i \in V} E(G_i)} \, ,
\end{equation*}
where $G_i$ is the local subgraph including only of a node $i$'s one-hop neighbors, but not the node $i$ itself. The range of $E_{local}(G)$ is [0, 1].
    
\end{itemize}


\section{Implementation details of ResiNet}
This section provides the implementation details of ResiNet, including dataset, network structure training strategies, and node feature construction.

\subsection{Dataset}\label{appendix:datasets}
We first present the data generation strategies. Table \ref{tb:dataset} summarizes the statistics of each dataset.
Synthetic datasets are generated using the Barabasi-Albert (BA) model (known as scale-free graphs) \cite{albert2002statistical}, with the graph size varying from $|N|$=10 to $|N|$=1000. 
During the data generation process, each node is connected to two existing nodes for graphs with no more than 500 nodes, and each node is connected to one existing node for graphs with near 1000 nodes. BA graphs are chosen since they are vulnerable to malicious attacks and are commonly used to test network resilience optimization algorithms\cite{bollobas2004robustness}.
We test the performance of ResiNet on both transductive and inductive settings.
\begin{itemize}
    \item \textbf{Transductive setting.} \quad The algorithm is trained and tested on the same network. 
         \begin{itemize}
              \item Randomly generated synthetic BA networks, denoted by BA-$m$, are adopted to test the performance of ResiNet on networks of various sizes, where $m \in \{15, 50, 100, 500, 1000 \}$ is the graph size.
            \item  The Gnutella peer-to-peer network file sharing network from August 2002~\cite{leskovec2007graph, ripeanu2002mapping} and the real EU power network \cite{zhou2005approximate} are used to validate the performance of ResiNet on real networks. The random walk sampling strategy is used to derive a representative sample subgraph with hundreds of nodes from the Gnutella peer-to-peer network~\cite{leskovec2006sampling}.
            \end{itemize}
    \item \textbf{Inductive setting.} \quad Two groups of synthetic BA networks denoted by BA-$m$-$n$ are randomly generated to test ResiNet's inductivity, where $m$ is the minimal graph size, and $n$ indicates the maximal graph size. We first randomly generate the fixed number of BA networks as the training data to train ResiNet and then evaluate ResiNet's performance directly on the test dataset without any additional optimization. 
\end{itemize}


 \begin{table}[t]
  \caption{Statistics of graphs used for resilience maximization. Both transductive and inductive settings ($\star$) are included. Consistent with our implementation, we report the number of edges by transforming undirected graphs to directed graphs. The edge rewiring has a fixed execution order. For the inductive setting, we report the maximum number of edges. The action space size of the edge rewiring is measured by $2|E|^2$.}
  \label{tb:dataset}
  \centering
    \scalebox{0.9}{
  \begin{tabular}{llllll}
    \toprule
      Dataset   & Node & Edge & Action Space Size & Train/Test  & Setting \\
    \midrule
    BA-15  & 15 & 54 & 5832 & \ding{55}  & Transductive \\
    BA-50   & 50 & 192 & 73728 & \ding{55} &  Transductive\\
    BA-100   & 100 & 392  & 307328 & \ding{55} & Transductive\\
      BA-500   & 500 &  996 & 1984032 & \ding{55} & Transductive\\
        BA-1000   & 1000 & 999  & 1996002 & \ding{55} & Transductive\\
    EU & 217  & 640 & 819200 & \ding{55} & Transductive\\
     p2p-Gnutella05 & 400  & 814  & 1325192 & \ding{55} & Transductive\\
      p2p-Gnutella09 & 300  & 740 & 1095200 & \ding{55} & Transductive\\
    BA-10-30 ($\star$) & 10-30  & 112 & 25088 & 1000/500 & Inductive\\
      BA-20-200 ($\star$) & 20-200  & 792 & 1254528 & 4500/360 & Inductive \\
    \bottomrule
  \end{tabular}}
\end{table}

\subsection{ResiNet setup} \label{appendix:resinet_setup}
In this section, we provide detailed parameter setting and training strategies for ResiNet.

Our proposed FireGNN is used as the graph encoder in ResiNet, including a 5-layer defined GIN \cite{xu2018how} as the backbone. The hidden dimensions for node embedding and graph embedding in each hidden layer are set to 64 and the SeLU activation function \cite{DBLP:conf/nips/KlambauerUMH17} is used after each message passing propagate. 
Graph normalization strategy is adopted to stabilize the training of GNN \cite{cai2020graphnorm}. The jumping knowledge network \cite{xu2018representation} is used to aggregate node features from different layers of the GNN.

The overall policy is trained by using the highly tuned implementation of proximal policy optimization (PPO) algorithm \cite{schulman2017proximal}. 
Several critical strategies for stabilizing and accelerating the training of ResiNet are used, including advantage normalization \cite{andrychowicz2020matters}, the dual-clip PPO (the dual clip parameter is set to 10) \cite{ye2020mastering}, and the usage of different optimizers for policy network and value network. Additionally, since the step-wise reward range is small (around 0.01), we scale the reward by a factor of 10 to aim the training of ResiNet. The policy head model and value function model use two separated FireGNN encoder networks with the same architecture. 
ResiNet is trained using two separate Adam optimizers \cite{kingma2015adam} with batch size 256 and a linearly decayed learning rate of 0.0007 for the policy network and a linearly decayed learning rate of 0.0007 for the value network. The aggregation function of FireGNN is defined as an attention mechanism-based linear weighted combination.

\textbf{Hardware}: We run all experiments for ResiNet on the platform with two GEFORCE RTX 3090 GPU and one AMD 3990X CPU.

\subsection{Node feature construction}\label{appendix:node_features}
 The widely-used node degree feature cannot significantly benefit the network resilience optimization of a single graph due to the degree-preserving rewiring. Therefore, we construct node features for each input graph to aid the transductive learning and inductive learning, including

\begin{itemize}
  \item The distance encoding strategy \cite{li2020distance}. Node degree feature is a part of it.
  
    \item The 8-dimensional position embedding originating from the Transformer \cite{NIPS2017_3f5ee243} as the measurement of the vulnerability of each node under attack. If the attack order is available, we can directly encode it into the position embedding. If the attack order is unknown, node degree, node betweenness, and other node priority metrics can be used for approximating the node importance in practice. In our experiments, we used the adaptive node degree for the position embedding.
  
\end{itemize}

\subsection{Baseline setup}
All baselines share the same action space with ResiNet and use the same action masking strategy to block invalid actions as ResiNet does. The maximal objective evaluation is consistent for all algorithms. Other settings of baselines are consistent with the default values in their paper. The early-stopping strategy is used for baselines, which means that the search process terminates if no improvement is obtained in successive 1000 objective function calling trials.
For

\section{Deep analysis of why regular GNNs fail in the resilience task}\label{sec:whygnnfails}

It is well-known that GNNs generally work well for graphs with rich features. Unluckily, the graph network in the resilience task has no node/edge/graph feature, with only the topological structure available. No rich feature means that the output of the GNNs is not distinguishable, and then it is difficult for the RL agent to distinguish different vertices/edges, causing large randomness in the output of the policy. 
This may cause the rewiring process to alternate between two graphs, forming an \textit{infinite loop}.  
And we suspect that this infinite loop failure may explain the poor empirical performance of optimizing network resilience by selecting edges using existing GNNs and reinforcement learning (RL). The infinite loop failure is presented as follows.

Consider the graph $G_t$ with $N$ nodes and containing two edges $AB$ and $CD$. 
The agent selects $AB$ and $CD$ for rewiring, leading to $G_{t+1}$ with news edges $AC$ and $BD$. 
A frequent empirical failure of regular GNNs for the resilience task is the \textit{infinite loop} phenomenon. 
The agent would select $AC$ and $BD$ at step $t+1$, returning back to $G_t$ and forming a cycled loop between $G_t$ and $G_{t+1}$. Formally, the infinite loop is formulated as
\begin{equation}
 \begin{aligned}
       ((A, B), (C, D)) &=  \argmax_{i, j, m, n \in 1: N} \text{SIM}\left(((h_t^i, h_t^j), (h_t^m, h_t^n)), h_{G_t} \right)   \nonumber  \\
      ((A, C), (B, D)) &=  \argmax_{i, j, m, n \in 1: N} \text{SIM}\left(((h^i_{t+1}, h_{t+1}^j), (h_{t+1}^m, h_{t+1}^n)), h_{G_{t+1}} \right) \, ,  \nonumber 
\end{aligned}   
\end{equation}
where SIM is a similarity metric, $h^i_t$ and $h_{G_t}$ are embeddings of node $i$ and graph $G_t$ at step $t$, and ($A$,$B$) is one edge.

Table \ref{tb:compare2othercotask} compares and summarizes different graph related tasks' characteristics. We can see that
\textit{the resilience task is more challenging from many aspects}. No prior rule like action masking or negative penalty can be used to avoid selecting visited nodes as in TSP. For the resilience task, all previously visited edges are also possibly valid to be selected again, resulting in insufficient training signals.

The desired GNN model should not depend on rules like action masking to distinguish edge and graph representations for graphs with little node features.
Our proposed FireGNN fulfills these requirements to obtain proper training signals.  FireGNN has a distinct expressive power and learns to create more meaningful and distinguishable features for each edge. FireGNN is not a simple aggregation of higher-order information of a static graph. It was inspired by homology filtration and the multi-view graph augmentation. Persistence homology motivates us to aggregate more distinct node features by observing how the graph evolves towards being empty, leading to more distinct and meaningful features for each node/edge, thus avoiding the infinite loop. Extensive experimental results in Table  \ref{tb:res_resi_gain_fixed_budget} validate the necessity and effectiveness of FireGNN. Existing GNNs perform worse while FireGNN performs well.

\begin{table}[t]

\caption{Characteristics of different graph related tasks.}
\label{tb:compare2othercotask}
\begin{adjustbox}{width=\columnwidth,center}

\begin{tabular}{|llllllclcl|}
\hline
\multirow{2}{*}{\textbf{Approach}}  & 
\multirow{2}{*}{\textbf{Task}}  & \multicolumn{3}{c}{\textbf{RL component}}  & 
\textbf{Problem}  &
\multicolumn{4}{c|}{\textbf{Training \& Inference}} 

\\  \cline{3-5} \cline{7-10} 

 &  & \textbf{State}  & \textbf{Action} &  \textbf{Reward} & \textbf{Complexity}& \textbf{Size Extrapolate} & \textbf{Encoder} &\textbf{Action Masking} &  \textbf{Scalability}

  \\ [0.5ex]  \hline\hline
\multirow{3}{*}{S2V-DQN \cite{CO_on_graph}} 
     & MVC & node level & add node to subset & -1 & $\mathcal{O}(N)$ & \ding{55} & S2V & \checkmark & 500 \\ 
     & Max-Cut  & node level & add node to subset & change in cut weight & $\mathcal{O}(N)$ & \ding{55} & S2V  & \checkmark  & 300   \\ 
     & TSP  & node level  & add node to tour & change in tour cost & $\mathcal{O}(N)$ & \ding{55} & S2V   & \checkmark  & 300
     \\ \hline
Local search~\cite{hudson2021graph} & TSP  & edge level  & relocate node in tour & global regret & $\mathcal{O}(N^2)$ & \checkmark & GNN  & \checkmark & 100  \\ \hline
RNN-RL~\cite{bello2016neural}   & TSP & node level  & add node to tour  & change in tour cost  &  $\mathcal{O}(N)$ & \ding{55}  & RNN & \checkmark & 100  \\ \hline
GNN-RL~\cite{joshi2022learning}  & TSP & node level & add node to tour  & change in tour cost & $\mathcal{O}(N)$ & \checkmark  & GNN & \checkmark   & 50  \\ \hline
\multirow{2}{*}{Attention-RL~\cite{kool2018attention}} & TSP & node level & add node to tour  & change in tour cost & $\mathcal{O}(N)$ & \ding{55}  & Attention & \checkmark & 100  \\ 
     & VRP  & node level  & add node to tour  & change in tour cost & $\mathcal{O}(N)$ & \ding{55} & Attention & \checkmark & 100  \\ \hline
Local search~\cite{DBLP:journals/corr/abs-2201-10494}& MIS  & node level  & add node to subset & change in IS size & $\mathcal{O}(N)$ & \checkmark & GNN  & \checkmark  & 800     \\ \hline

\textbf{ResiNet}  & Resilience & \textbf{\textit{graph level}}  & \textit{\textbf{edge rewiring}}   & change in resilience and utility & $\mathcal{O}(N^4)$  & \checkmark  & \textit{\textbf{FireGNN}}    & \ding{55}    & \textit{\textbf{1000}}      \\ \hline
\end{tabular}

\end{adjustbox}

\end{table}

\section{Extended experimental results}
\label{appendix:extra_results}
In this section, we present additional experimental results.

This section provides additional experimental results, including optimizing different resilience and utility metrics and validating ResiNet's inductivity on larger datasets. Finally, we describe future work.


\subsection{Learning to balance more utility and resilience metrics} 
\label{sec:res_multiobjs}
\begin{figure}[t]
  \centering
  \begin{subfigure}[b]{0.17\textwidth}
         \centering
    \includegraphics[width=\textwidth]{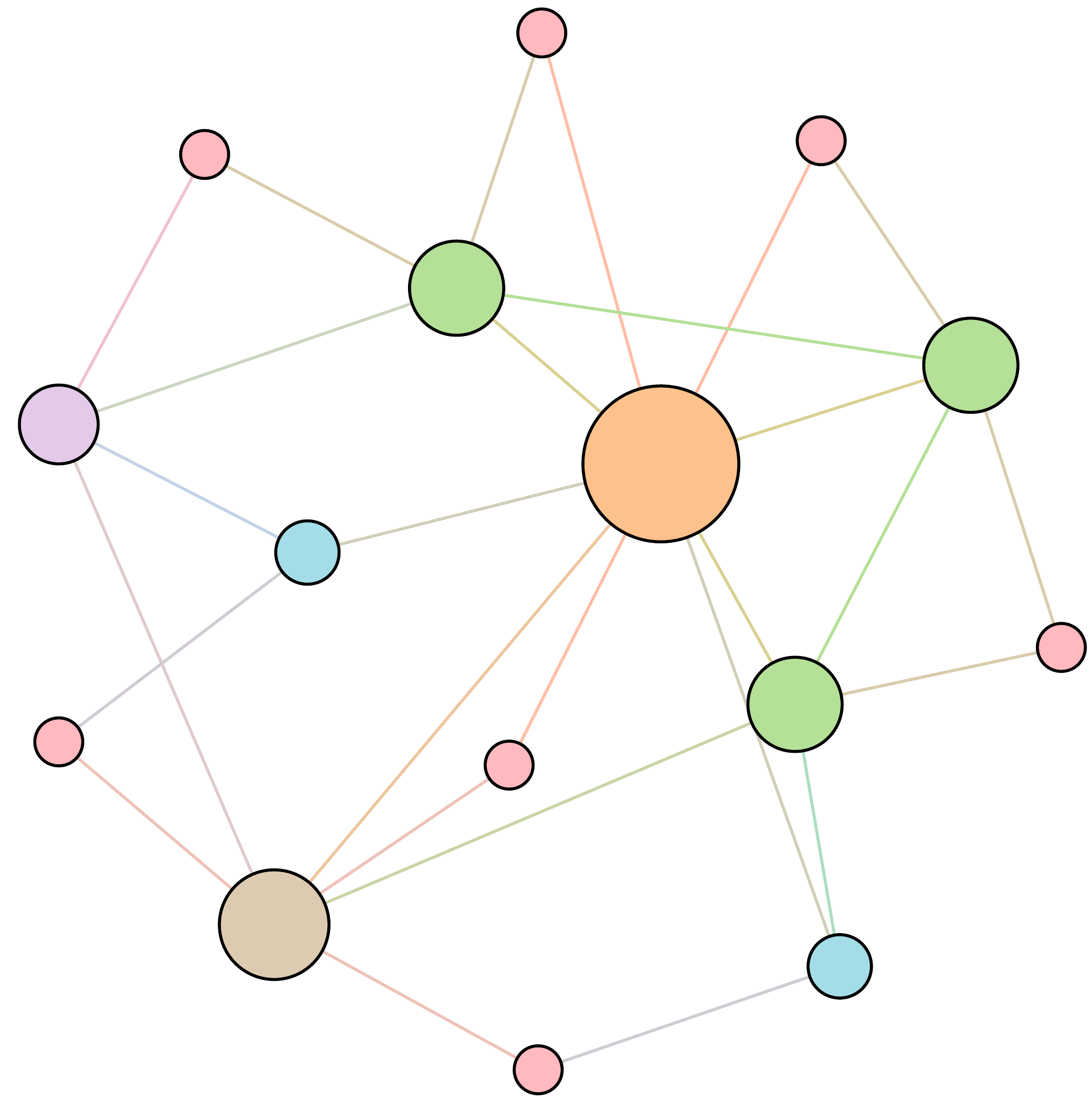}
    \caption{Original}
     \end{subfigure}%
        \hfill
     \begin{subfigure}[b]{0.17\textwidth}
         \centering
        \includegraphics[width=\textwidth]{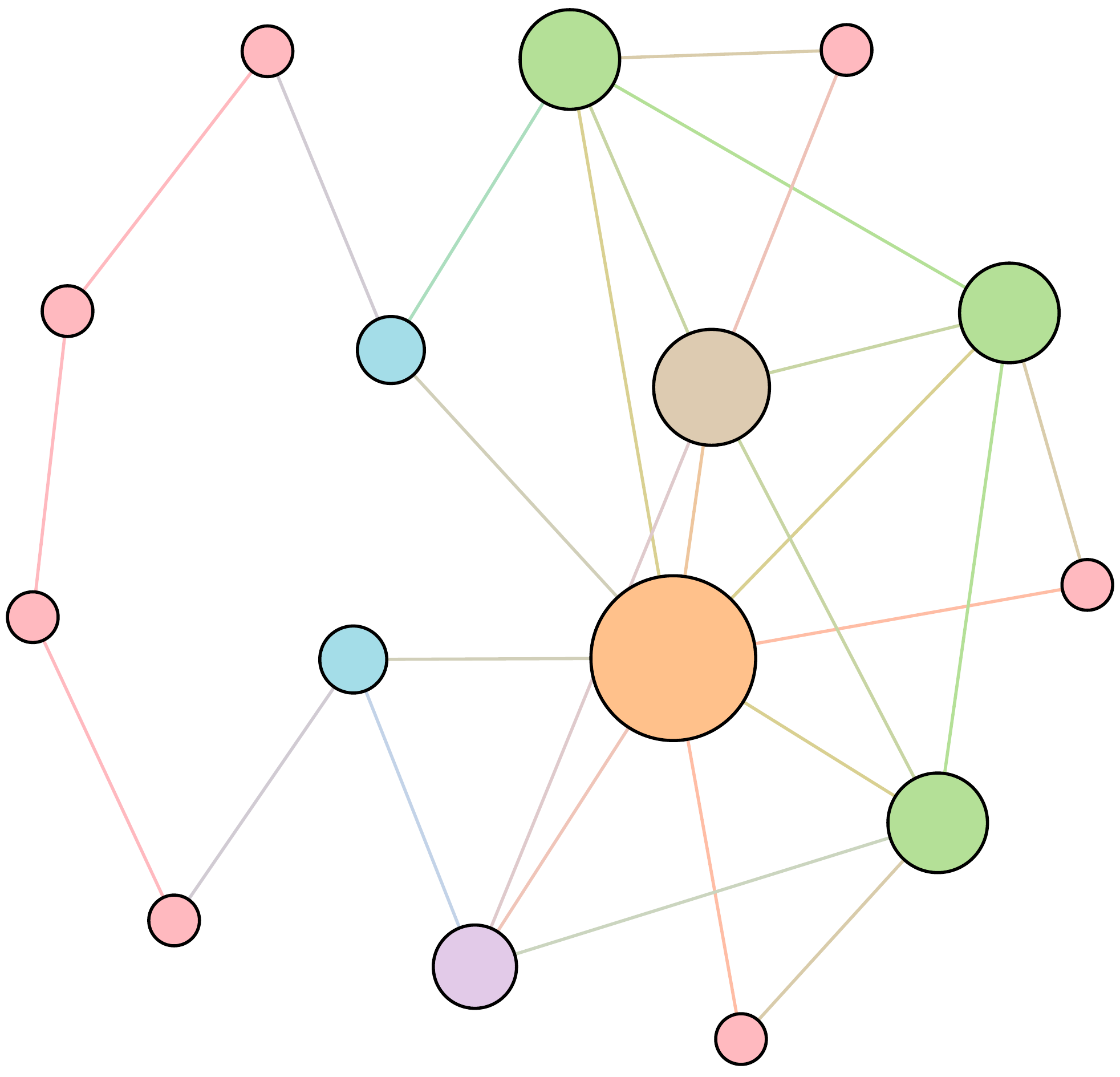}
        \caption{$\mathcal{R}_D$}
     \end{subfigure}%
             \hfill
     \begin{subfigure}[b]{0.17\textwidth}
         \centering
        \includegraphics[width=\textwidth]{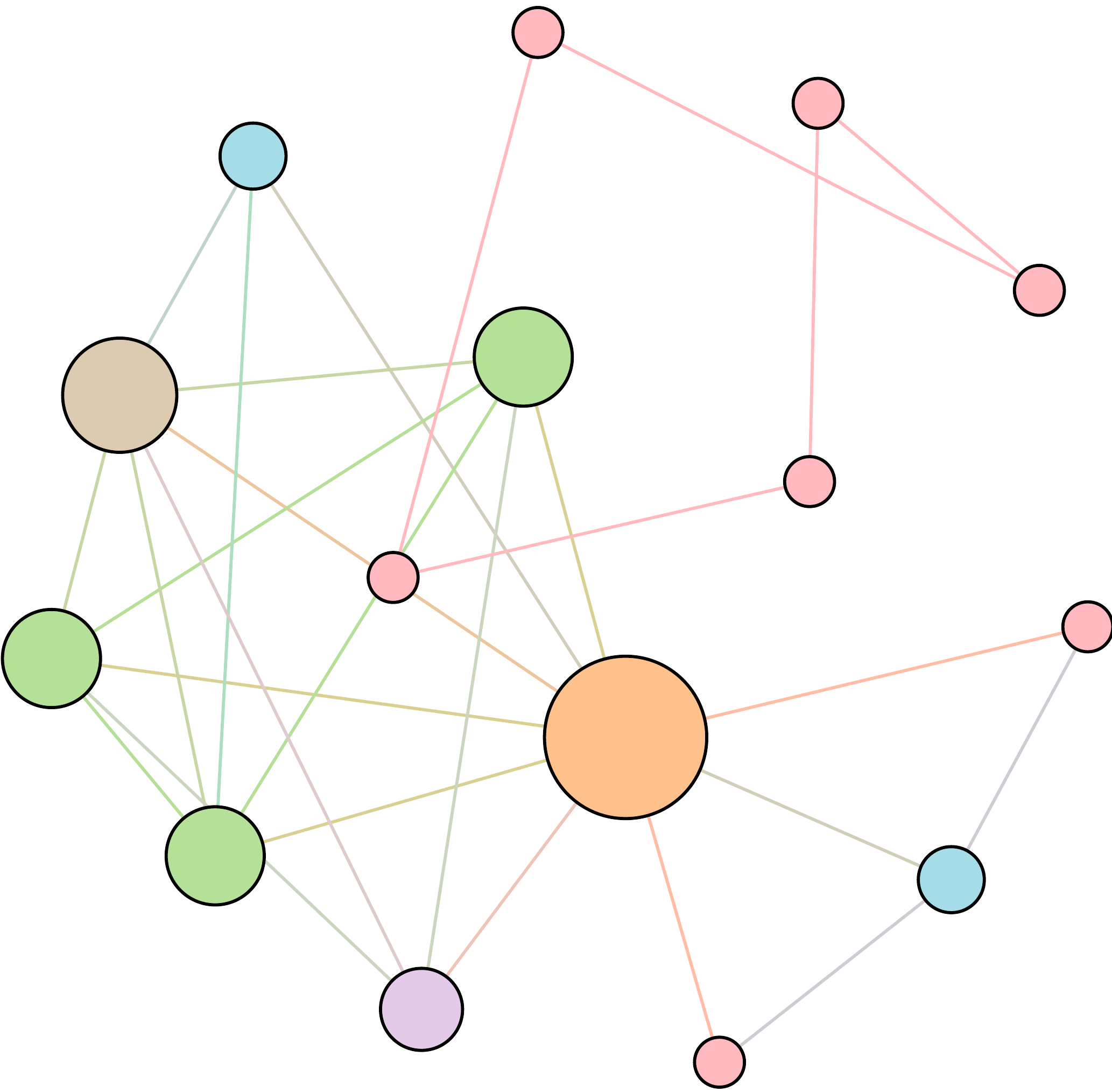}
        \caption{$\mathcal{SR}_D$}
         \end{subfigure}
            \hfill
     \begin{subfigure}[b]{0.17\textwidth}
         \centering
        \includegraphics[width=\textwidth]{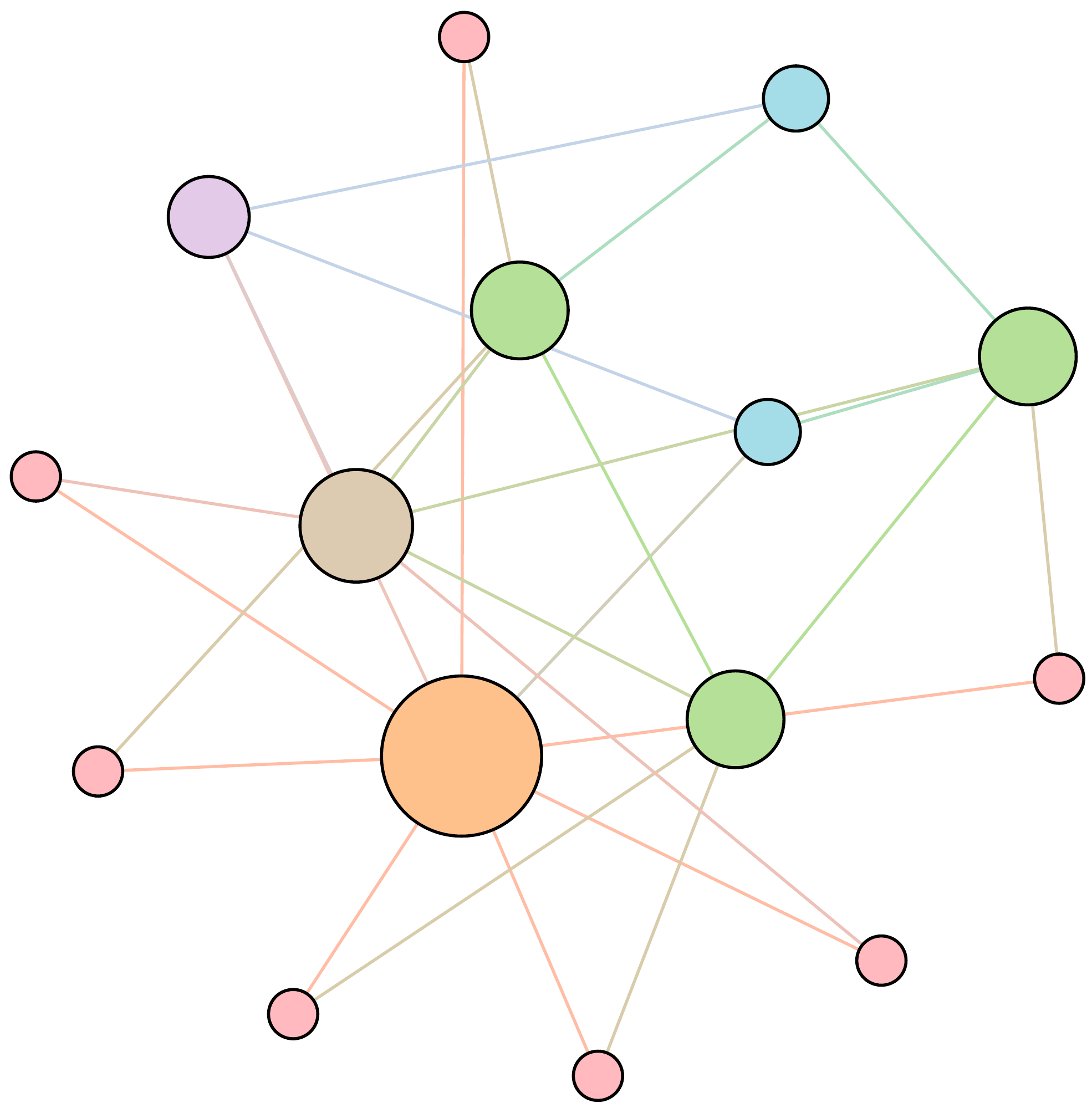}
        \caption{$\mathcal{AC}_D$}
         \end{subfigure}
         \hfill
      \begin{subfigure}[b]{0.17\textwidth}
             \centering
            \includegraphics[width=\textwidth]{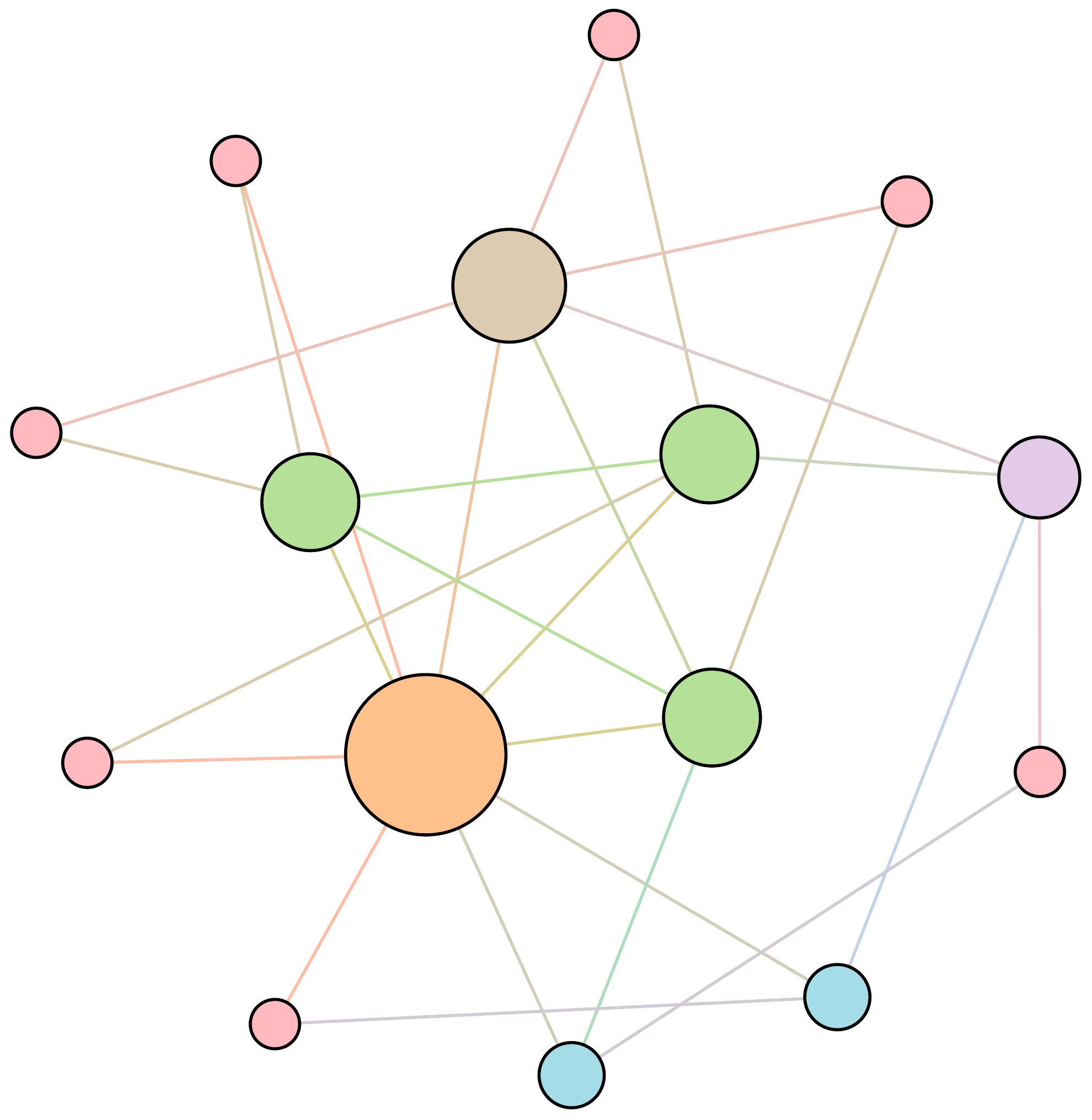}
            \caption{$\mathcal{R}_D$ + $E_{global}$}
             \end{subfigure}
  \begin{subfigure}[b]{0.17\textwidth}
         \centering
    \includegraphics[width=\textwidth]{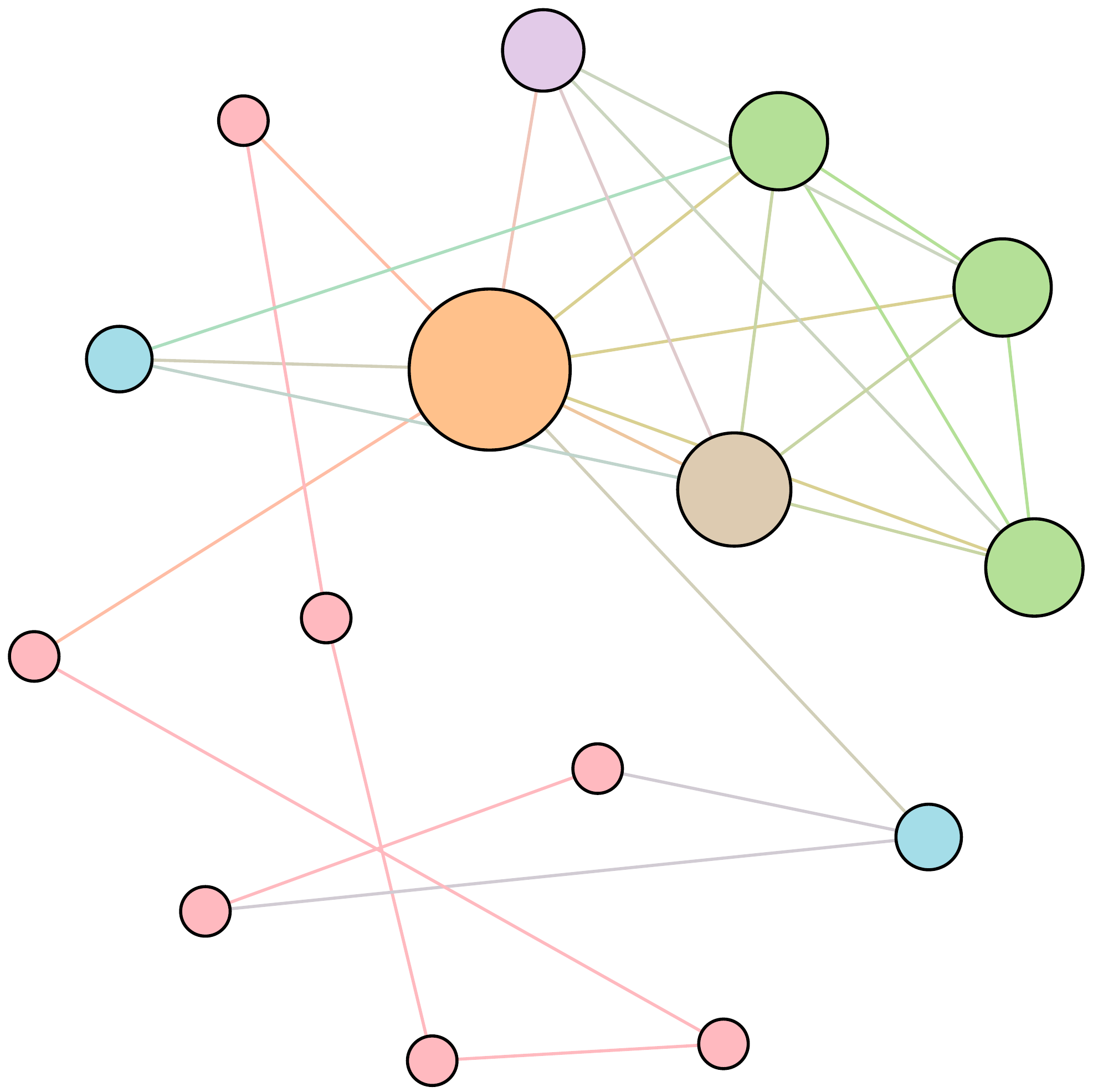}
    \caption{$\mathcal{SR}_D$ + $E_{global}$}
     \end{subfigure}%
        \hfill
     \begin{subfigure}[b]{0.17\textwidth}
         \centering
        \includegraphics[width=\textwidth]{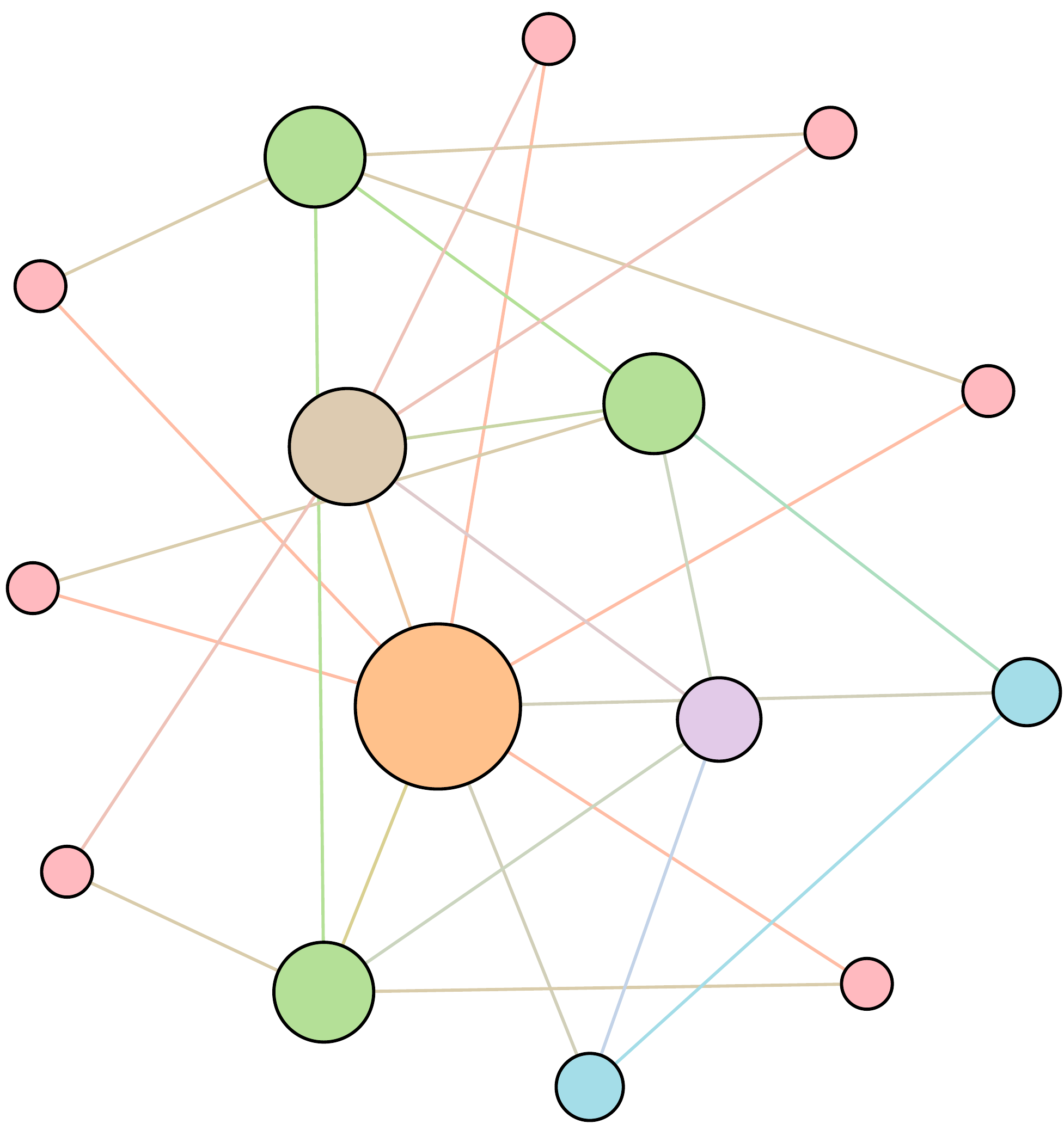}
        \caption{$\mathcal{AC}_D$ + $E_{global}$}
     \end{subfigure}%
             \hfill
     \begin{subfigure}[b]{0.17\textwidth}
         \centering
        \includegraphics[width=\textwidth]{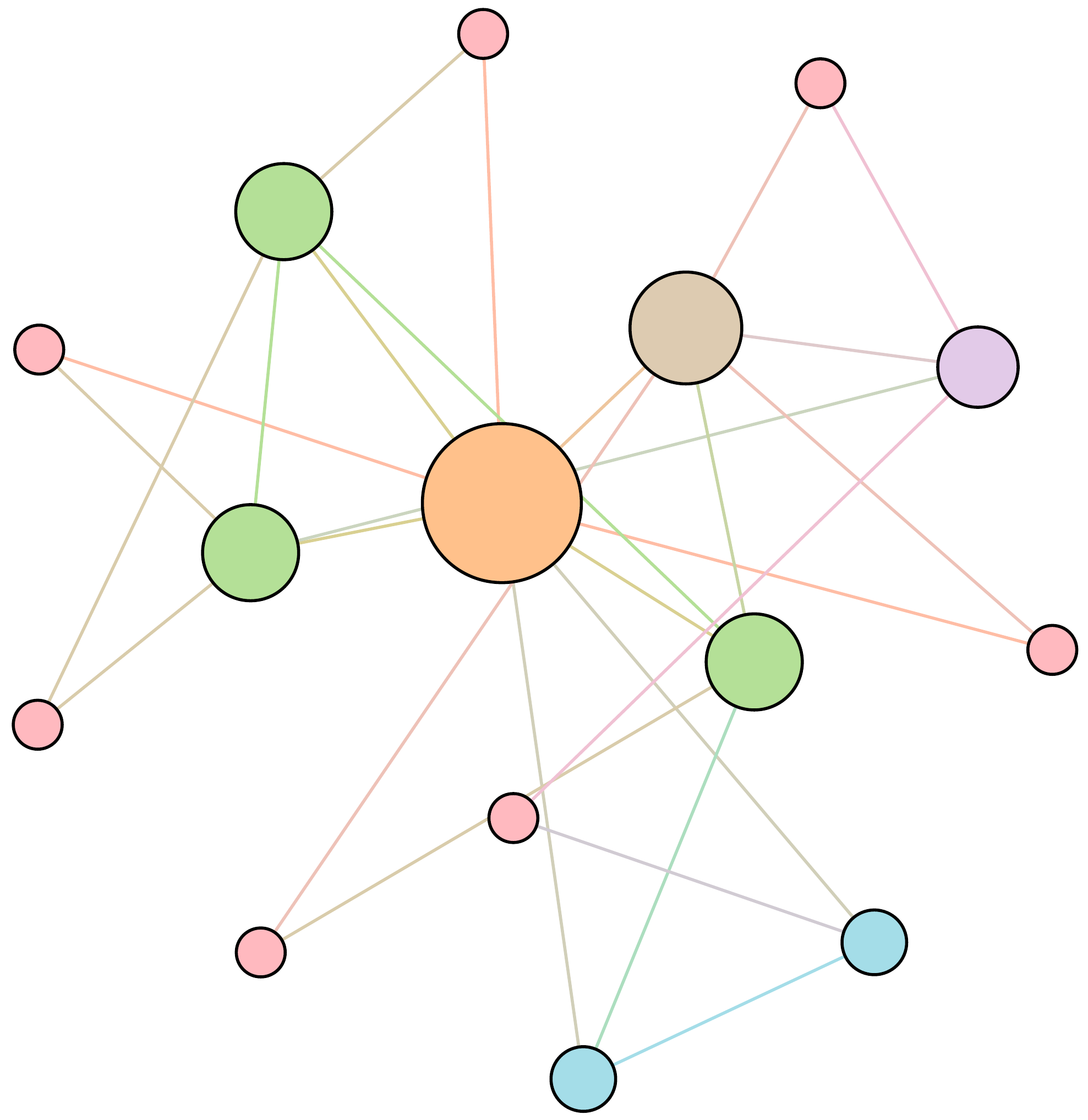}
        \caption{$\mathcal{R}_D$ + $E_{local}$}
         \end{subfigure}
            \hfill
     \begin{subfigure}[b]{0.17\textwidth}
         \centering
        \includegraphics[width=\textwidth]{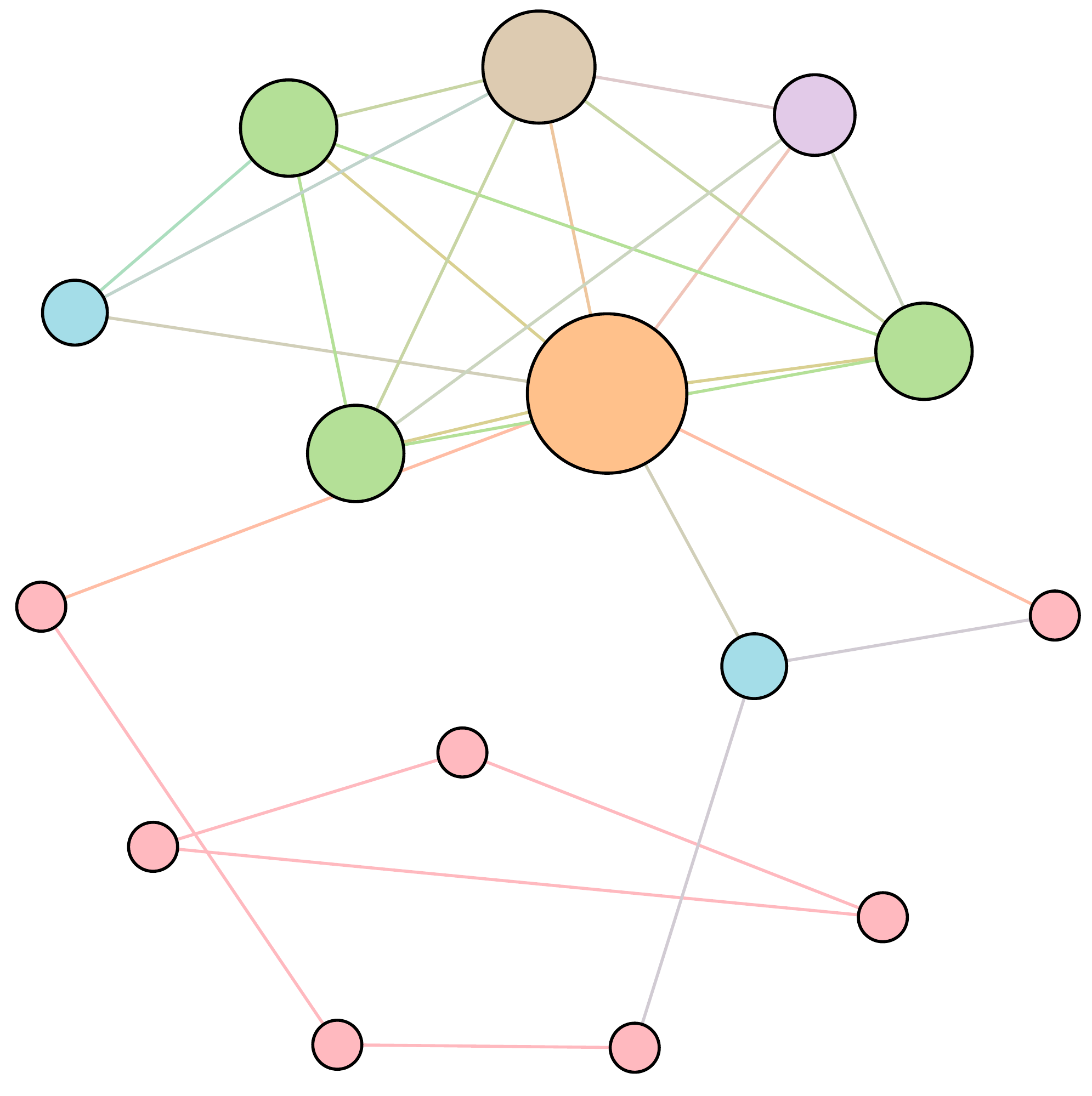}
        \caption{$\mathcal{SR}_D$ + $E_{local}$}
         \end{subfigure}
         \hfill
      \begin{subfigure}[b]{0.17\textwidth}
             \centering
            \includegraphics[width=\textwidth]{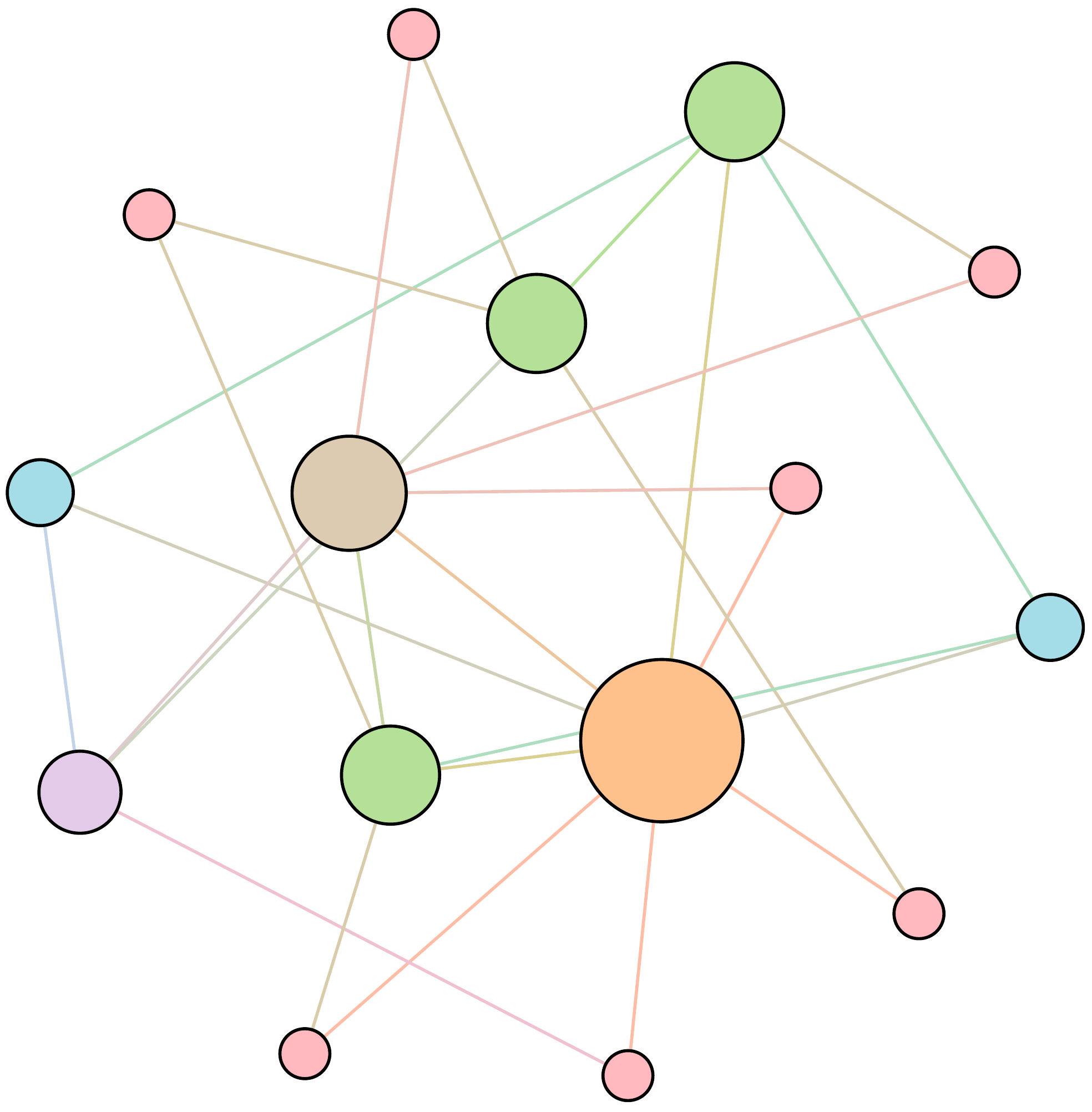}
            \caption{$\mathcal{AC}_D$ + $E_{local}$}
             \end{subfigure}
     \begin{subfigure}[b]{0.17\textwidth}
         \centering
    \includegraphics[width=\textwidth]{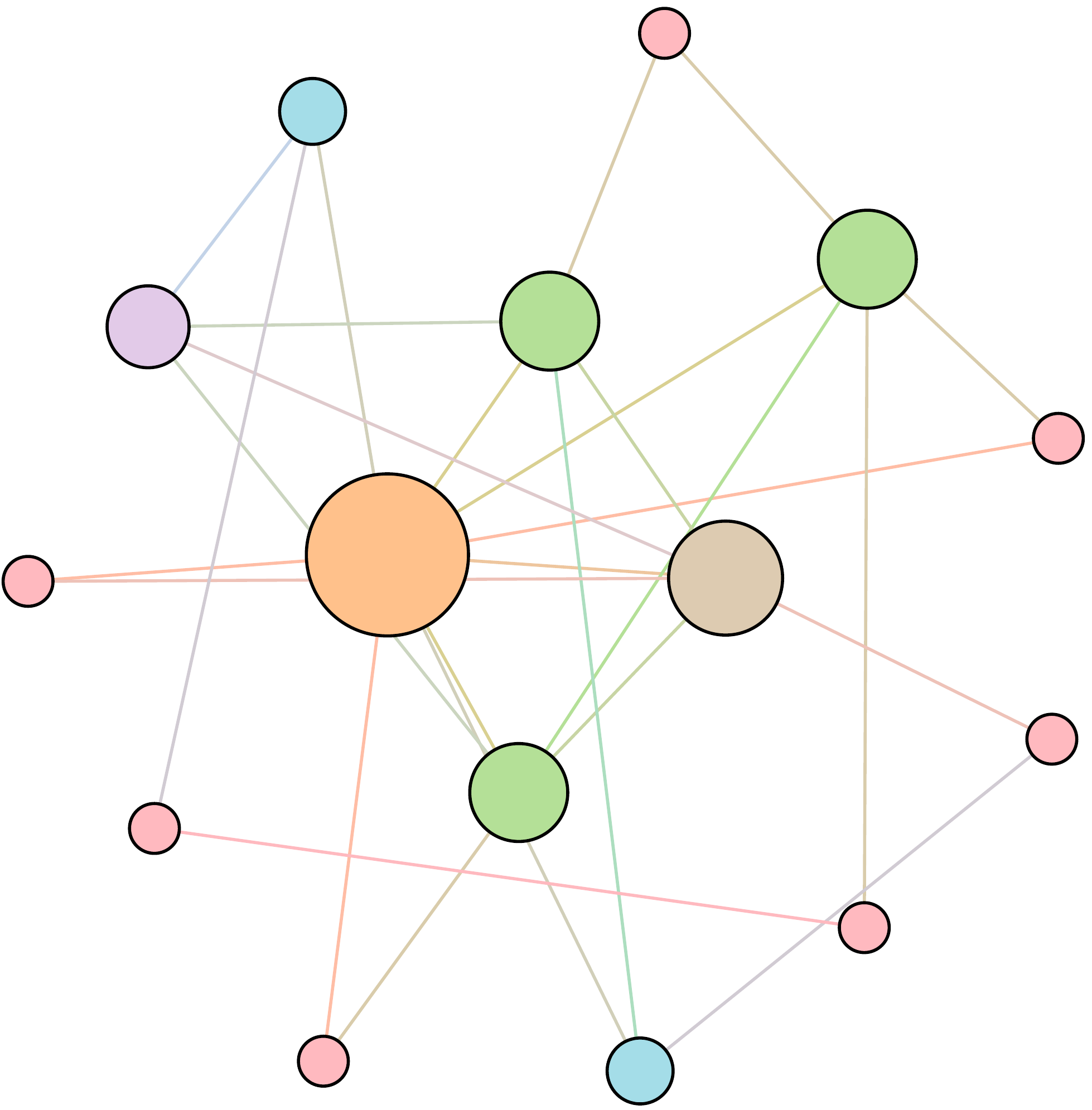}
    \caption{$\mathcal{R}_B$}
     \end{subfigure}%
        \hfill
     \begin{subfigure}[b]{0.17\textwidth}
         \centering
        \includegraphics[width=\textwidth]{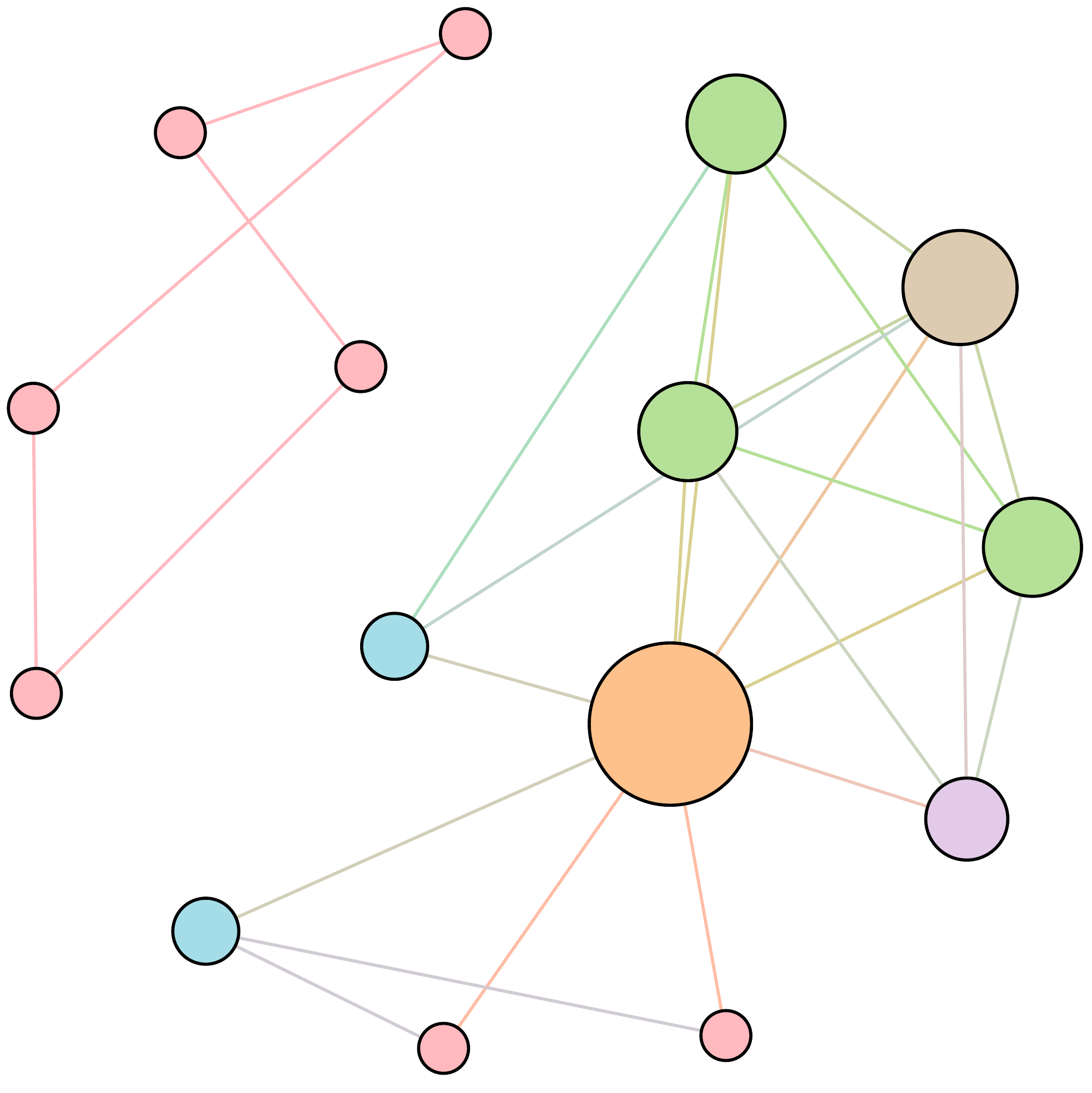}
        \caption{$\mathcal{SR}_B$}
     \end{subfigure}%
             \hfill
     \begin{subfigure}[b]{0.17\textwidth}
         \centering
        \includegraphics[width=\textwidth]{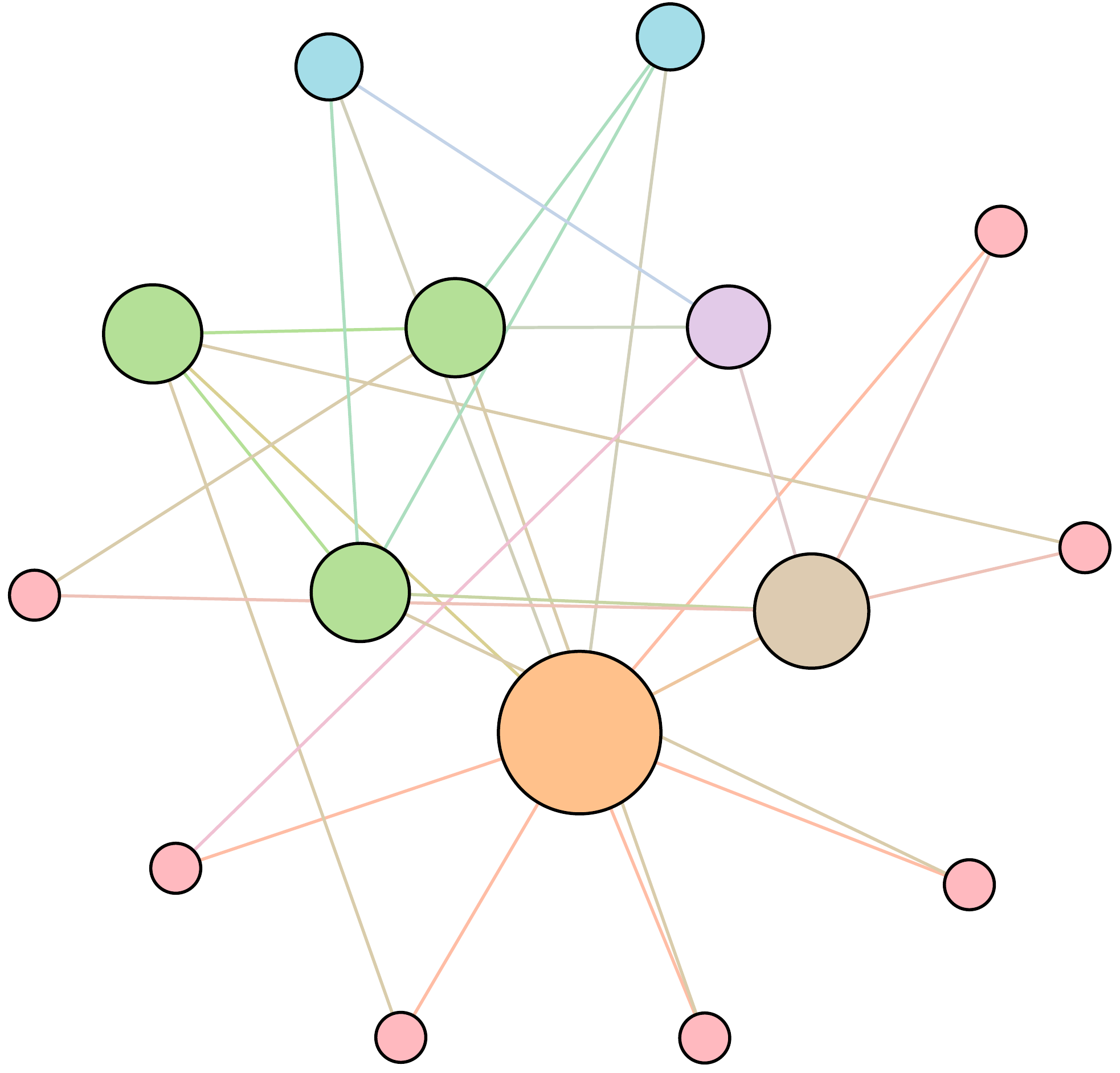}
        \caption{$\mathcal{AC}_B$}
         \end{subfigure}
            \hfill
     \begin{subfigure}[b]{0.17\textwidth}
         \centering
        \includegraphics[width=\textwidth]{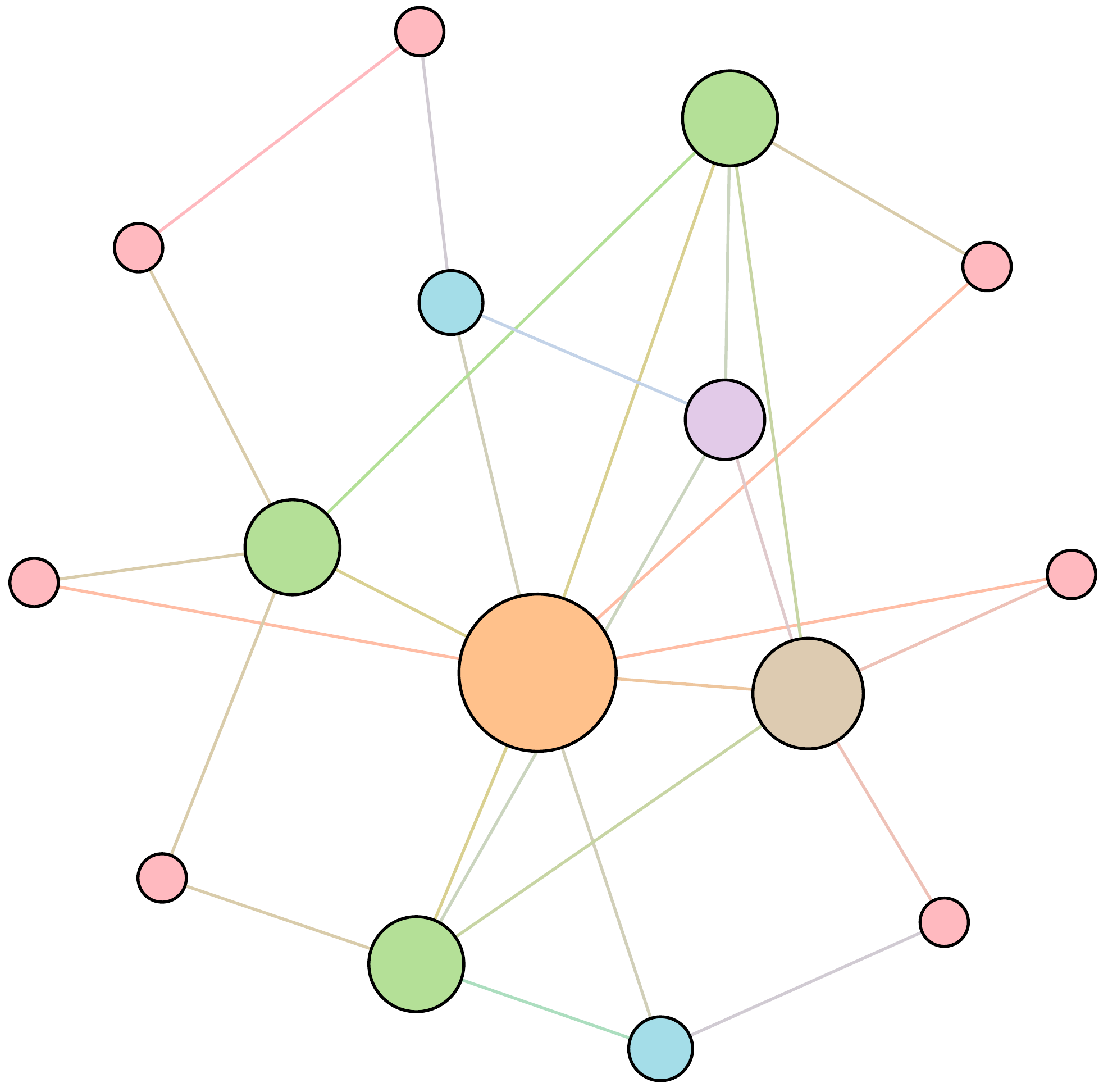}
        \caption{$\mathcal{R}_B$ + $E_{global}$}
         \end{subfigure}
         \hfill
      \begin{subfigure}[b]{0.17\textwidth}
             \centering
            \includegraphics[width=\textwidth]{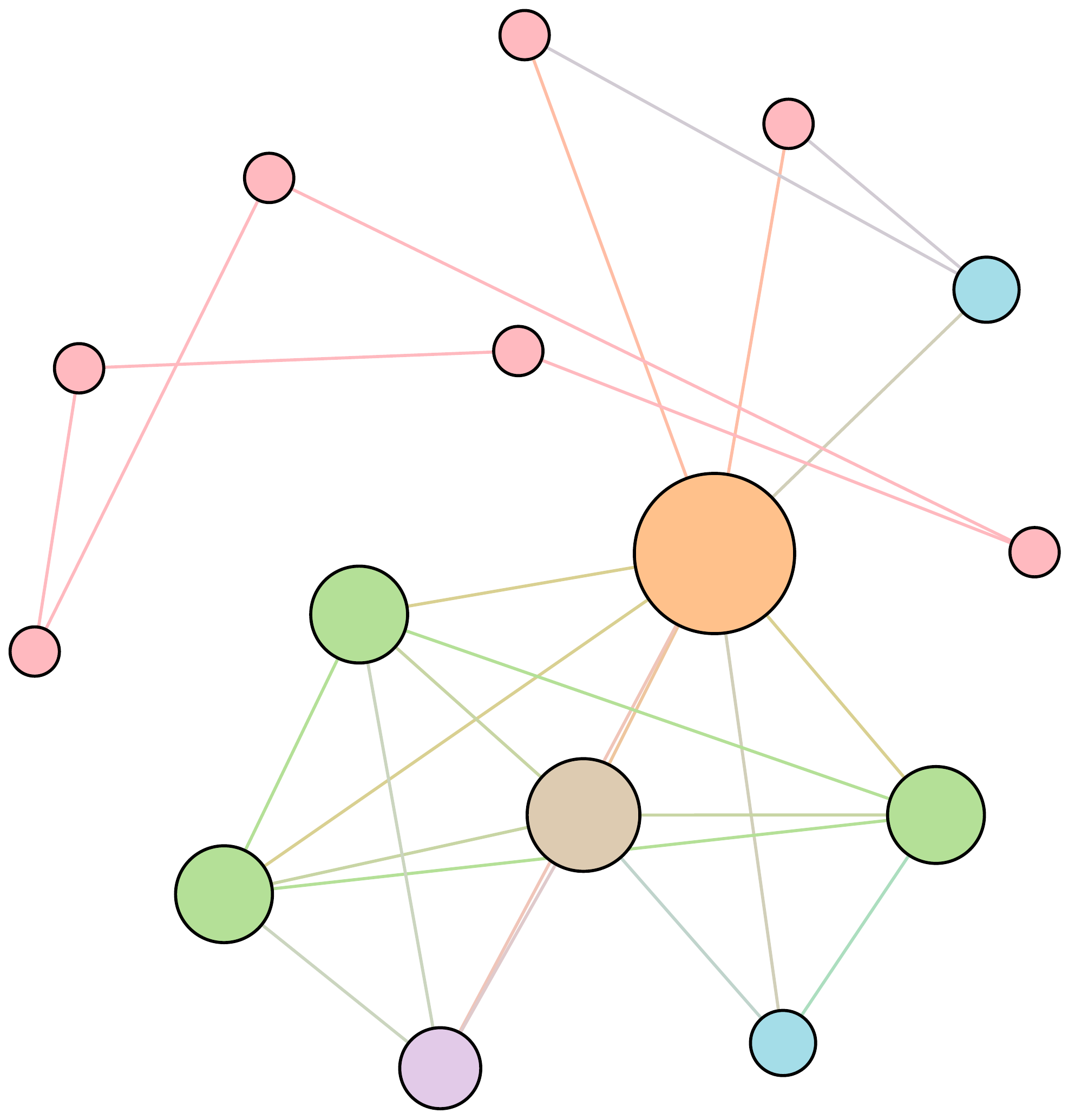}
            \caption{{\scriptsize $\mathcal{SR}_B$+$E_{global}$}}
             \end{subfigure}
           \\
    \begin{subfigure}[b]{0.17\textwidth}
         \centering
    \includegraphics[width=\textwidth]{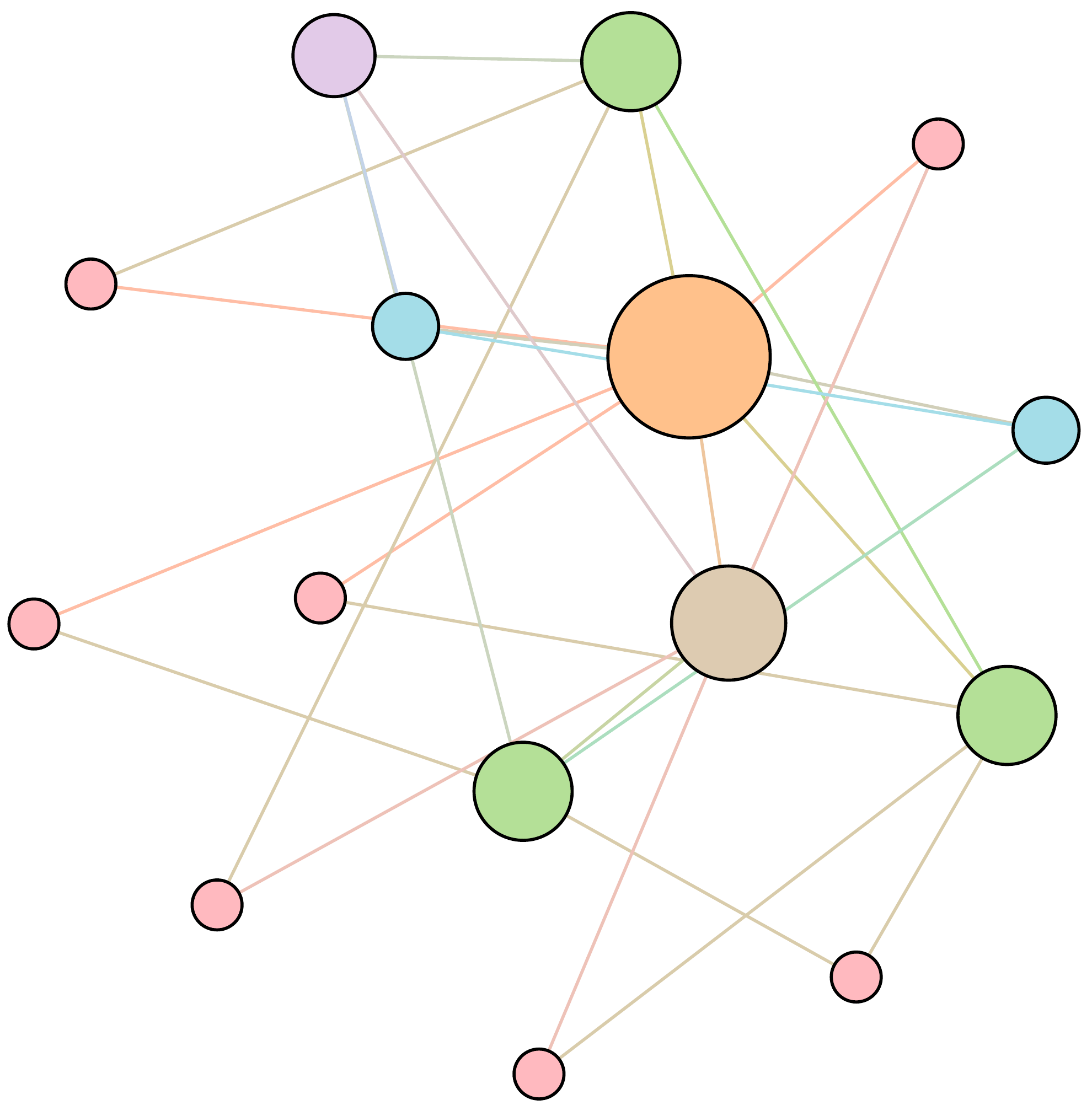}
    \caption{$\mathcal{AC}_B$ + $E_{global}$}
     \end{subfigure}%
        \hfill
     \begin{subfigure}[b]{0.17\textwidth}
         \centering
        \includegraphics[width=\textwidth]{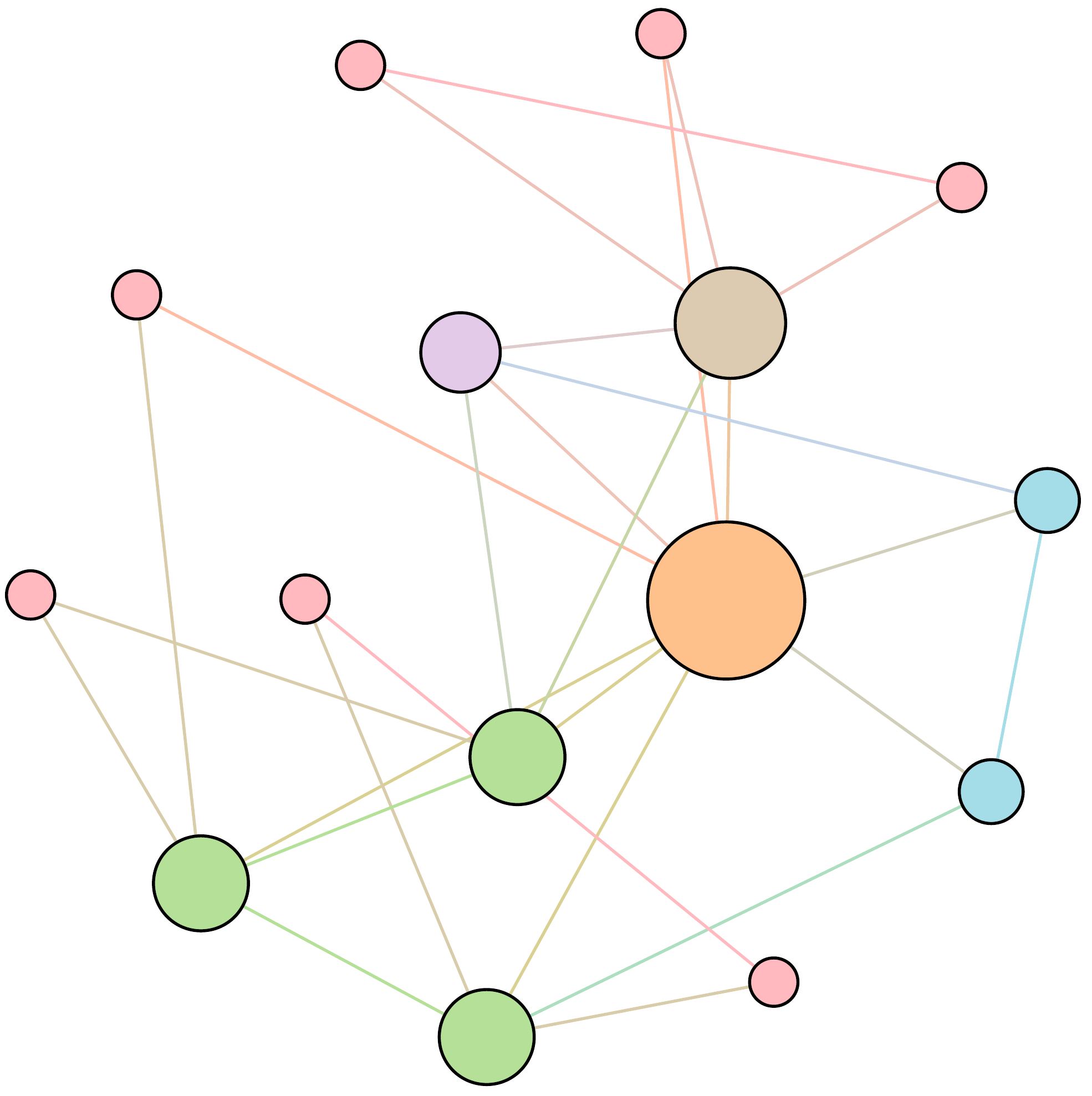}
        \caption{$\mathcal{R}_B$ + $E_{local}$}
     \end{subfigure}%
             \hfill
     \begin{subfigure}[b]{0.17\textwidth}
         \centering
        \includegraphics[width=\textwidth]{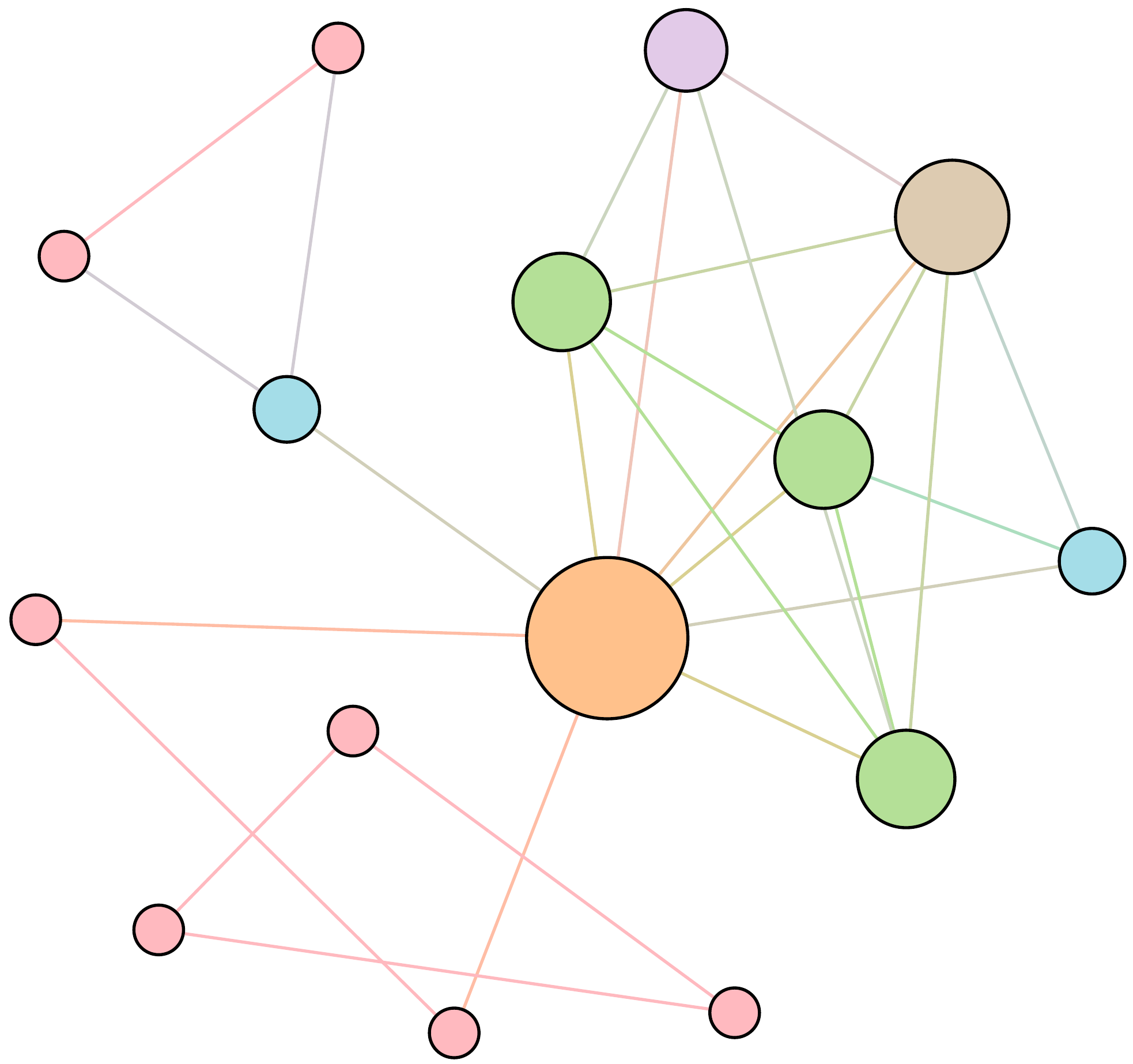}
        \caption{$\mathcal{SR}_B$+$E_{local}$}
         \end{subfigure}
            \hfill
     \begin{subfigure}[b]{0.17\textwidth}
         \centering
        \includegraphics[width=\textwidth]{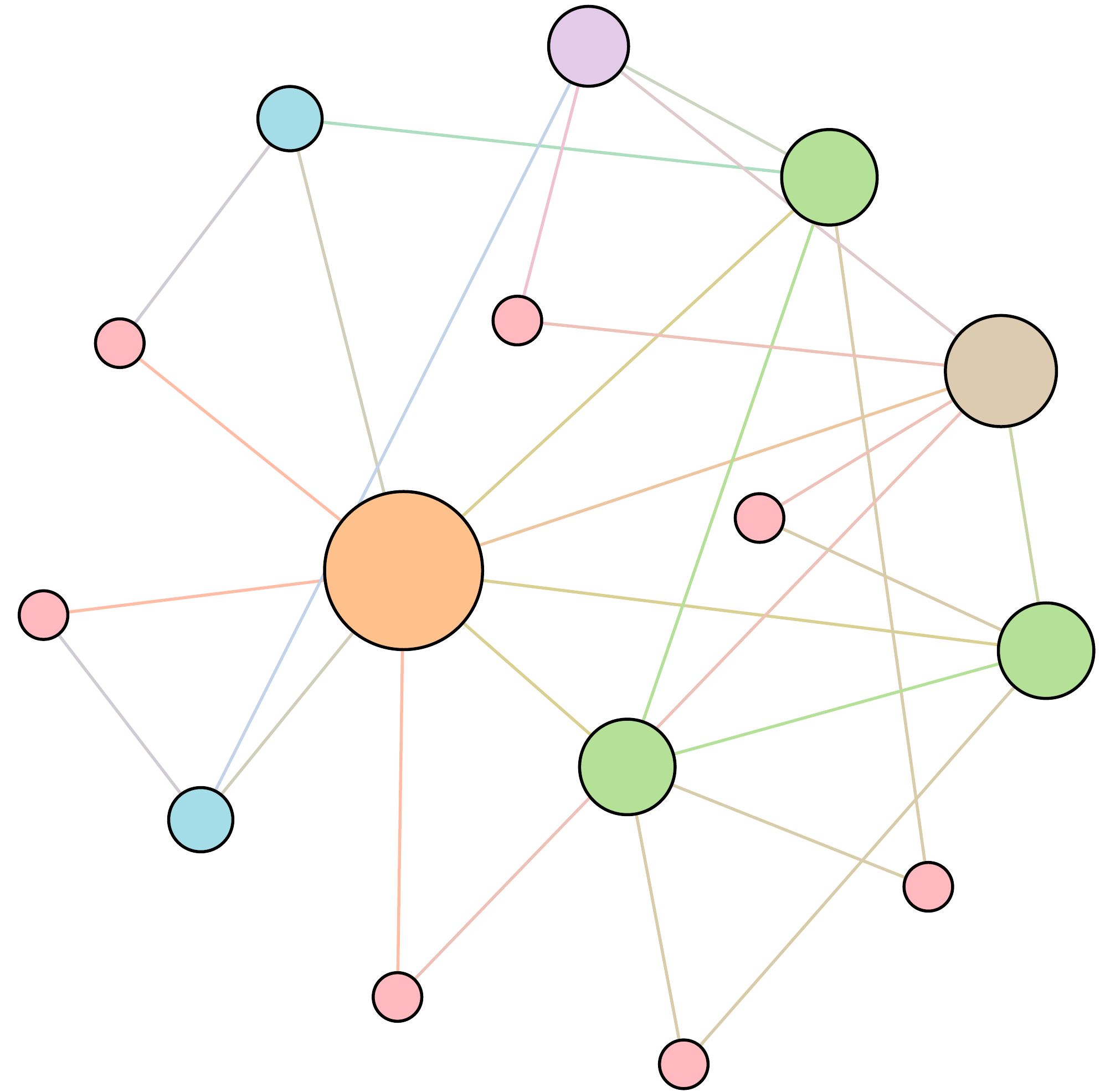}
        \caption{$\mathcal{AC}_B$ + $E_{local}$}
         \end{subfigure}
  
\caption{The resilience maximization on the BA-15 dataset with 15 nodes and 27 edges with  (a) original network, (b)-(j) results of defending the node degree-based attack with different combinations of resilience and utility, and (k)-(s) results of defending against the node betweenness-based attack with varying combinations of resilience and utility. For three resilience metrics, $\mathcal{R}$ denotes the graph connectivity-based resilience metric; $\mathcal{SR}$ is the spectral radius;  $\mathcal{SR}$ represents the algebraic connectivity. For two utility metrics, $E_{global}$ denotes the global efficiency, and $E_{local}$ is the local efficiency.}
\label{fig:casestudy}
\end{figure}

As shown in Figure \ref{fig:casestudy}, we conduct extensive experiments on the BA-15 network to demonstrate that ResiNet can learn to optimize graphs with different resilience and utility metrics and to defend against other types of attacks besides the node degree-based attack, such as the node betweenness-based attack.

Table \ref{tb:ba_15_varying_objs} records the improvements in percentage of ResiNet  for varying objectives on the BA-15 dataset. As visualized in Figure \ref{fig:casestudy}, ResiNet is not limited to defend against the node degree-based attack (Figure \ref{fig:casestudy} (b)-(j)) and also learns to defend against the betweenness-based attack (Figure \ref{fig:casestudy} (k)-(s)). Total three resilience metrics are used, with $\mathcal{R}$ denoting the graph connectivity-based resilience metric, $\mathcal{SR}$ being the spectral radius and $\mathcal{SR}$ representing the algebraic connectivity. Total two utility metrics are adopted, including the global efficiency $E_{global}$ and the local efficiency $E_{local}$.
Not surprisingly, the optimized network with an improvement of about 3.6\% for defending the betweenness-based attack also has a higher resilience (around 7.8\%) against the node-degree attack. This may be explained as the similarity between node degree and betweenness for a small network.


     


 \begin{table}[t]
  \caption{Performance gain (in percentage) of ResiNet in optimizing varying objectives on the BA-15 network. All objectives are optimized with the same hyper-parameters, which means that we did not tune hyper-parameters for objectives except for $R_D$.}
  \label{tb:ba_15_varying_objs}
  \centering
    \scalebox{0.9}{
  \begin{tabular}{ll|ll}
    \toprule
      Objective & Gain (\%)  & Objective & Gain(\%)  \\
    \midrule
    $\mathcal{R}_D$  & 35.3  & $\mathcal{R}_B$ & 14.6  \\
    $\mathcal{SR}_D$  & 15.3  &$\mathcal{SR}_B$ & 15.3  \\
    $\mathcal{AC}_D$  & 48.2 & $\mathcal{AC}_B$ & 43.2   \\
    $\mathcal{R}_D + E_{global}$  & 14.2 & $\mathcal{R}_B$ + $E_{global}$ & 13.1 \\
    $\mathcal{SR}_D$ + $E_{global}$ & 14.6  & $\mathcal{SR}_B$+$E_{global}$ &  15.0 \\
    $\mathcal{AC}_D$ + $E_{global}$ & 34.0  &  $\mathcal{AC}_B$ + $E_{global}$ & 31.3  \\
    $\mathcal{R}_D$ + $E_{local}$ & 24.4 &   $\mathcal{R}_B$ + $E_{local}$ & 39.4\\
    $\mathcal{SR}_D$ + $E_{local}$ & 17.3 & $\mathcal{SR}_B$ + $E_{local}$ & 21.2\\
    $\mathcal{AC}_D$ + $E_{local}$  & 9.4 & $\mathcal{AC}_B$ + $E_{local}$ & 15.1 \\
    \bottomrule
  \end{tabular}}
\end{table}

\subsection{Inductivity on larger datasets}
\label{appendix:extra_datasets_results}
\begin{figure}[t]
    \centering
   \begin{subfigure}{0.45\textwidth}  
         \centering
    \includegraphics[width=\textwidth]{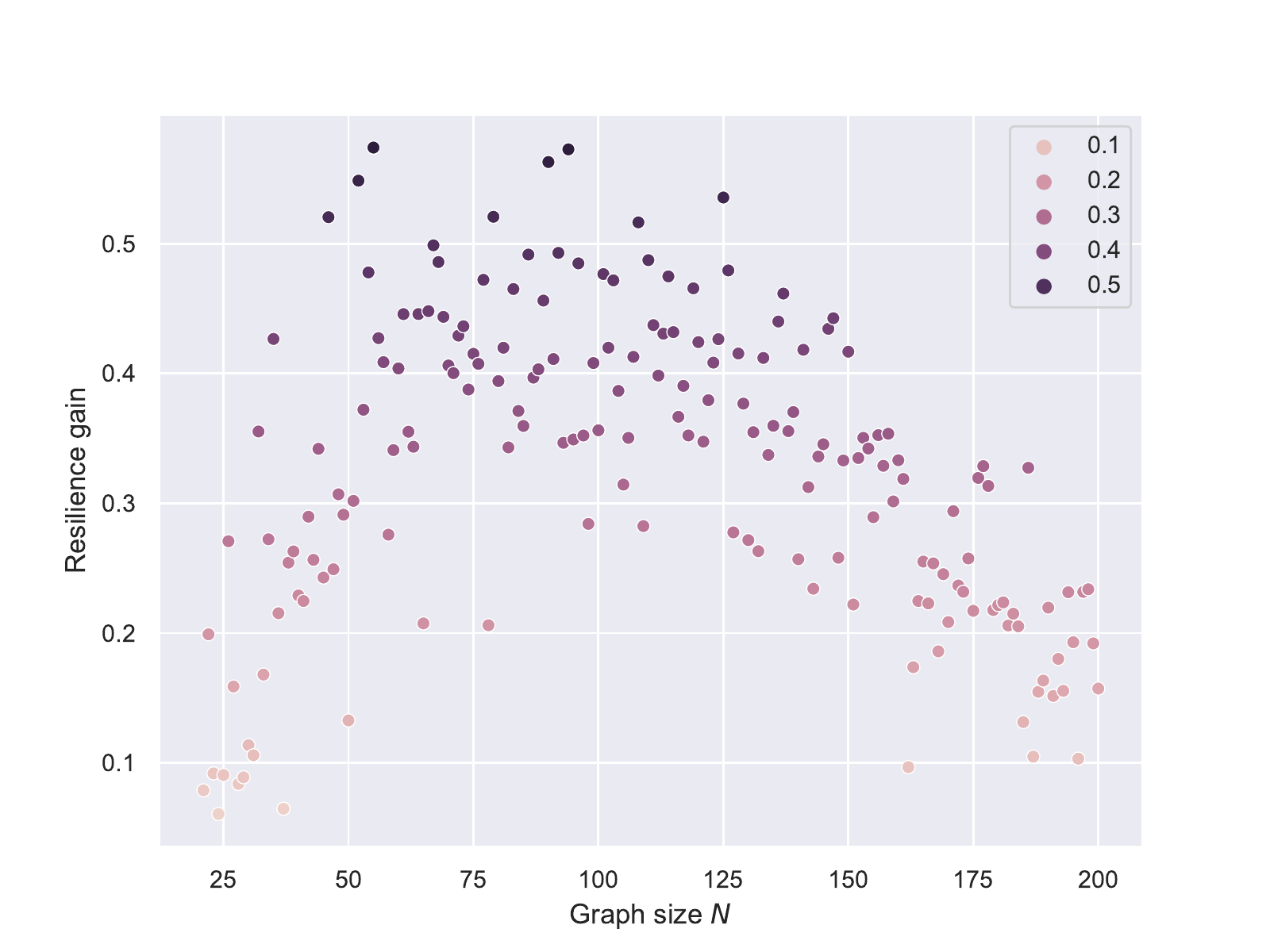}
    \caption{Inductivity on resilience}
     \end{subfigure}
      \hfill
      \begin{subfigure}{0.45\textwidth}  
         \centering
    \includegraphics[width=\textwidth]{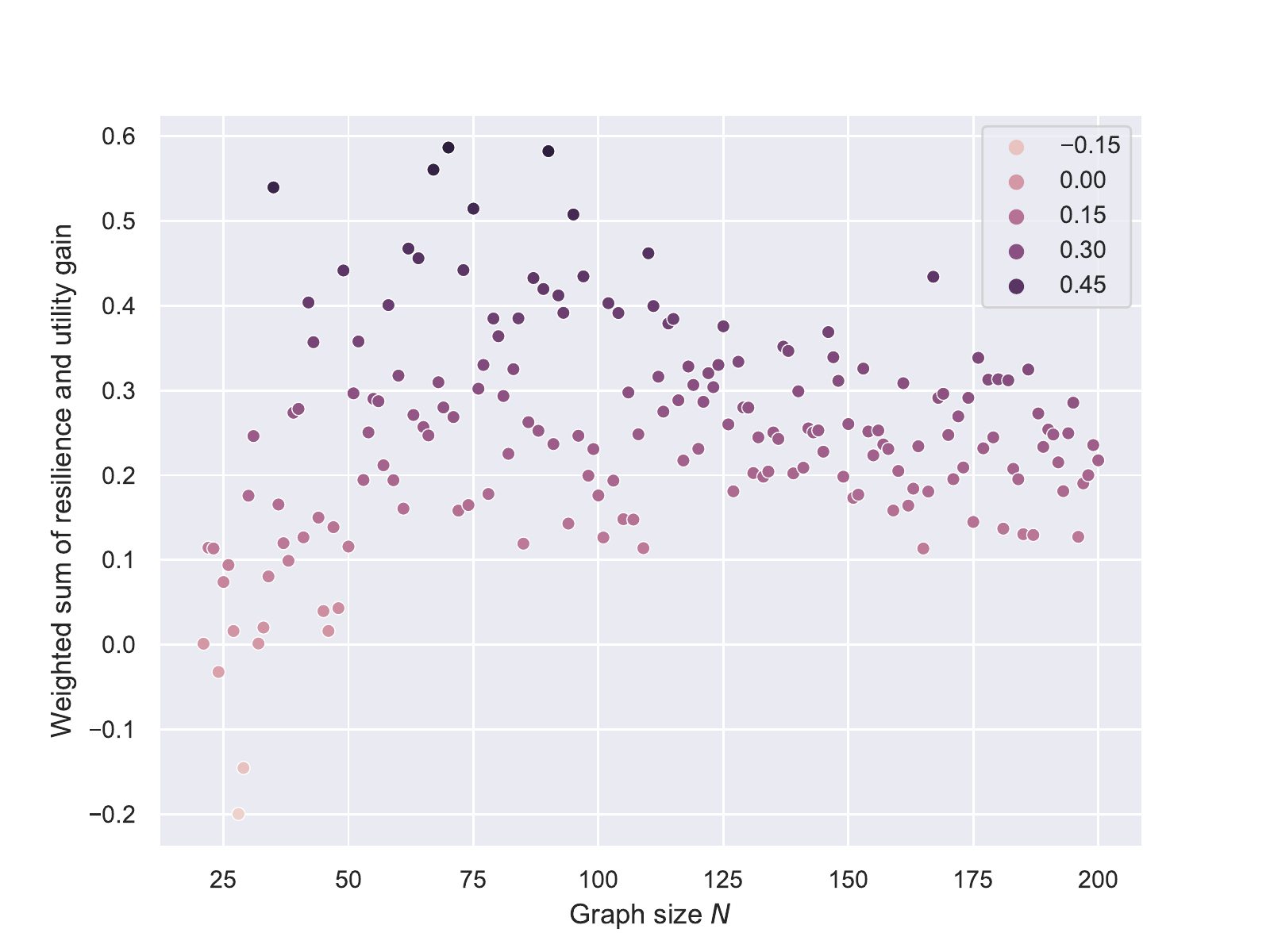}
    \caption{Inductivity on resilience and utility}
     \end{subfigure}
\caption{The inductive ability of ResiNet on the test dataset (BA-20-200) when optimizing (a) network resilience  and (b) the combination of resilience and utility.}
\label{fig:inductivity_ba_mixed}
\end{figure}

Even with limited computational resources, armed with the autoregressive action space and the power of FireGNN, ResiNet can be trained fully end-to-end on graphs with thousands of nodes using RL.  
We demonstrate the inductivity of ResiNet on graphs of different sizes by training ResiNet on the BA-20-200 dataset, which consists of graphs with the size ranging from 20 to 200, and then report its performance on directly guiding the edges selections on unseen test graphs. The filtration order $K$ is set to 1 for the computational limitation. As shown in Figure \ref{fig:inductivity_ba_mixed}, we can see that ResiNet has the best performance for $N \in [70, 100]$. The degrading performance with the graph size may be explained by the fact that larger graphs require a larger filtration order for ResiNet to work well. A more stable performance improvement of ResiNet is observed with the increment of graph size when trained to optimize network resilience and utility simultaneously, and ResiNet possibly finds a strategy to balance these two metrics.

\subsection{Inspection of optimized networks}

        



\begin{figure}
     \centering
     \begin{subfigure}[b]{0.25\textwidth}
         \centering
         \includegraphics[width=\textwidth]{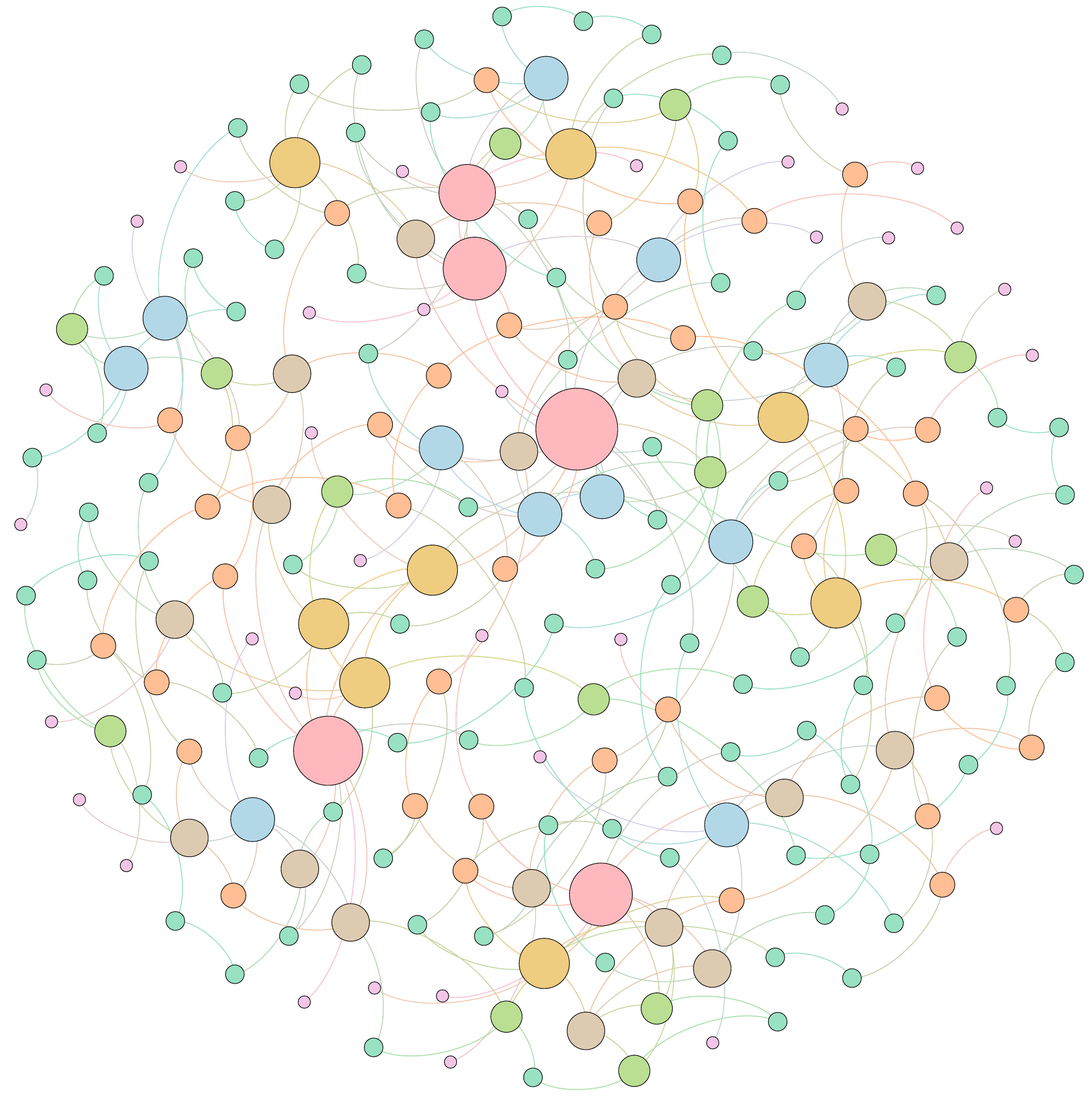}
         \caption{Original EU network}
     \end{subfigure}
     \hfill
     \begin{subfigure}[b]{0.25\textwidth}
         \centering
         \includegraphics[width=\textwidth]{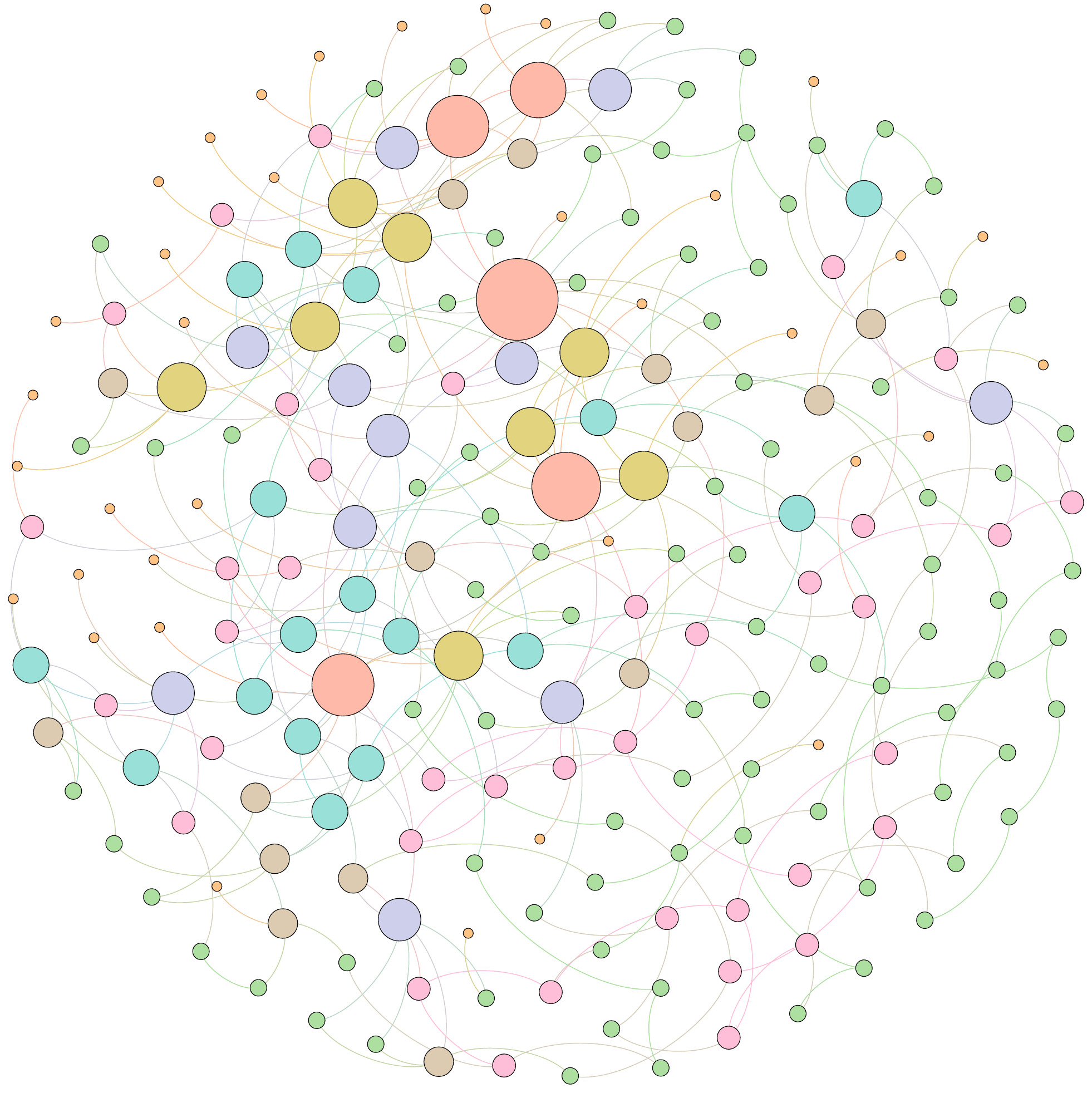}
         \caption{$\mathcal{R}$}
     \end{subfigure}
     \hfill
     \begin{subfigure}[b]{0.25\textwidth}
         \centering
         \includegraphics[width=\textwidth]{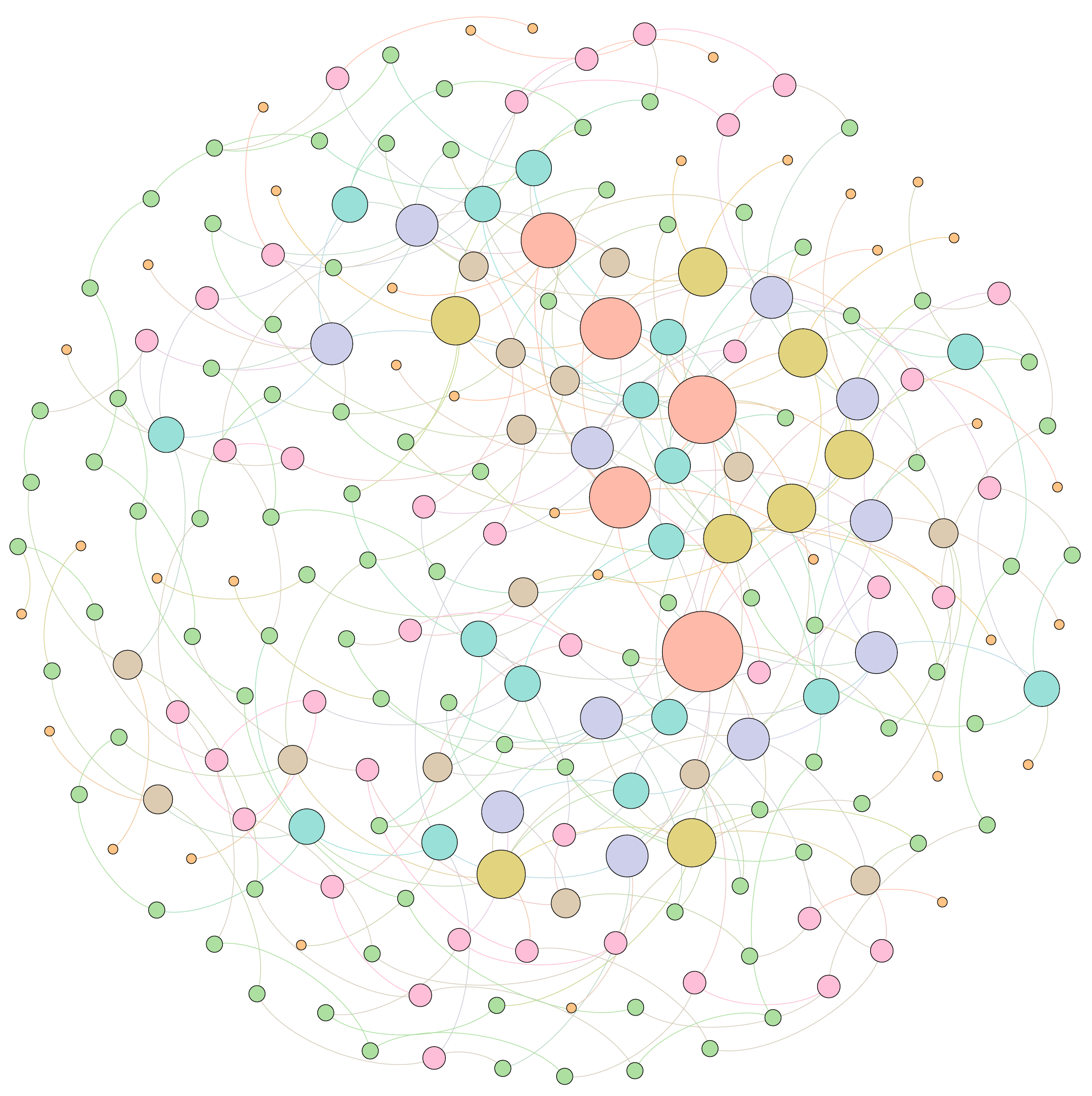}
         \caption{$E_{global}$}
     \end{subfigure}
        \caption{Visualizations of the original EU network and  optimized networks using ResiNet with different objectives: $\mathcal{R}$ means the connectivity-based resilience measurement and $E_{global}$ is the global efficiency.}
        \label{fig:results_EU}
\end{figure}

Moreover, to provide a deeper inspection into the optimized network structure, we take the EU power network as an example to visualize its network structure and the optimized networks given by ResiNet with different objectives. Compared to the original EU network, Figure \ref{fig:results_EU} (b) is the network structure obtained by only optimizing the graph connectivity-based resilience. We can observe a more crowded region on the left, consistent with the ``onion-like'' structure concluded in previous studies. If we consider the combination gain of both resilience and utility, we observe a more compact clustering ``crescent moon''-like structure as shown in Figure \ref{fig:results_EU} (c).





\subsection{Performance comparisons under a large rewiring budget}
In this section, we present the resilience improvement and the required number of edge rewiring of each algorithm under a large rewiring budget of 20. 
The running speed is also presented to compare the running time efficiency of each algorithm. 

As shown in Table \ref{tb:res_resi_gain_budget_200}, traditional methods improve the network resilience significantly compared to ResiNet under a large rewiring budget of 200. However, traditional methods are still undesired in such a case since a solution with a large rewiring budget is not applicable in practice due to the vast cost of adding many new edges into a real system. For example, the actual number of rewiring budget for EA is hard to calculate since it is a population-based algorithm, so it is omitted in Table~ \ref{tb:res_resi_gain_budget_200}. All baselines adopt the early-stopping strategy that they will terminate if there is no positive resilience gain in a successive 1000 steps.

Table \ref{tb:res_speed} indicates that the time it takes for the benchmark algorithm to solve the problem usually increases as the test data set size increases.
 In contrast, our proposed ResiNet is suitable for testing on a large dataset once trained.
\begin{table}
  \caption{Resilience optimization algorithm under the fixed maximal rewiring number budget of 200. 
  Entries are in the format of $X(Y)$, where 1) $X$: weighted sum of the graph connectivity-based resilience and the network efficiency improvement (in percentage); 2) $Y$: required rewiring number.
  Results are averaged over 3 runs and best performance is in bold.} 
 
  \label{tb:res_resi_gain_budget_200}
   
  \centering
    \scalebox{0.67}{
  
  \begin{tabular}{llllllllllll}
    \toprule
    Method  & $\alpha$ & BA-15 & BA-50 &  BA-100 & BA-500 & BA-1000 &  EU & P2P-Gnutella05 & P2P-Gnutella09  \\
     
    \midrule
    
    \multirow{2}{4em}{HC} & $0$ & 26.8 (10.0)  & 52.1 (47.0) & 76.9 (97.3) & 45.8 (200) & 302.5 (200) & 71.9 (152.7)  & 37.5 (193.3) & 40.2 (137.7) \\
    &$0.5$ & 18.6 (11.3) & 43.1 (62.7) & 56.9 (121) & 30.0 (200)  & 66.3 (200)  & 63.2 (200) & 27.7 (200) & 34.7 (196.3) \\

\hline
  \multirow{2}{4em}{SA} & $0$ & 26.8 (20)  & 49.7 (59.0) & 84.5 (119.7) & 43.2 (200) & 271.8 (200) & 73.5(160.3) & 37.1 (200) & 37.2 (134) \\
    &$0.5$ & 17.8 (21) & 41.1 (79.7) & 57.7 (127.7)  &  31.4 (200) & 64.9 (200) & 62.8 (200) & 37.1 (200) &  35.2 (200)\\
    
\hline
 \multirow{2}{4em}{Greedy} & $0$ & 23.5 (6)  & 48.6 (13) & 64.3 (20) & \ding{55} & \ding{55} & 0.5 (3) & \ding{55} & \ding{55} \\
    &$0.5$ & 5.3 (15)  & 34.7 (13) & 42.7 (20)  & \ding{55} & \ding{55} & 0.3 (3) & \ding{55} & \ding{55} \\   
\hline
 \multirow{2}{4em}{EA} & $0$ & 35.3 (\ding{55}) & 50.2 (\ding{55})  &  61.9 (\ding{55}) & 9.9 (200) & 174.1 (200) & 66.2 (\ding{55}) & 2.3 (200) & 0 (200) \\
    &$0.5$ & 27.1 (\ding{55}) & 38.3 (\ding{55}) & 46.6 (\ding{55}) & 6.8 (200) & 18.7 (200) & 58.4 (\ding{55}) & 3.2 (200)  & 0 (200) \\
\hline
\hline
\multirow{2}{4em}{DE-GNN-RL} & $0$ & 13.7 (2) &  0 (1)  & 0 (1)  & 1.6 (20)  & 41.7 (20) & 9.0 (20) & 2.2 (20) & 0 (1) \\
&$0.5$ &  10.9 (2) &  0 (1) & 0 (1) & 2.7 (20) & 20.1 (14) & 2.1 (20) & 0 (1) &  1.0 (20) \\
\hline
\multirow{2}{4em}{$k$-GNN-RL} & $0$ & 13.7 (2) &  0 (1)  & 0 (1)  & 0 (1)  & 8.8 (20) & 4.5 (20) & -0.2 (20) & 0 (1) \\
&$0.5$ &  6.3 (2) &  0 (1) & 0 (1) & 0 (20) & -24.9 (20) & 4.8 (20) & -0.1 (20) &  0 (1) \\
\hline
 \multirow{2}{4em}{ResiNet} & $0$ & \textbf{35.3} (6) & \textbf{61.5} (20) & \textbf{70.0} (20) & \textbf{10.2} (20) & 172.8 (20) &  \textbf{54.2} (20) & \textbf{14.0} (20) & \textbf{18.6} (20) \\
    &$0.5$ & \textbf{26.9} (20) & \textbf{53.9} (20) &  \textbf{53.1} (20) &\textbf{ 15.7} (20) & \textbf{43.7} (20) & \textbf{51.8} (20) & \textbf{12.4} (20) & \textbf{15.1} (20) \\

    \bottomrule
  \end{tabular}
  }
\end{table}

\begin{table}
  \caption{Running speed (in second) of the resilience optimization algorithm  under the fixed maximal rewiring number budget.
    Entries are in the format of $X(Y)$, where 1) $X$: speed under the budget of 20; 2) $Y$: speed under the budget of 200 .
  \ding{55} means that the result is not available at a reasonable time. Results are averaged over 3 runs and best performance is in bold.}  
  \label{tb:res_speed}
   
  \centering
   \scalebox{0.67}{
  \begin{tabular}{llllllllllll}
    \toprule
    Method  & $\alpha$ & BA-15 & BA-50 &  BA-100 & BA-500 & BA-1000 &  EU & P2P-Gnutella05 & P2P-Gnutella09 \\
     
    \midrule
    
    \multirow{2}{4em}{HC} & $0$ & 1.0 (1.0)  & 1.1 (6.4) & 1.3 (22.2) & 21.9 (354.1) & 80.3 (1288.3) & 3.1 (94.2) & 15.3 (358.1) & 4.5 (89.1) \\
    &$0.5$ & 1.5 (11.5) & 1.1 (12.8) & 2.0 (49.0) & 40.9 (589.5) & 148.7 (2603.7) & 5.3 (193.7) & 24.7 (462.8) & 7.0 (190.6) \\

\hline
  \multirow{2}{4em}{SA} & $0$ & 0.5 (0.5)  & 0.3 (6.6) & 0.6 (22.6) & 12.2 (313.0) & 45.7 (1051.8) &2.4 (91.2) & 10.8 (286.4) & 2.6 (89.4)\\
    &$0.5$ & 0.7 (1.7) & 0.7 (13.2) & 1.7 (47.5)  & 33.9 (568.9) & 99.8 (2166.3) & 5.0 (193.5) & 23.9 (454.5) & 6.3 (188.5)\\
    
\hline
\multirow{2}{4em}{Greedy} & $0$ & 0.2 (6.0)  & 34.1 (34.5) & 766.3 (\ding{55}) & \ding{55} & \ding{55} & 3061.7 (\ding{55}) & \ding{55} & \ding{55}   \\
&$0.5$ & 0.7 (0.7) & 64.1 (65.4) & 1478.9 (\ding{55})  & \ding{55} & \ding{55} & 6192.6 (\ding{55}) & \ding{55} & \ding{55} \\
\hline
 \multirow{2}{4em}{EA} & $0$ & 0.01 (\ding{55}) & 0.1 (\ding{55})  &  1.6 (\ding{55}) & 2.5 (\ding{55}) & 10.3 (\ding{55}) & 0.2 (\ding{55}) & 1.6 (\ding{55}) & 0.4 (\ding{55})  \\
    &$0.5$ & 0.01 (\ding{55})  & 0.1 (\ding{55}) & 0.8 (\ding{55}) & 4.7 (\ding{55}) &  15.0 (\ding{55}) & 0.4 (\ding{55}) & 3.0 (\ding{55}) & 0.8 (\ding{55}) \\
\hline
\hline
\multirow{2}{4em}{DE-GNN-RL} & $0$ & 0.1 (\ding{55}) & 0.1 (\ding{55}) & 0.1 (\ding{55}) & 14.9 (\ding{55}) & 70.3 (\ding{55}) & 3.6 (\ding{55}) & 8.7 (\ding{55}) & 0.5 (\ding{55}) \\
    &$0.5$ & 0.1 (\ding{55}) & 0.1 (\ding{55})&  0.2 (\ding{55}) & 13.7 (\ding{55}) & 60.9 (\ding{55}) & 4.5 (\ding{55}) & 1.0 (\ding{55}) & 6.7 (\ding{55}) \\
\hline
\hline
\multirow{2}{4em}{$k$-GNN-RL} & $0$ &  0.02 (\ding{55}) & 0.03 (\ding{55}) & 0.07 (\ding{55}) & 1.3 (\ding{55}) & 56.5 (\ding{55}) & 2.6 (\ding{55}) & 8.2 (\ding{55}) & 0.5 (\ding{55}) \\
    &$0.5$ & 0.02 (\ding{55}) & 0.04 (\ding{55})& 0.08  (\ding{55}) & 18.3  (\ding{55}) & 76.1  (\ding{55}) & 3.6  (\ding{55}) & 11.5 (\ding{55}) & 0.6 (\ding{55}) \\
\hline

 \multirow{2}{4em}{ResiNet} & $0$ & 0.5 (\ding{55}) & 1.8  (\ding{55}) & 2.2 (\ding{55}) & 17.5 (\ding{55}) & 66.8 (\ding{55}) & 4.5  (\ding{55}) & 14.7  (\ding{55}) & 9.3 (\ding{55}) \\
    &$0.5$ & 0.5 (\ding{55}) & 1.9 (\ding{55}) & 2.4  (\ding{55}) & 18.0 (\ding{55}) & 67.5 (\ding{55}) &  5.2 (\ding{55}) & 15.0 (\ding{55}) &  10.3 (\ding{55}) \\
    \bottomrule
  \end{tabular}
  }
\end{table}

\end{document}